\pdfoutput=1

\documentclass[11pt]{article}

\usepackage[final]{acl}
\usepackage{amssymb}
\usepackage{mathrsfs}
\usepackage{hyperref}
\usepackage{amsmath}
\usepackage{float}
\usepackage{tabularx} 
\usepackage[title]{appendix}
\usepackage{mwe}
\usepackage{wrapfig}
\usepackage{caption}
\usepackage{subcaption}
\usepackage{booktabs}
\usepackage{bbm}
\usepackage{xcolor}
\usepackage{graphicx}
\usepackage{pifont}
\usepackage{multirow} 
\usepackage{xcolor,colortbl}
\usepackage{times}
\usepackage{latexsym}
\usepackage[ruled,vlined,linesnumbered]{algorithm2e}

\usepackage{times}
\usepackage{latexsym}

\usepackage[T1]{fontenc}

\usepackage[utf8]{inputenc}

\usepackage{microtype}

\usepackage{inconsolata}

\usepackage{graphicx}

\newcommand{\method}{SSU\xspace} 

\title{Avoiding Copyright Infringement via Large Language Model Unlearning }

\author{Guangyao Dou$^1$ \quad Zheyuan Liu$^2$ \quad Qing Lyu$^1$ \quad \textbf{Kaize Ding}$^3$ \quad {\bf Eric Wong$^1$}\\ \\
        $^1$University of Pennsylvania \quad $^2$University of Notre Dame \quad $^3$Northwestern University \\ \\
       {\texttt gydou@seas.upenn.edu}
}       

\begin{document}
\maketitle
\begin{abstract}
Pre-trained Large Language Models (LLMs) have demonstrated remarkable capabilities but also pose risks by learning and generating copyrighted material, leading to significant legal and ethical concerns. In real-world scenarios, model owners need to continuously address copyright infringement as new requests for content removal emerge at different time points. This leads to the need for sequential unlearning, where copyrighted content is removed sequentially as new requests arise. Despite its practical relevance, sequential unlearning in the context of copyright infringement has not been rigorously explored in existing literature.
To address this gap, we propose \textbf{S}table \textbf{S}equential \textbf{U}nlearning (\textbf{SSU}), a novel framework designed to unlearn copyrighted content from LLMs over multiple time steps. Our approach works by identifying and removing specific weight updates 
in the model's parameters that correspond to copyrighted content. We improve unlearning efficacy by introducing random labeling loss and ensuring the model retains its general-purpose knowledge by adjusting targeted parameters. 
Experimental results show that \method achieves an effective trade-off between unlearning efficacy and general-purpose language abilities, outperforming existing baselines. \footnote{Code is available at \href{https://github.com/guangyaodou/SSU_Unlearn}{guangyaodou/SSU}.}

\end{abstract}

\section{Introduction}

\begin{figure}[t]
    \centering
    \includegraphics[width=1\linewidth]{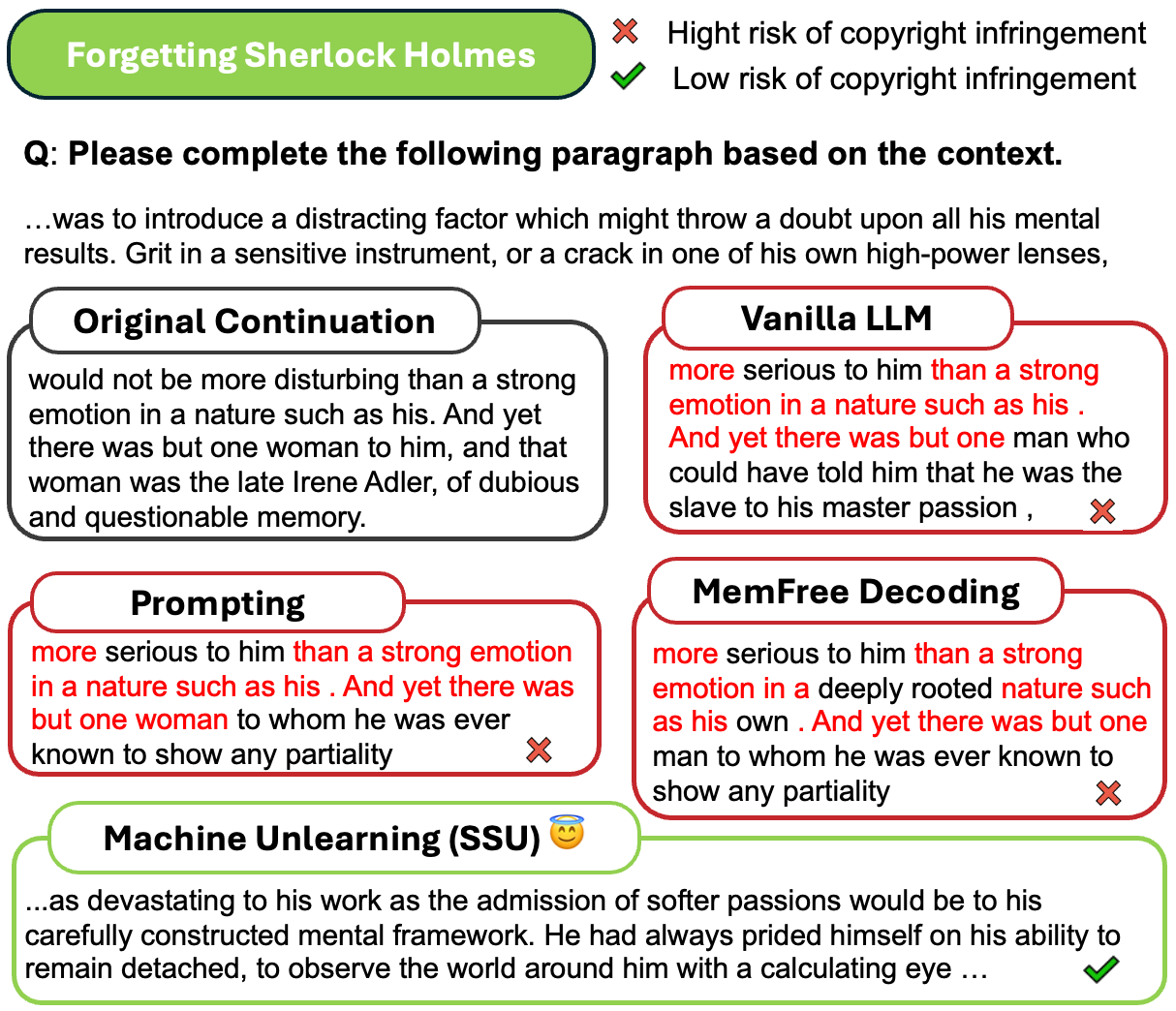}
    \vspace{-0.20in}
    \caption{Continuations of a passage from Sherlock Holmes under different copyright takedown methods. The original continuation serves as the ground truth. The vanilla model, prompting, and MemFree decoding exhibit high risk of copyright infringement. In contrast, SSU produces a continuation that is transformative to avoid copyright infringement. \protect\footnotemark}
    \label{fig:thumbnail_example}
    \vspace{-0.24in}
\end{figure}
\footnotetext{This example is for illustration purposes only, as Conan Doyle's Sherlock Holmes entered the public domain in 2023.}

In December 2023, the New York Times filed a lawsuit against OpenAI, accusing it of training its Large Language Models (LLMs) on copyrighted material without permission\footnote{\href{https://nytco-assets.nytimes.com/2023/12/NYT_Complaint_Dec2023.pdf}{NYT Complaint, Dec 2023}}. This legal challenge highlights the growing concern over LLMs incorporating copyrighted content from vast pre-training datasets, which are often composed of publicly available text ~\cite{brittain2023us, rahman2023beyond}. Despite the significant progress LLMs have made through learning from diverse text data ~\cite{brown2020language,chowdhery2023palm, touvron2023llama}, screening out copyrighted material remains an immense challenge~\cite{duarte2024cop}. These issues raise broader questions about fair use of generative models.

There are two times when copyright interacts with LLMs. The first is when LLMs learn from copyrighted materials, which is arguably fair use (but this has not been tested in court). The second is when LLMs generate outputs. If a generated output is substantially similar to copyrighted work it has trained on, then this is more likely to be a copyright infringement. If a court found an AI model developer to be in violation of copyright, then the court may require that the developer to remove that copyrighted work from the model. The cost of retraining from scratch leaving out one copyrighted work is exorbitantly high. Therefore, as an alternate remedy, a court may ask for a
copyright takedown that does not require a full retraining of the model. This motivates research into unlearning and other copyright takedown methods.

Previous works have investigated post-hoc copyright takedown methods -- mitigating risks of generating copyrighted contents -- using system prompt and decoding time intervention such as the MemFree Decoding \cite{ippolito2022preventing}. Additionally, an alternative solution is \textit{machine unlearning}~\cite{cao2015towards}, which removes unwanted knowledge after pre-training, reconfiguring the model as if it had never learned that data (Figure \ref{fig:thumbnail_example}). Recent works proposed practical machine unlearning algorithms for LLMs \cite{zhang2024negative, chen2023unlearn, jang2023knowledge, zhao2024towards}. However, few have addressed the challenge of \textit{sequentially} unlearning copyrighted content, where multiple unlearning requests must be processed over time without retraining from scratch. A key difficulty lies in controlling the changes in model weights during this process—existing methods often struggle to prevent unintended drift, leading to drastic degradation of general-purpose language abilities, leaving it unclear if they are suitable as copyright takedown methods. An effective unlearning algorithm should be \textit{stable}, meaning it should ensure \textit{unlearning efficacy} while preserving non-targeted knowledge, knowledge that are not subject to unlearning, and general-purpose abilities.

The core of many previous LLM unlearning methods have focused on Gradient Ascent (GA) and further combined it with an in-distribution retained dataset to preserve general-purpose language abilities, known as Gradient Difference~\cite{maini2024tofu, zhao2024towards, liu2024towards, yao2024machine}. However, Gradient Difference requires collection of a substantial amount of in-distribution retained data to maintain general-purpose abilities. Moreover, GA-based methods risk catastrophic collapse, where the model's general-purpose language abilities degrade significantly~\cite{liu2024towards}. \citet{zhang2024negative} proposed Negative Preference Optimization (NPO), framing unlearning as preference optimization. However, NPO relies on a reference model, and if the reference model contains copyrighted information, unlearning efficacy is compromised.

To address these challenges in sequentially unlearning copyrighted books, we propose \textbf{S}table \textbf{S}equential \textbf{U}nlearning (\method), that achieves a better trade-off between effective unlearning and maintaining general-purpose language abilities in sequential settings. Specifically, \method first fine-tunes the model on the copyrighted books, followed by fine-tuning with random labels. During gradient updates, \method applies targeted weight adjustments through weight saliency.
Afterwards, it extracts task vectors \cite{ilharco2022editing} corresponding to the copyrighted books and subsequently negates these task vectors to achieve unlearning. Unlike Gradient Difference methods, \method does not require additional retained data collection to maintain its performance on other tasks, thereby avoiding the complexity and overhead associated.

Our experiments on the Llama-3.1-8B-Instruct \cite{dubey2024llama} and Mistral-7B-Instruct-v0.3 \cite{jiang2023mistral} show that \method achieves a superior trade-off between unlearning efficacy and general-purpose language abilities, avoiding the catastrophic collapse. Moreover, \method consistently outperforms NPO, which employs preference optimization frameworks. Additionally, fine-tuning with random loss and targeted model updates play distinct roles in facilitating the unlearning process within \method. In contrast, copyright takedown methods that do not involve model weight updates, such as system prompts and MemFree Decode, fail to  mitigate the risks of copyright infringement.

Our main contributions are:

\begin{itemize}
   \item We present the first investigation into the sequential unlearning of copyrighted literary works to address copyright infringement, formalizing the task and defining evaluation metrics for effectiveness. 
    \item We propose \method, a stable unlearning algorithm for sequential settings, which achieves a superior trade-off between mitigating copyright infringement and preserving reasoning capabilities compared to existing methods.
    \item We systematically evaluate existing methods within the sequential unlearning setting and demonstrate that they either provide limited mitigation of copyright infringement or suffer from catastrophic collapse.
  
\end{itemize}\

\section{Related Work}

\paragraph{LLM Memorization and Copyright} 
LLMs are known to memorize and reproduce training data from the pre-training stage~\cite{carlini2021extracting, carlini2022quantifying, zhang2023counterfactual, nasr2023scalable, liu2024shield}. This behavior has prompted recent studies exploring the connection between verbatim memorization and copyright infringement~\cite{chu2024protect, huang2024demystifying, meeus2024copyright, karamolegkou2023copyright}, as well as the fair use of foundation models~\cite{henderson2023foundation}. Various methods have been proposed to mitigate these risks. For instance, \citet{ippolito2022preventing} introduced the MemFree decoding to reduce the likelihood of copyright infringement but failed to capture non-consecutive verbatim content. Additionally, \citet{min2023silo} proposed SILO, a framework that utilizes a nonparametric datastore containing high-legal-risk data. However, the SILO framework does not address the risks associated with retrieval augmented generation~\cite{wei2024evaluating}, nor does it adequately address the practical challenges of ensuring that the training data for parametric models is limited to permissive content.

\paragraph{Machine Unlearning}
Machine Unlearning \cite{cao2015towards} have been proposed for LLMs to remove harmful knowledge~\cite{liu2024towards, yao2023large}, private information \cite{dong2024unmemorization}, and copyrighted content~\cite{jang2022knowledge, eldan2023s, liu2024breaking}. Additionally, \citet{maini2024tofu} and \citet{yao2024machine} explored the \textit{right to be forgotten} and introduced benchmarks for evaluating the unlearning effectiveness of private data. \citet{li2024wmdp} further introduced a benchmark for forgetting hazardous knowledge in LLMs. However, none of these works have addressed the unlearning of copyrighted literary works in a sequential setting, nor the limitations of existing methods in this context.

\paragraph{Model Editing}
A related but distinct process is model editing~\cite{ meng2022mass, hewitt2024model,gupta2024unified}. While unlearning aims to erase specific information from a model, model editing focuses on correcting errors without removing the underlying learned information~\cite{liu2024machine}. Recent studies have shown that model editing in sequential settings can lead to model collapse~\cite{gupta2024model, yang2024butterfly}, and potential solutions have been proposed to mitigate this issue~\cite{gupta2024rebuilding}.

\begin{figure*}[ht]
  \centering 
  \includegraphics[width=0.92
  \textwidth]{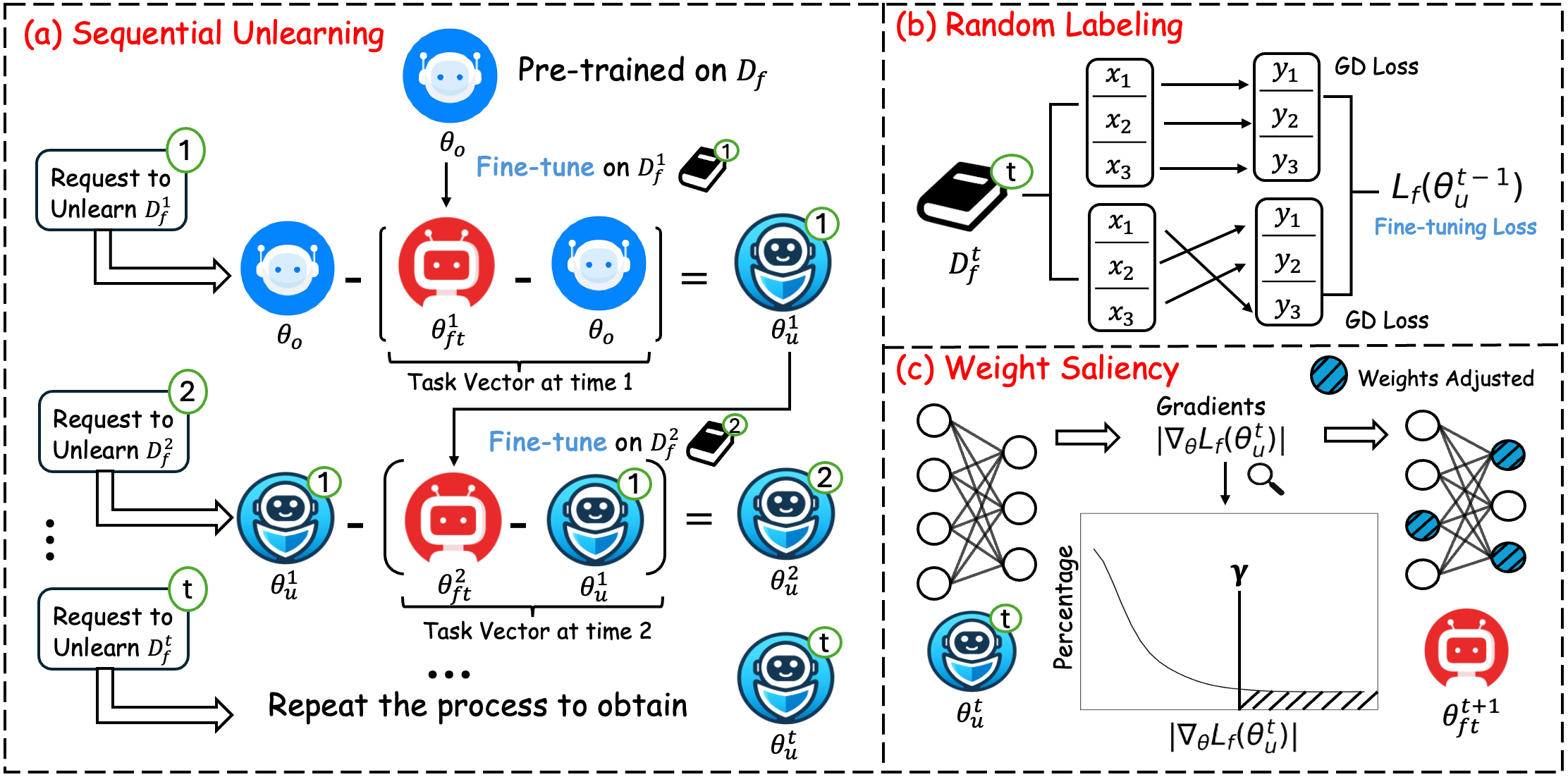}
  \vspace{-0.1in}
  \caption{
  \textbf{Overall process of our unlearning framework.} \textbf{(a)} At each time step \( t \), an unlearning request is received to forget the dataset \( D_f^t \). The unlearning algorithm involves first fine-tuning $\theta_{u}^{t-1}$ on $D_f^t$ to obtain $\theta_{ft}^{t}$, and then subtracting the task vector from previously unlearned model $\theta_{u}^{t-1}$. \textbf{(b)} At each time step t. we compute the gradient loss and random labeling loss to obtain the objective $L_f(\theta_{u}^{t-1})$ that will be used for fine-tuning. \textbf{(c)} At time step \(t+1\), we fine-tune $\theta_{u}^{t}$ using the objective we obtained in (b), and only update model weights that are most salient using weight saliency mapping.
  } 
  \vspace{-0.15in}
  \label{fig:method_pipeline}
\end{figure*}

\section{Preliminaries}

\subsection{Machine Unlearning for LLMs}

Consider the original model and its weights, denoted as $\theta_o$. Machine unlearning involves the problem where, given a dataset $D = \{(x_i, y_i)\}_{i=1}^N$ that $\theta_o$ was trained on, we aim to intentionally forget a subset of data, denoted as $D_f$, to obtain a modified model, denoted as $\theta_{u}$, such that $\theta_{u}$ behaves as if it has never seen $D_f$ during pre-training. 

In the context of machine unlearning, we often use a retrained model excluding $D_f$ during pre-training as a gold baseline. However, retraining a model for LLMs is extremely expensive and impractical in real-world settings.

\subsection{Task Arithmetic}

Unlearning via negating \textit{task vectors} has recently gained attention~\cite{ilharco2022editing, liu2024towards} and has become an important baseline approach for many unlearning tasks. The rationale behind this approach is that by negating the gradient updates of the unwanted data, we can achieve a more localized unlearning algorithm to effectively erase $D_f$ from $\theta_o$.

Specifically, the process involves two stages. First, we perform standard gradient descent to fine-tune $\theta_o$ on $D_f$, resulting in $\theta_{ft}$. Next, we calculate the task vector as the element-wise difference $\theta_{ft} - \theta_o$. We then negate this task vector from $\theta_o$ to derive the unlearned model $\theta_{u}$, expressed as $\theta_{u} = \theta_o - (\theta_{ft} - \theta_o)$.

\subsection{Unlearning with Multiple Time Steps}

We formally define sequential unlearning as the process where a model, originally trained on a dataset $D$, is incrementally modified to forget subsets of data at multiple time steps while preserving knowledge from the remaining data. Let $D$ be the original dataset, and let $D_f^t \subseteq D$ denote the subset of data that must be forgotten at time step $t$, where $t = 1, 2, \dots, T$. The cumulative set of all data to be forgotten over time is defined as: $D_f = \cup_{t=1}^T D_f^t$. Let $D_r$ represent the subset of data to be retained, such that
$D_r = D \setminus D_f$,  $\ D_f \cap D_r = \emptyset$, and $D_f \cup D_r = D$.

At each time step $t$, the unlearning process modifies the model to forget the subset $D_f^t$, resulting in a sequence of models $\{\theta_u^1, \theta_u^2, \ldots, \theta_u^T\}$. Each $\theta_u^t$ is the model obtained after unlearning the data subsets $D_f^1, D_f^2, \dots, D_f^t$ sequentially. Formally, the goal of sequential unlearning is to ensure that after each unlearning step, the model $\theta_u^t$ minimizes the influence of $D_f^t$ while retaining as much general-purpose knowledge from $D_r$ as possible.

\section{Methods}

This section presents \method, 
 which leverages task vectors, incorporates additional loss term for ensuring unlearning efficacy and uses a gradient-based weight saliency map to ensure general-purpose language abilities. The overall process is shown in Figure \ref{fig:method_pipeline}. We first explain unlearning at each time step in section \ref{methods:learning_with_tv}, and then generalize it to sequential setting in section \ref{methods:learning_sequential}.

\subsection{Learning Stable Task Vectors}
\label{methods:learning_with_tv}
First, we present the case of unlearning during the first time step. This means that $t = 1$, $D_f^1 = D_f$, and $\theta_u^0 = \theta_o$. Following the intuition from task vectors, we first need to fine-tune a model on $D_f$. To do this, we define $h_\theta(x, y_{y<i}) = \mathbb{P}(y_i | (x, y_{<i}); \theta)$, which is the probability of the token $y_i$ conditioned on the prompt $x$ and the already generated tokens $y_{<i} = [y_1, y_2, ..., y_{i-1}]$. Next, we define the LLM's loss on $y$ as: \begin{equation}
    L(x, y; \theta) := \sum_{i=1}^{|y|} \ell (h_\theta(x, y_{<i}), y_i),
\end{equation}
in which $l$ is the cross-entropy loss. 

Suppose $\theta_t$ is the current LLM through unlearning process. The first goal is to obtain a model that forgets $D_f$. Specifically, we define our first gradient descent loss term as:
\begin{equation}
    \mathcal{L}_{\text{fgt}} = \sum_{(x_{\text{fgt}}, y_{\text{fgt}}) \in D_f} L(x_{\text{fgt}}, y_{\text{fgt}}, \theta_o).
\end{equation}

\paragraph{Random Labeling Loss} Inspired by previous works demonstrating that injecting noise during training improves learning outcomes \cite{miyato2016adversarial, srivastava2014dropout, neelakantan2015adding}, we propose enhancing the effectiveness of unlearning by introducing data augmentation. Specifically, as shown in Figure \ref{fig:method_pipeline} (b), we randomly mismatch the outputs of $D_f$ with the inputs of $D_f$. During the first stage of the task vector approach, we include the following loss:

\begin{equation}
    \mathcal{L}_{\text{rnd}} := \sum_{(x_{\text{fgt}}, ) \in D_{f}} \frac{1}{|D_{f}|} \sum_{(,y_{\text{rnd})} \in D_{f}} L(x_{\text{fgt}}, y_{\text{rnd}}, \theta_t),
\end{equation}
in which $y_{\text{rnd}}$ is any output from $D_f$ and not corresponds to $x_{\text{fgt}}$.

By incorporating this random labeling loss, we introduce controlled noise into the unlearning process. This helps to prevent ``overfitting'' and enhance the stability of unlearning. Combining two loss terms, the final objective can be expressed as: 
\begin{equation}
\label{equation:ssu_main_no_weight_saliency}
    L_f(\theta_t) = \epsilon_1   \mathcal{L}_{\text{fgt}} + \epsilon_2  \mathcal{L}_{\text{rnd}}.
\end{equation} 

\paragraph{Weight Saliency} 
While performing sequential unlearning, we implicitly leverage the knowledge of previous steps. Specifically, as new requests are introduced, the most recent unlearning step adjusts the model weights based on the parameters from earlier unlearning steps. One of the main challenges with existing methods is that they often fail to control the changes in model weights during this process, which can lead to instability and faster degradation of the model's general-purpose capabilities.

Therefore, to preserve general-purpose language abilities, we should mitigate the risk of catastrophic collapse during each time step of sequential unlearning. We can achieve this by steering the unlearning process towards specific parts of the model weights that are most relevant to the data to be forgotten. 

As shown in Figure \ref{fig:method_pipeline} (c), we use a gradient-baased weight saliency map \cite{fan2023salun} during the first stage of fine-tuning to further ensure localized unlearning by only adjusting specific weights that are most influenced by the data to be forgotten. The weight saliency map is defined as:
\begin{equation}
\label{equation:ssu_main_weight_saliency_mask}
     m_s = \mathbbm{1}(\lvert \nabla_{\theta} L_f (\theta_t) \rvert \geq \gamma),
\end{equation}
in which $\mathbbm{1}(f \geq \gamma)$ is an element-wise indicator function which outputs one for the $i$-th element if $f_i \geq \gamma$, and 0 otherwise, and $\nabla_{\theta} L_f (\theta_t)$ is a gradient vector. 

Next, we apply the weight saliency mapping on the parameters that are most salient to unlearning and have the learned model as at each gradient accumulation step as:
\begin{equation}
\label{equation:objective_equation}
    \theta_{t+1} = m_s \odot (\Delta\theta + \theta_t) + (1 - m_s) \odot \theta_t,
\end{equation}
where $\Delta\theta$ indicates model updates. After training for $T$ gradient accumulation steps using Equation \ref{equation:objective_equation}, we obtain a fine-tuned model $\theta_{ft}^1$. Finally, we obtain our modified model using task vector by negating the knowledge of $D_f$ learned during the fine-tuning process from the original model as:
\begin{equation}
    \theta_u^1 = \theta_o - (\theta_{ft}^1- \theta_{o}).
\end{equation}

\subsection{Sequential Unlearning}
\label{methods:learning_sequential}
In this section, we explain how we generalize unlearning to sequential setting. As shown in Figure \ref{fig:method_pipeline}, at each new time step \( t \), we have the previously unlearned model $\theta_{u}^{t-1}$. Once we receive a new unlearning request, we will fine-tune $\theta_{u}^{t-1}$ using equation \ref{equation:objective_equation} as discussed in section \ref{methods:learning_with_tv} and obtain $\theta_{ft}^{t}$. Lastly, we obtain a new unlearned model at time step \( t \) as:
\begin{equation}
    \theta_{u}^t = \theta_{u}^{t-1} - (\theta_{ft}^t - \theta_{u}^{t-1}).
\end{equation}
If more unlearning requests are received, we will iteratively apply the same process to obtain a newer unlearned model. Notably, unlike Gradient Difference, \method does not require any additional retained dataset when calculating equation \ref{equation:ssu_main_no_weight_saliency}, ensuring efficiency and simplicity in real-world applications.

\section{Experimental Setup} In this paper, we choose the removal of copyrighted books from LLMs as a representative scenario for authors to exercise their \textit{intellectual property rights}. Machine unlearning can be applied to these LLMs to unlearn these books, thereby preventing the generation of verbatim copyrighted content.

\subsection{Setting}
We aim to evaluate the effectiveness of sequential unlearning of copyrighted books. At each time step, we unlearn one book following the experimental design of \cite{zhou2023making, carlini2022quantifying}. For each book, we split the text into chunks of 200 tokens and use the system prompt, instruction prompt, and the first 100 tokens as prompt text to ask the model to continue the story, with the following 100 tokens serving as the ground truth. To assess the amount of copyrighted information being leaked, we compared the LLM's completion with the remaining 100 tokens of each chunk from the original book using techniques for extract training data proposed by \citet{yu2023bag}. Specifically, we use nucleus sampling \cite{holtzman2019curious} by setting the temperature to be $0.4$ and $\eta = 0.6$, which means that the smallest set of the most likely tokens with total probabilities equal to or greater than 0.6 are selected. 

In addition to books in $D_f$, we evaluate performance on previously unlearned books ($D_{prev}$, when $t > 1$) and books that should not be unlearned, denoted as $D_{nor}$ (representing non-targeted knowledge), where $D_{nor} \subset D_r$. 
Details about experiment settings are in Appendix \ref{sec:appendix-experiment_details:experiment_settings}.

\subsection{Evaluation Metrics}
To assess the effectiveness of copyright takedown methods in reducing copyright risks, we assess the transformativeness of the generated outputs in comparison to the original continuations \cite{henderson2023foundation}. 
Following \citet{wei2024evaluating}, we use Rouge-1 and Rouge-L scores \cite{lin2004rouge} to measure the similarities between the model's outputs and the original content. For the books we aim to unlearn, lower Rouge-1 and Rouge-L scores indicate greater   transformativeness, thereby reducing the risk of copyright infringement. In our experiments, we evaluated Rouge-1 and Rouge-L scores on the datasets $D_f$, $D_{prev}$, and $D_{nor}$. \footnote{In alignment with previous copyright evaluation metrics \cite{maini2024tofu, yao2024machine}, and recognizing that semantic similarity alone is insufficient to determine copyright infringement, we include evaluation metrics that focus on lexical similarity (e.g., Rouge) and exclude those that solely reflect semantic similarity.}

In addition to evaluating the model's unlearning effectiveness, we also assessed general-purpose language abilities after copyright takedown methods. The tasks considered include Massive Multitask Language Understanding (MMLU) \cite{hendrycks2020measuring} and MT-Bench \cite{zheng2023judging}. More details are in Appendix \ref{sec:appendix-experiment_details:eval_metrics}.

\begin{figure*}
\centering
         \begin{subfigure}[b]{\textwidth}
            \centering
            \includegraphics[width=0.7\textwidth]{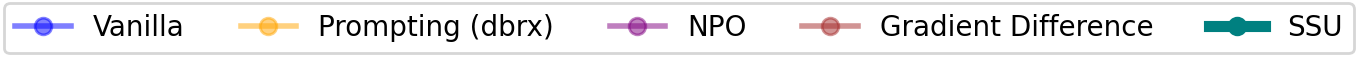}
          \end{subfigure}
        \begin{subfigure}{0.410\textwidth}
        \includegraphics[width=\textwidth]{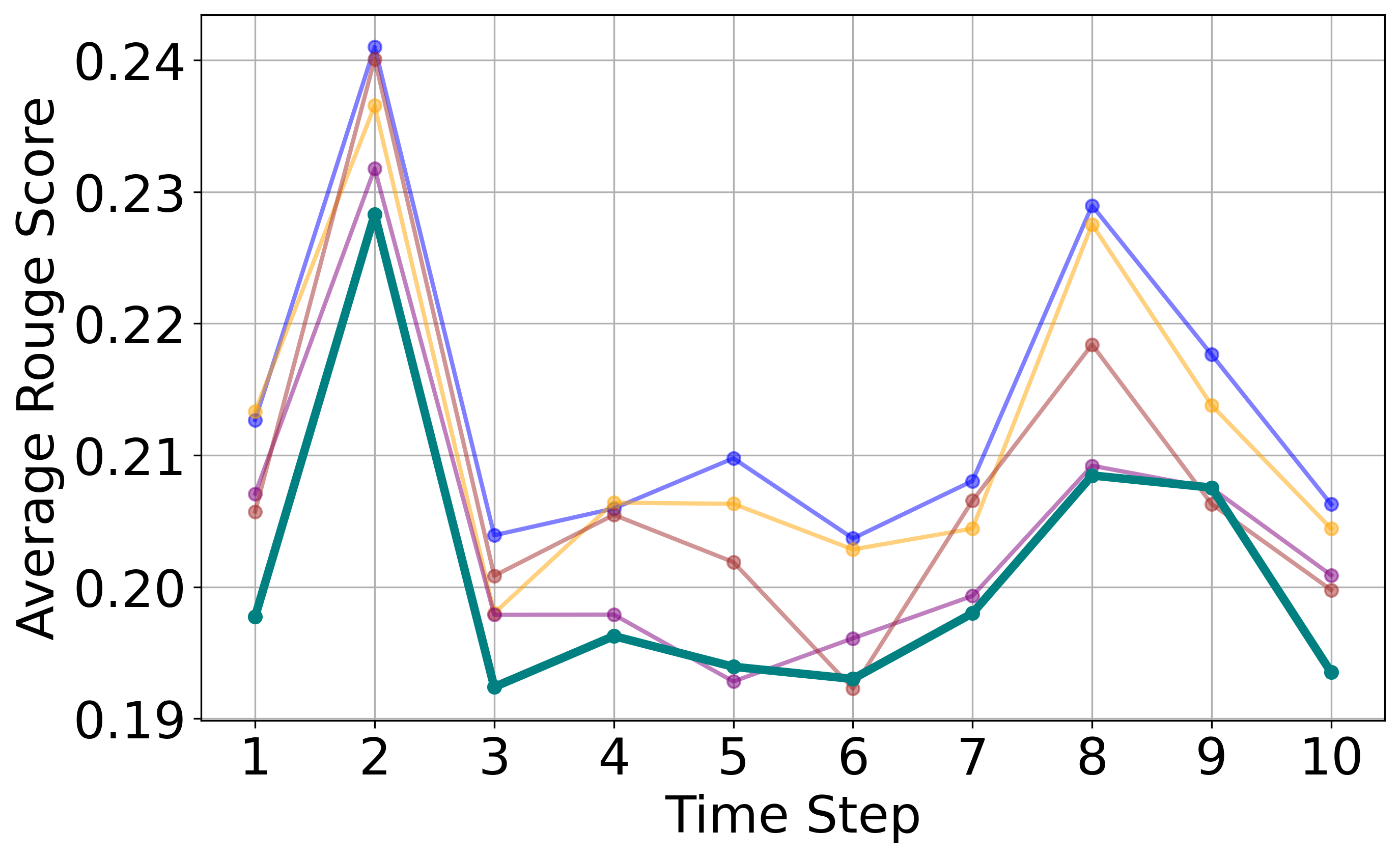}
        \subcaption{Avg Rouge Score on $D_f$}
        \label{fig:main_book_forget_unlearn_main}
        \end{subfigure}
        \begin{subfigure}{0.430\textwidth}
        \includegraphics[width=\textwidth]{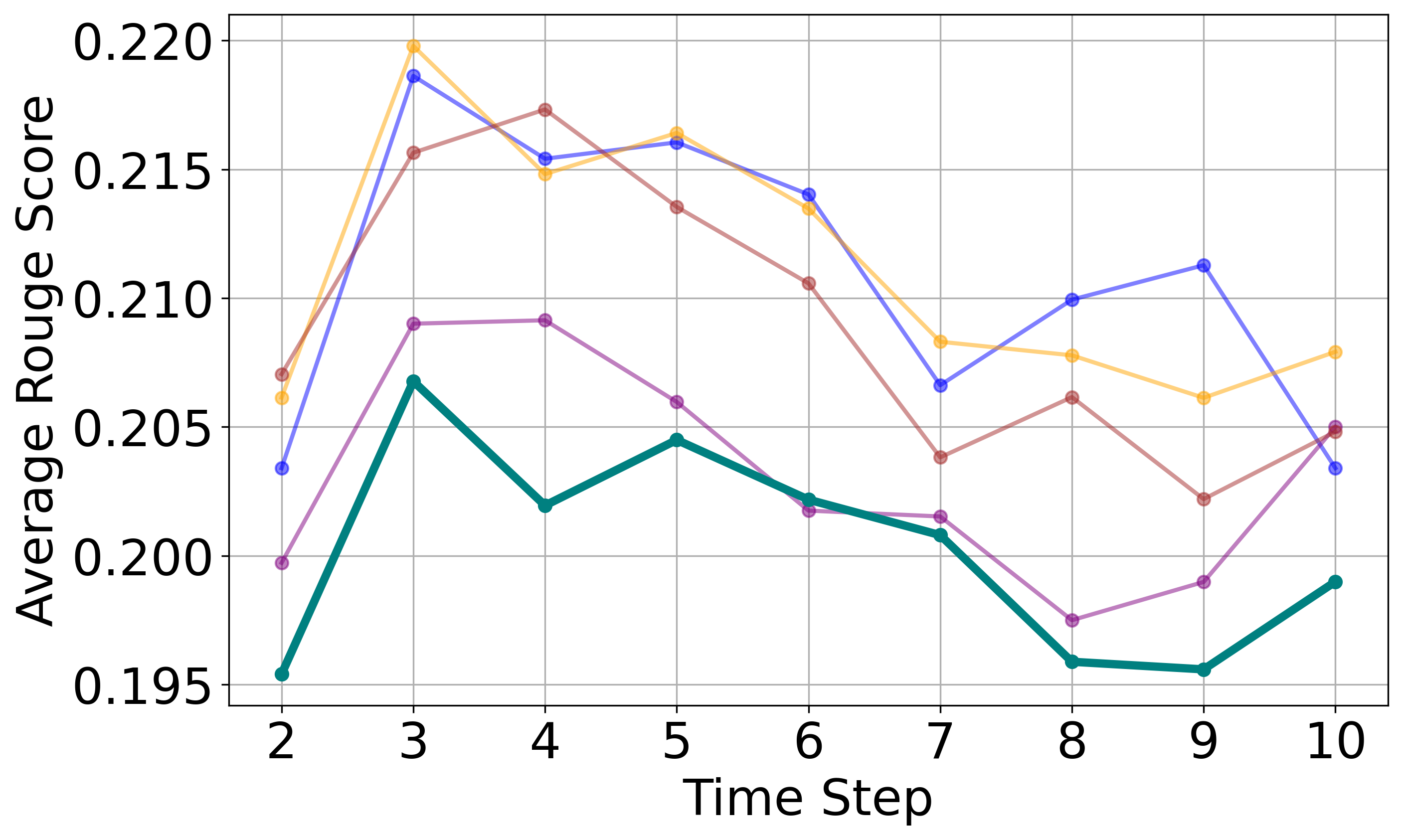}
        \subcaption{Avg Rouge Score on $D_{prev}$}
        \label{fig:main_book_forget_previous_main}
        \end{subfigure} 
        
      \begin{subfigure}{0.430\textwidth}
        \includegraphics[width=\textwidth]{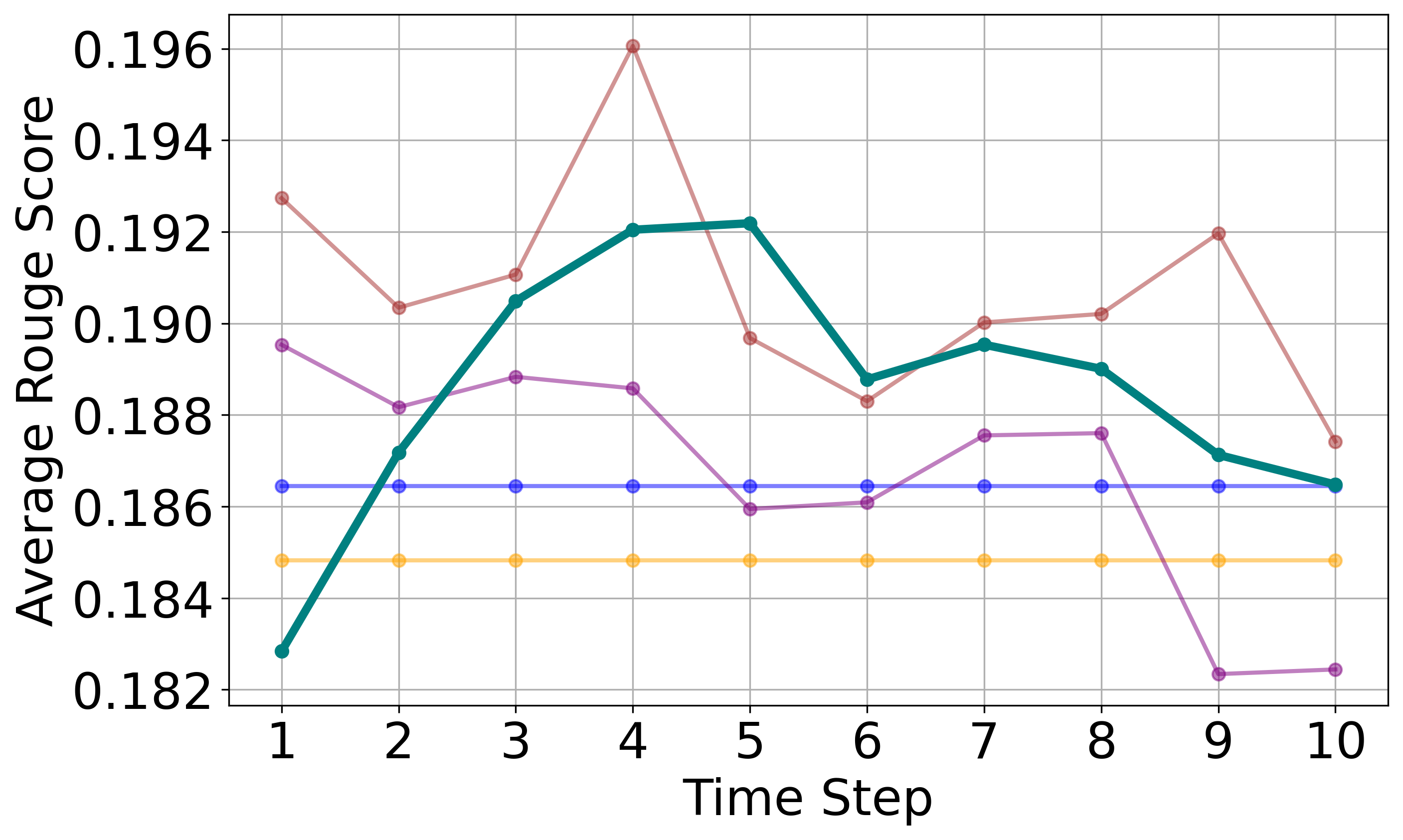}
        \subcaption{Avg Rouge Score on $D_{nor}$}
        \label{fig:main_book_forget_normal_main}
        \end{subfigure}
        \begin{subfigure}{0.410\textwidth}
        \includegraphics[width=\textwidth]{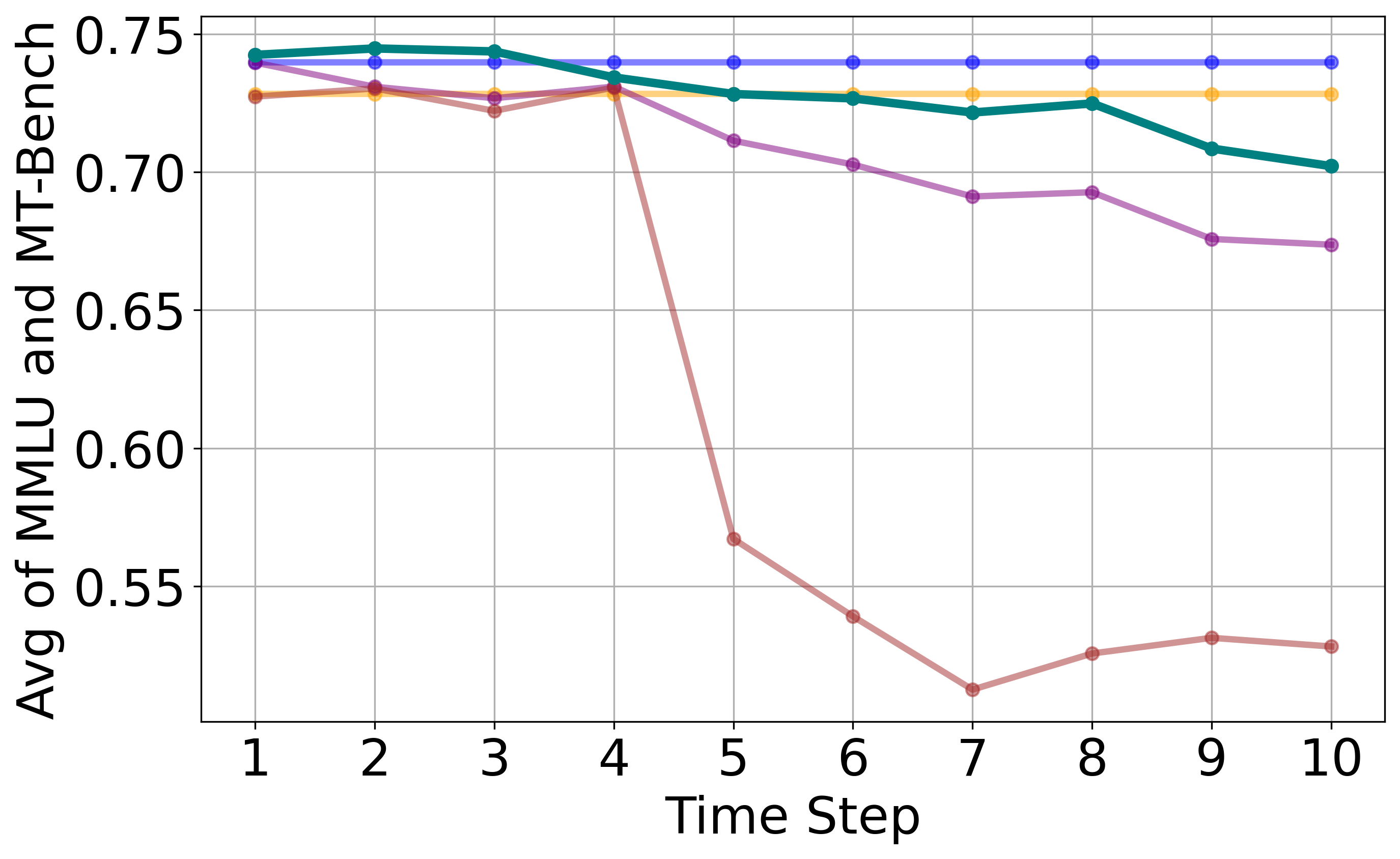}
        \subcaption{Avg MMLU and MT-Bench score}
\label{fig:main_book_forget_general_llama3.1_main}
        \end{subfigure}
    \vspace{-0.1in}

\caption{The averaged Rouge-1 and Rouge-L scores and benchmark scores for Llama3.1, omitting baseline methods that either consistently have low unlearning efficacy or easy to collapse (collapse details in Appendix \ref{sec:appendix-full_experiment_numbers}): (a) books to forget $D_f$ ($\downarrow$); (b) previously unlearned books $D_{prev}$ ($\downarrow$); (c) $D_{nor}$ ($\uparrow$). and (d) averaged normalized MMLU and MT-Bench scores ($\uparrow$). Lower Rouge scores on $D_f$ and $D_{prev}$ indicate better unlearning, while higher scores for $D_{nor}$ and benchmarks reflect better performance. Result with all methods is in Figure \ref{fig:main_book_forget_all_appendix} in Appendix \ref{sec:appendix-full-mistral-llama-results}.}
\vspace{-0.2in}
\label{fig:main_book_forget_all_main}
\end{figure*}

\subsection{Datasets and Models} We use the open-source language models Llama-3.1-8B-Instruct (Llama3.1) \cite{dubey2024llama} and Mistral-7B-Instruct-v0.3 (Mistral-7B) \cite{jiang2023mistral}, both fine-tuned on a dataset of 10 books from Project Gutenberg \footnote{\href{https://www.gutenberg.org/}{gutenberg.org}} ($D_f$) for one epoch as the vanilla models for our experiments.

For Llama-3.1, we unlearned 10 books across 10 time steps, and for Mistral-7B, we unlearned the first six books as many methods collapsed by time step 6 (see Figure \ref{fig:main_book_forget_all_mistral}). We intentionally selected {The Adventures of Sherlock Holmes} by Arthur Conan Doyle, \textit{Pride and Prejudice} by Jane Austen, and \textit{Alice's Adventures in Wonderland} by Lewis Carroll at the first, fifth, and eighth time steps, as these books are listed in Project Gutenberg's "Top 100 EBooks of the Last 30 Days." The remaining books were randomly sampled. All books were pre-processed following the methodology of \citet{gerlach2020standardized}. At $t>1$, we constructed $D_{prev}$ by aggregating all previously unlearned books. Additionally, we selected 200 text chunks from 100 other books in Project Gutenberg, pre-processed as above, to form the $D_{nor}$. Further details are provided in
Appendix \ref{sec:appendix-experiment_details:datasets}.

\subsection{Baseline Methods}
Following \citet{wei2024evaluating}, we compared our approach with several baseline methods:

\paragraph{System prompts} Following \citet{wei2024evaluating}, we use two system prompts to directly instruct the model to refrain from generating copyrighted content. The first, denoted as Prompting (a), adds instructions to not generate copyrighted contents to the default system prompt . The second, denoted as Prompting (dbrx), is used by the DBRX model from Databricks \cite{mosaic2024introducing}. This prompt specifies that the model was not trained on copyrighted books. It explicitly mentions that the model does not provide such content (More details can be found in Appendix \ref{sec:appendix-experiment_details:experiment_settings}). 

\paragraph{MemFree Decoding} The Memfree decoding checks whether the model's next token would create an n-gram found in the Bloom filter \cite{bloom1970space}. If it would, we will choose the next most probable token and do the check again, until the generated n-gram will not be found by the filter. 

\paragraph{Unlearning Methods} Unlearning methods such as Gradient Difference involves using books not in $D_f$ to maintain performance and a random mismatch loss to force random outputs for unlearned data. NPO treats $D_f$ as dispreferred data and formulates the unlearning problem within a preference optimization framework. More details, including hyperparameter choices, are in Appendix \ref{sec:appendix-experiment_details:baseline_methods}.

\section{Results} 
\label{sec:results}
We present experimental results of Llama3.1 for different unlearning time steps in Figure \ref{fig:main_book_forget_all_main} and \ref{fig:main_book_forget_all_appendix}.  See the results for Mistral-7B in Appendix \ref{sec:appendix-full-mistral-llama-results}, and the results with exact numbers in Appendix \ref{sec:appendix-full_experiment_numbers}.

\subsection{Sequential Unlearning of Books}

This section examines the impact of unlearning on \( D_{f} \) and $D_{prev}$. Ideally, the model should have lower average Rouge scores to demonstrate lower risks of copyright infringement. 

\paragraph{\method consistently ranks among the top unlearning methods for mitigating copyright infringement.}
As shown in Figure \ref{fig:main_book_forget_unlearn_main} and \ref{fig:main_book_forget_unlearn}, for Llama3.1,  \method consistently achieves one of the lowest average Rouge scores on $D_f$. Similarly, as shown in Figure \ref{fig:main_book_forget_previous_main} and \ref{fig:main_book_forget_previous}, \method proves to be one of the most effective across all time steps for books in $D_{prev}$, more effective in mitigating copyright infringement than NPO and Gradient Difference. The results are similar for Mistral-7B. As shown in Figures \ref{fig:main_book_forget_unlearn_mistral} and \ref{fig:main_book_forget_previous_mistral}, Gradient Ascent and Gradient Difference achieve the lowest Rouge scores on $D_f$ at later time steps, effectively erasing copyrighted content through opposite gradient updates, ultimately leading to catastrophic collapse. TV also maintains effective unlearning, but also collapses at time step 5. In contrast, NPO’s average Rouge score on $D_f$ remains close to the vanilla model, indicating ineffectiveness in mitigating copyright infringement.  On the other hand, \method demonstrates less risks of copyright infringement.

\paragraph{Prompting and MemFree Decoding offer limited mitigation of copyright infringement.} As shown in Figures \ref{fig:main_book_forget_unlearn} and \ref{fig:main_book_forget_unlearn_mistral}, the Rouge scores for prompting and MemFree are largely indistinguishable from the vanilla model across many time steps and models. Although system prompts attempt to prevent generating copyrighted content, Llama3.1 often produces higher Rouge scores with prompting (a), and prompting (dbrx) is only marginally effective at certain time steps. We suspect this is due to the instruction-tuned Llama3.1 model’s inability to differentiate what constitutes copyrighted content. For the Mistral-7B model, both prompting methods are similarly ineffective in reducing infringement risks.  Lastly, the MemFree decoding is always ineffective in unlearning copyrighted books for for Llama3.1 and Mistral-7B.

\begin{figure*}
\centering
         \begin{subfigure}[b]{\textwidth}
            \centering
            \includegraphics[width=0.8\textwidth]{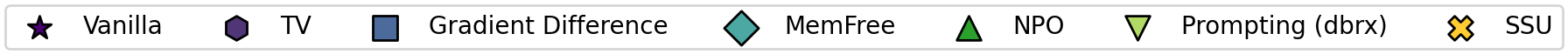}
          \end{subfigure}
        \begin{subfigure}{0.435\textwidth}
        \includegraphics[width=\textwidth]{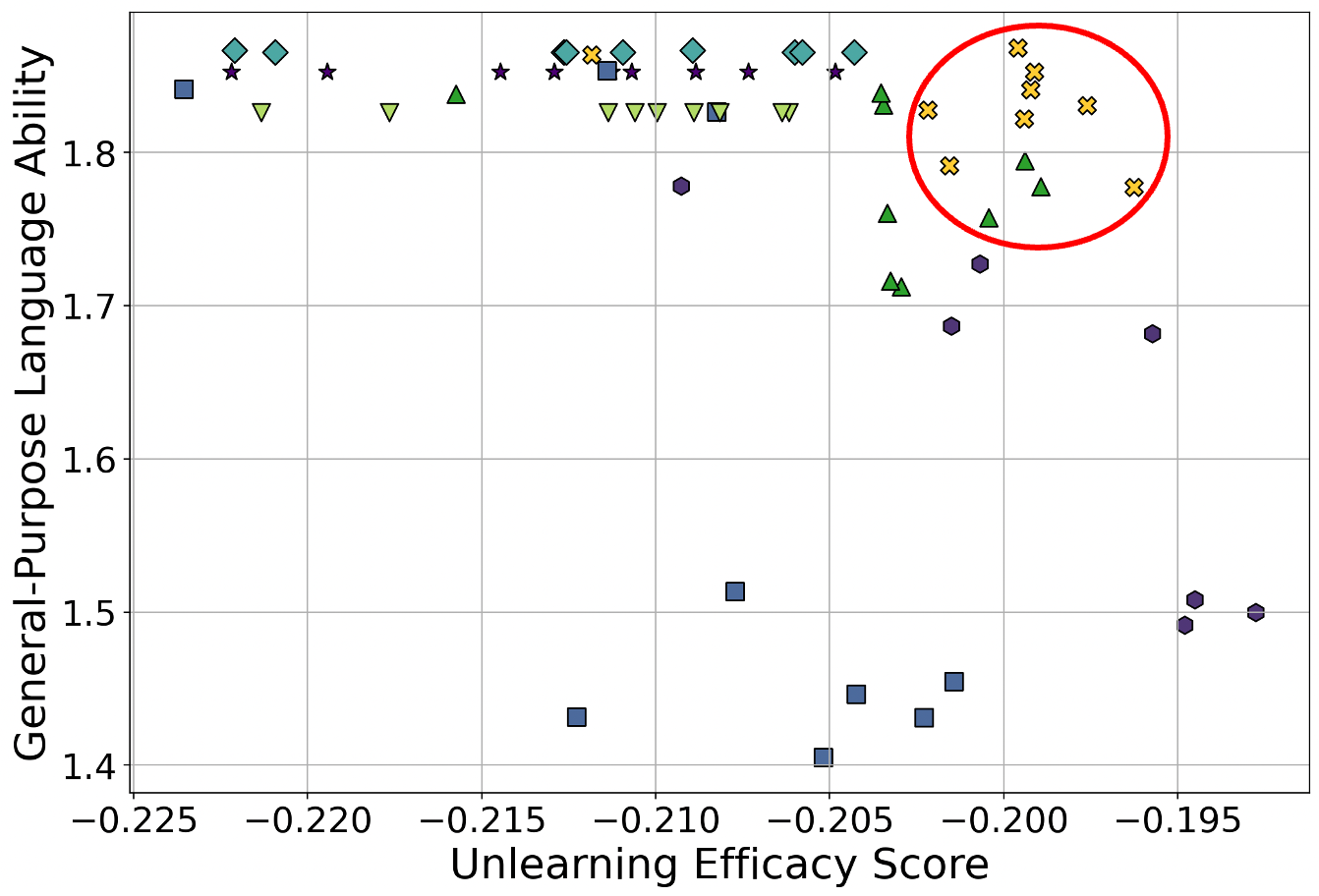}
        \subcaption{Llama3.1-8B-Instruct Trade-off}
        \label{fig:trade_off_llama3.1_main}
        \end{subfigure}
        \begin{subfigure}{0.45\textwidth}
        \includegraphics[width=\textwidth]{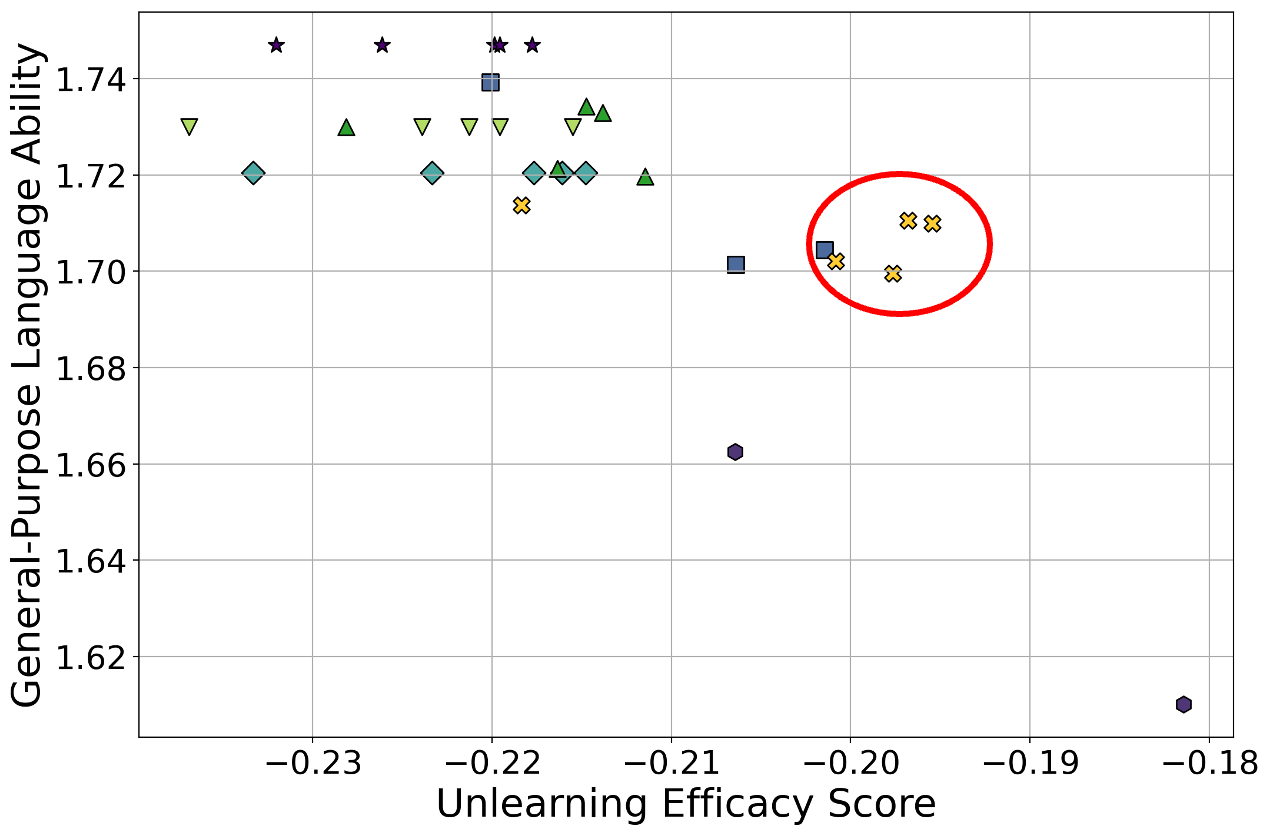}
        \subcaption{Mistral-7B-Instruct Trade-off}
        \label{fig:trade_off_mistral_main}
        \end{subfigure} 
\vspace{-0.1in}
\caption{
Trade-off analysis between general-purpose language abilities and unlearning efficacy for Llama3.1 and Mistral-7B, considering only methods at time steps greater than 1. For improved visualization, we exclude Prompting (a) and GA, which consistently exhibit low unlearning efficacy or collapse during the process. We also exclude TV beyond time step 9 (Llama3.1) and time step 3 (Mistral-7B), as well as Gradient Difference beyond time step 4 (Mistral-7B), due to collapose at these stage (see Appendix \ref{sec:appendix-full_experiment_numbers} for collapse details). General-purpose abilities are calculated using normalized averages of MMLU and MT-Bench scores, while unlearning efficacy is represented by the average of Rouge-1 and Rouge-L scores on $D_f$ (targeted data) and $D_{prev}$ (previously unlearned data). Lower Rouge scores indicate better unlearning performance; hence, these values are negated for clarity. The ideal method balances both metrics and is positioned in the top-right corner.  Full result with all methods is in Figure \ref{fig:trade_off_appendix}.}
\vspace{-0.2in}
\label{fig:trade_off_main}
\end{figure*}

\subsection{Non-Targeted Knowledge Retention}
\label{sec:results_knowledge_retention}

This section examines the impact of unlearning on \( D_{nor} \), the books not intended to be unlearned. Ideally, the model should maintain the average Rouge scores compared to the vanilla model.

\paragraph{\method maximally preserves non-targeted knowledge compared to other unlearning methods.} The results for Llama3.1 and Mistral-7B are notably different. For Mistral-7B, as shown in Figure \ref{fig:main_book_forget_normal_mistral}, the average Rouge scores for GA, Gradient Difference, and TV demonstrate significant loss of retained knowledge at later time steps. While NPO is less effective at mitigating copyright infringement, it consistently retains more knowledge of non-targeted books than \method. In contrast, the results for Llama3.1 fluctuate. As shown in Figures \ref{fig:main_book_forget_normal_main} and \ref{fig:main_book_forget_normal}, Gradient Difference and \method retain more knowledge of $D_{nor}$ than the vanilla model. The unexpected re-emergence of these knowledge after unlearning is possibly due to knowledge redistribution during the unlearning process. \citet{yang2024reawakening} also examined this anticipatory recovery behavior where LLMs recover knowledge from the forgetting on documents before encountering them again. Further research is needed to explore this phenomenon. Nevertheless, \method demonstrates stability by preserving knowledge in $D_{nor}$. 

\paragraph{Prompting and MemFree decoding maintain non-targeted knowledge retention.} As shown in Figures \ref{fig:main_book_forget_normal} and \ref{fig:main_book_forget_normal_mistral},
prompting and MemFree consistently retain non-targeted knowledge. Notably, for Llama3.1, both prompting (a) and MemFree retain more knowledge than the vanilla model. In contrast, for Mistral-7B, both prompting methods and MemFree exhibit slightly lower levels of knowledge retention compared to the vanilla model.

\begin{figure*}[th]
\centering
         \begin{subfigure}[b]{\textwidth}
            \centering
            \includegraphics[width=0.45\textwidth]{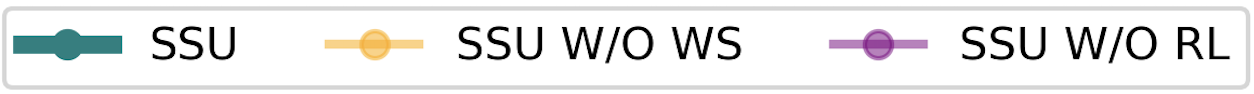}
          \end{subfigure}
        \begin{subfigure}{0.24\textwidth}
        \includegraphics[width=\textwidth]{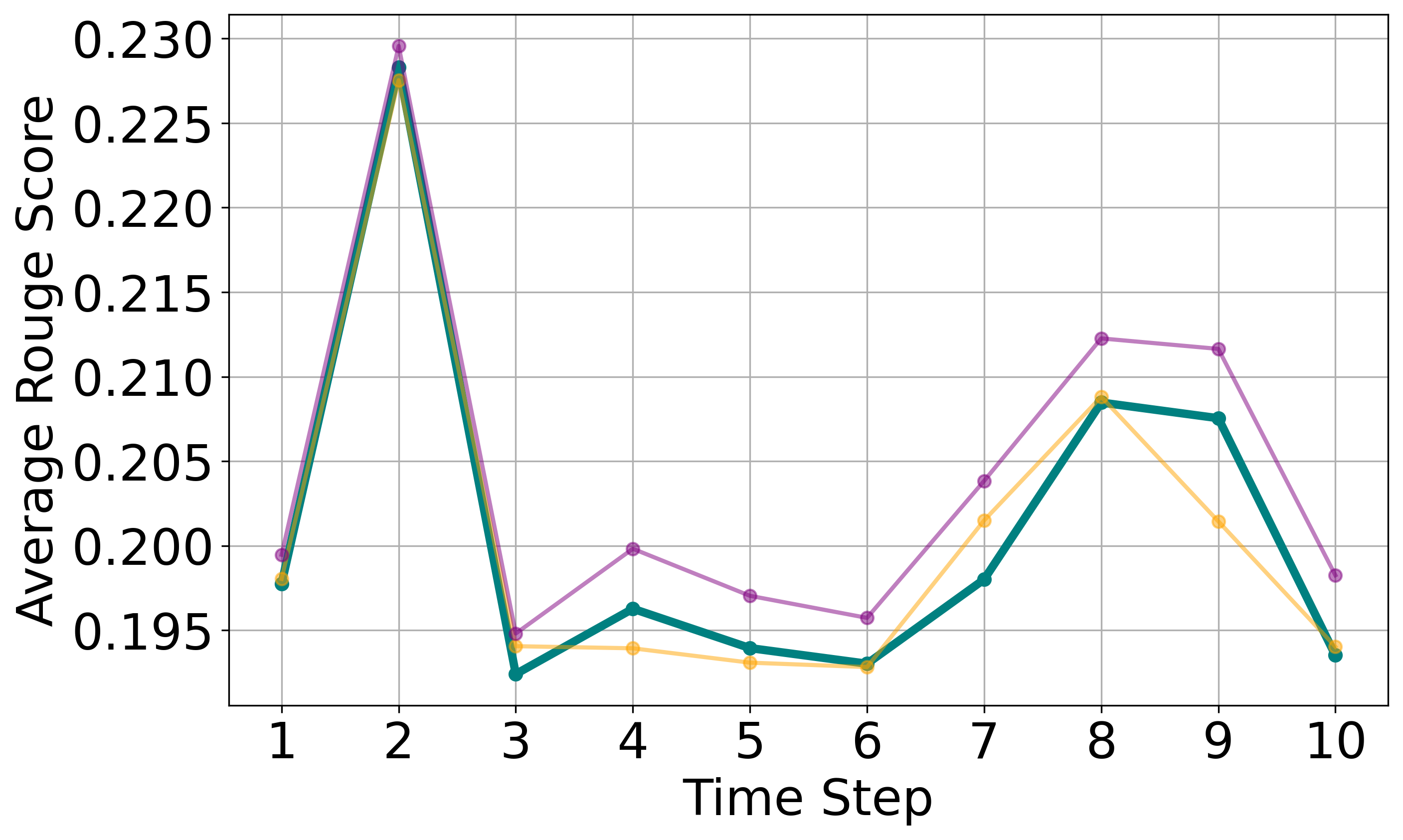}
        \subcaption{Avg Rouge on $D_f$}
        \label{fig:ablation_book_forget_llama3.1}
        \end{subfigure}
        \begin{subfigure}{0.24\textwidth}
        \includegraphics[width=\textwidth]{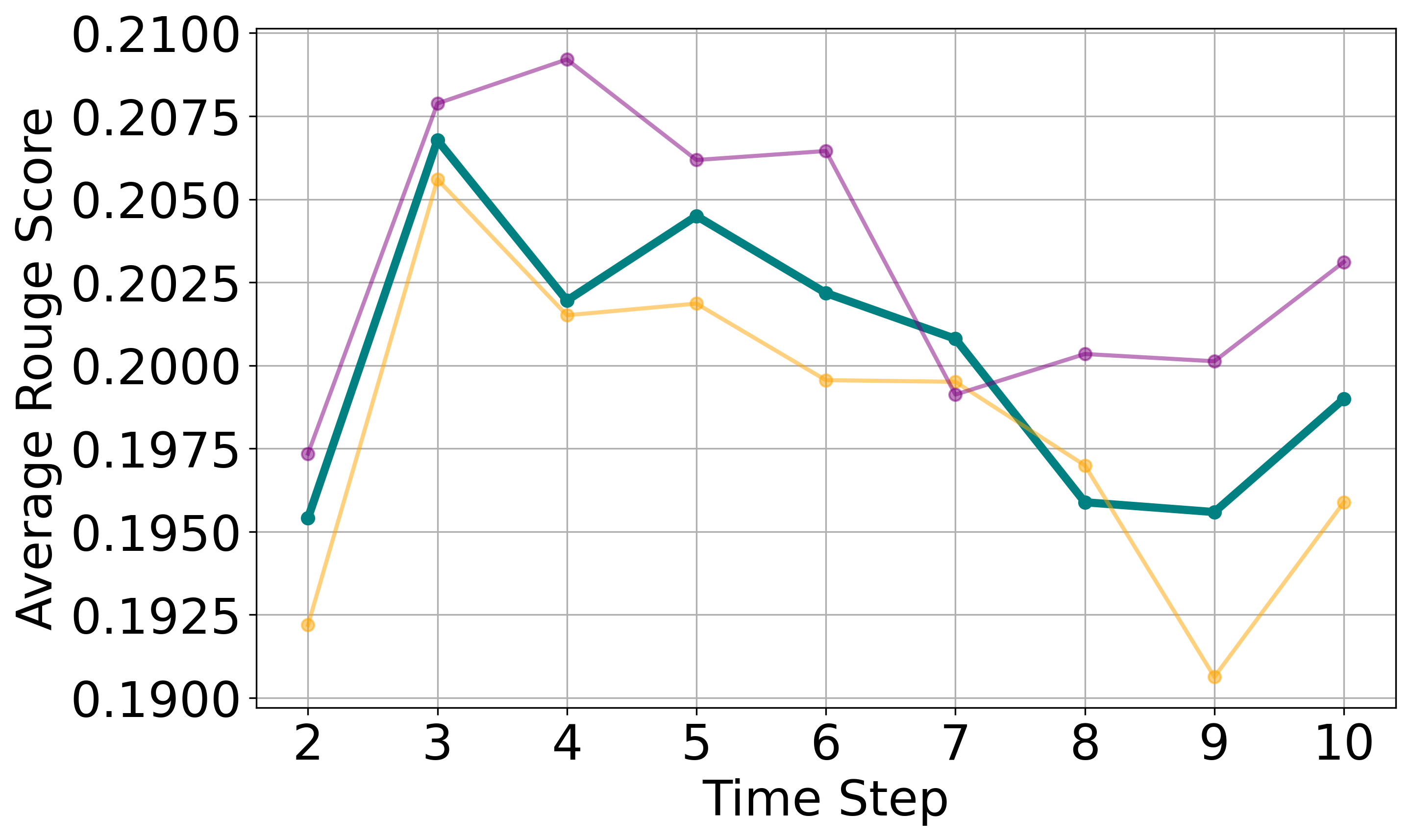}
        \subcaption{Avg Rouge Score on $D_{prev}$}
        \label{fig:ablation_book_prev_llama3.1}
        \end{subfigure}
        \begin{subfigure}{0.24\textwidth}
        \includegraphics[width=\textwidth]{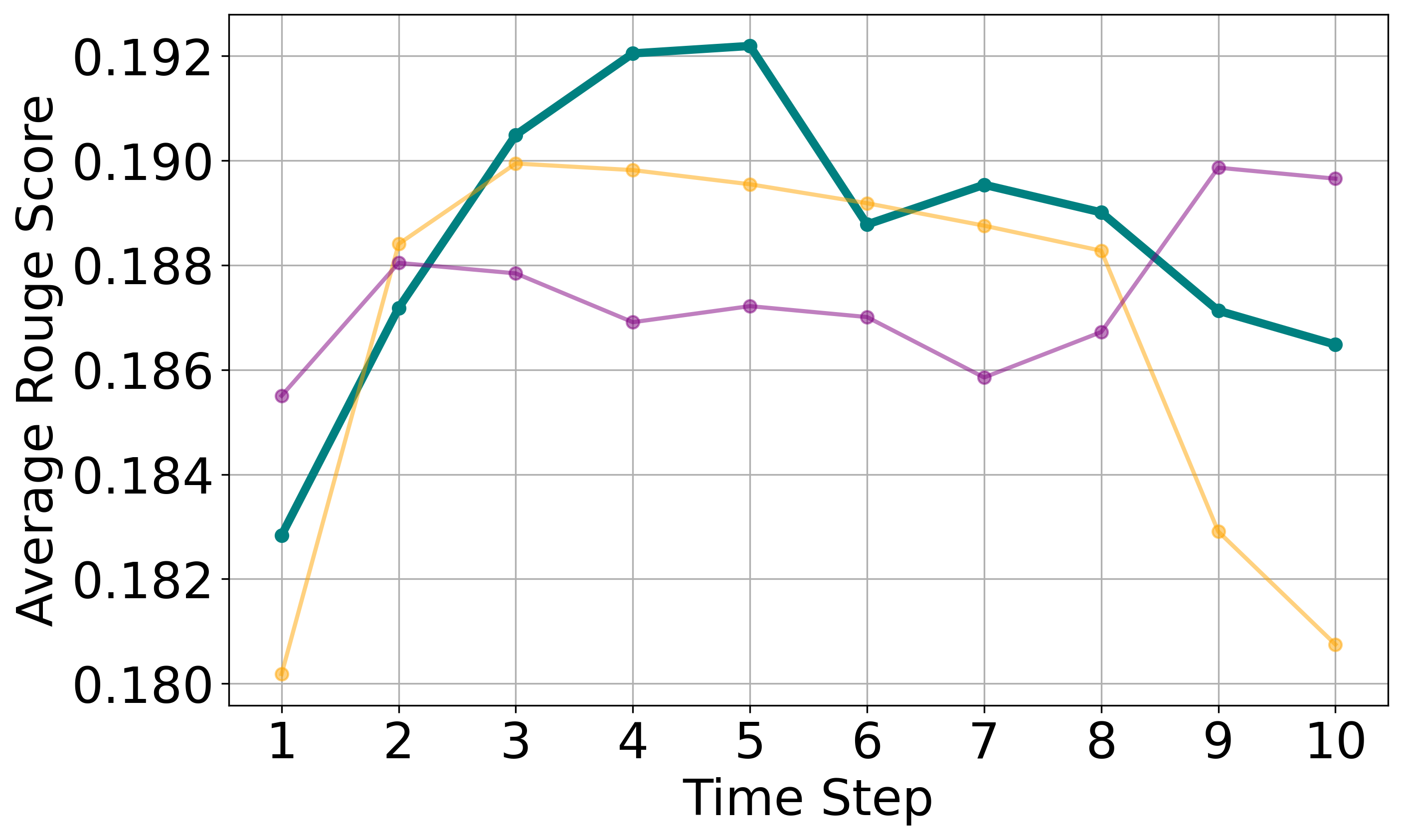}
        \subcaption{Avg Rouge Score on $D_n$}
        \label{fig:ablation_book_norm_llama3.1)}
        \end{subfigure}
        \begin{subfigure}{0.24\textwidth}
        \includegraphics[width=\textwidth]{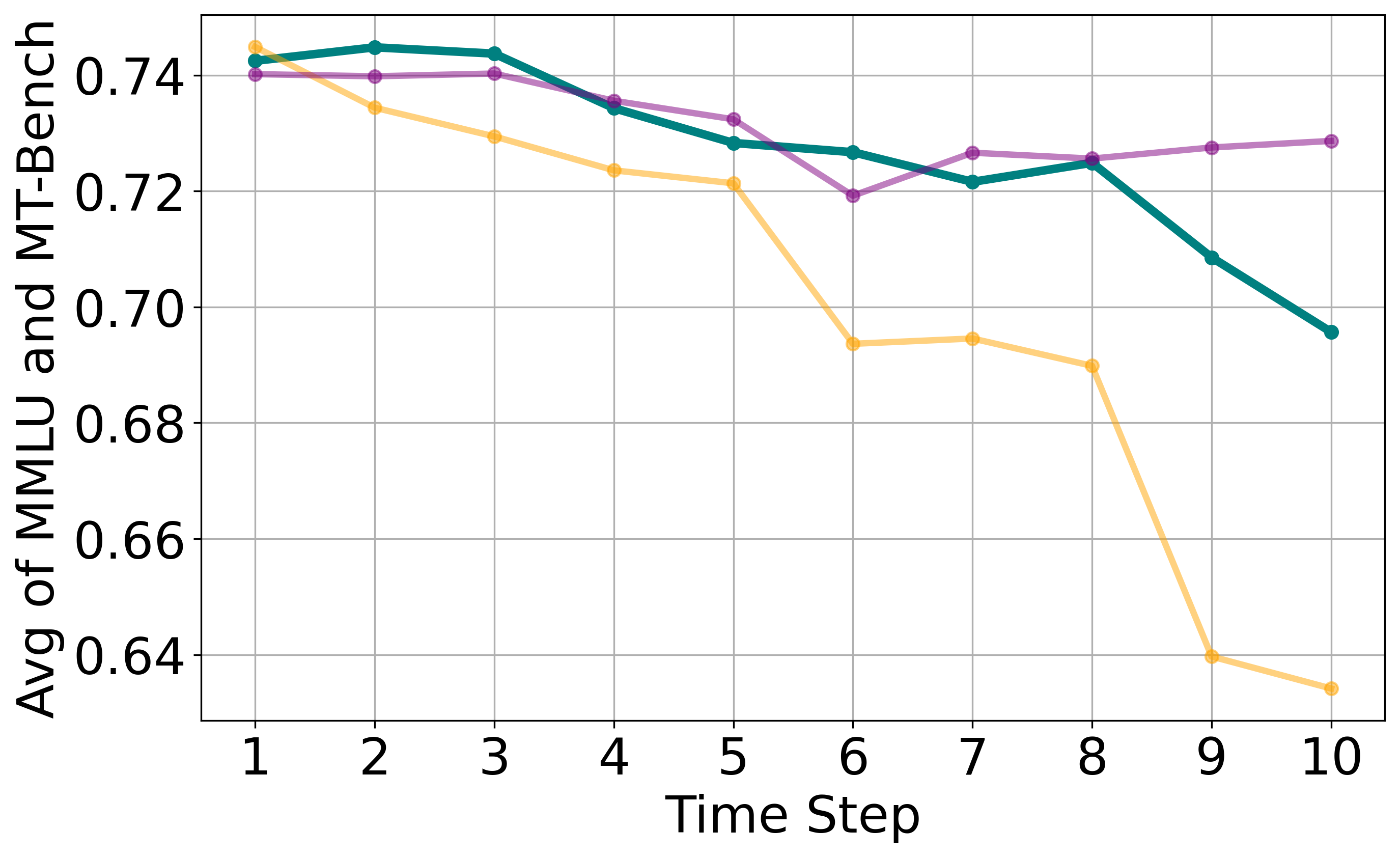}
        \subcaption{General Performance}
        \label{fig:ablation_general_llama3.1)}
        \end{subfigure}

\caption{Ablation study of \method for Llama3.1-8B-Instruct. The orange line represents unlearning without the weight saliency map, while the purple line shows the effect of removing the random labeling loss.}
\label{fig:ablation_llama}
\vspace{-0.2in}
\end{figure*}

\subsection{General-purpose Language Abilities}

\paragraph{GA, Gradient Difference, and TV experienced catastrophic collapse at later time steps.} Figures \ref{fig:main_book_forget_general_llama3.1_main}, \ref{fig:main_book_forget_general_llama3.1}, and \ref{fig:main_book_forget_general_mistral}, show that GA, Gradient Difference, and TV both have sudden significant degradation of general-purpose abilities at certain time steps. NPO consistently underperforms \method for Llama3.1 outperforms for Mistral-7B. Prompting and MemFree maintain stable general-purpose language abilities, close to the vanilla model. 

\subsection{Copyright Takedown Trade-offs}

We present the overall trade-off of each method in Figures \ref{fig:trade_off_llama3.1_main} and \ref{fig:trade_off_mistral_main}. The $x$-axis represents unlearning efficacy score, and the $y$-axis represents general-purpose language ability. An ideal copyright takedown method would aim for the top-right corner. Comparing to existing baseline methods across two different models, \textbf{\method achieves a better trade-off between unlearning efficacy and general-purpose language abilities retention.} 

\section{Analysis} 
We analyze the impact of different components of \method, including weight saliency maps and random labeling loss, on the sequential unlearning process. Figure \ref{fig:ablation_llama} presents results for Llama3.1-8B-Instruct, and Figure \ref{fig:ablation_mistral} shows ablation results for Mistral-7B-Instruct-v0.3. We further present the effect of varying $\epsilon_1$ and $\epsilon_2$ in equation \ref{equation:ssu_main_no_weight_saliency} in appendix \ref{sec:appendix-epsilon_results}. 

\subsection{Impact of Weight Saliency}
As shown in Figures  \ref{fig:ablation_book_forget_llama3.1} and \ref{fig:ablation_book_prev_llama3.1}, the performance of \method without weight saliency leads to lower average Rouge scores on both $D_f$ and $D_{nor}$. Additionally, Figure \ref{fig:ablation_general_llama3.1)} shows a sharper decline in benchmark performance with each time step, indicating that without weight saliency, the risk of catastrophic collapse increases as general-purpose language abilities deteriorate. \textbf{By updating only specific parts of the model's weights, weight saliency helps preserve the general-purpose language abilities.}

\subsection{Impact of Random Labeling Loss}
As seen in Figures \ref{fig:ablation_book_forget_llama3.1} and \ref{fig:ablation_book_prev_llama3.1}, \method without random labeling loss results in a higher average Rouge scores and slightly improved general-purpose language abilities. This suggests that \textbf{random labeling loss enhances the model’s ability to unlearn $D_f$ consistently across all time steps.}

It is also noteworthy that, as shown in Figures \ref{fig:ablation_book_norm_llama3.1)} and \ref{fig:ablation_book_norm_mistral)}, \method without weight saliency and \method without random loss lead to lower average Rouge scores for $D_{nor}$ in both models. This observation highlights the need for further research on the impact of unlearning algorithms on preserving content not intended for unlearning.

\section{Conclusion}
We explore sequential unlearning of copyrighted content in LLMs to mitigate copyright infringement. We propose \method, which utilizes random labeling loss and gradient-based weight saliency to achieve more stable sequential unlearning. Experiments show \method outperforms existing methods in balancing unlearning efficacy and language retention. Future research could focus on enhancing the robustness of unlearning algorithms and exploring other copyright takedown methods to further mitigate the risks of copyright infringement.


\section*{Limitations}
\paragraph{Robustness of Machine Unlearning}
Our evaluation relies on lexical-based metrics, which, as noted by \citet{ippolito2023preventing}, may provide a false sense of privacy. Additionally, \citet{lucki2024adversarial} and \citet{cooper2024machine} highlight that existing unlearning methods often lack robustness and may merely obscure data rather than true unlearning. 

In our experiments, we observe that knowledge from previously unlearned books can re-emerge at certain future time steps, a phenomenon present across all unlearning algorithms tested (see Appendix \ref{sec:appendix-sequential_unlearning_challenge}). This further underscores the challenges of sequential unlearning and the necessity for more robust unlearning algorithms.

\paragraph{Unintended Knowledge and Capability Loss}
While \method achieves a better trade-off between unlearning efficacy and general-purpose language retention, it still weakens the model’s reasoning and conversational abilities. Future work should further reduce this gap, ensuring better preservation of general knowledge during sequential unlearning.

\paragraph{Unlearning Is Not All You Need}
Mitigating copyright infringement risks requires more than unlearning alone. For example, \citet{liu2024infini} introduces Infini-gram, an efficient search engine for scanning large-scale LLM training corpora. A complementary approach could involve using such tools to detect copyrighted passages in generated outputs and intervene to ensure non-copyrighted regeneration. This generation-time copyright takedown method represents a promising research direction. Future work could expand search engine tools to cover a broader range of training datasets used in state-of-the-art LLMs. Additionally, regularly updating model checkpoints to align with \textit{the right to be forgotten} would be crucial for maintaining legal and ethical compliance.

\section*{Acknowledgements}
We thank Bryan Li, Yikai Mao, and anonymous reviewers for their valuable feedback and insightful discussions on this paper.

This research is supported in part by the Office of the Director of National Intelligence (ODNI), the Intelligence Advanced Research Projects Activity (IARPA) via the BENGAL Program, the Defense Advanced Research Projects Activity (DARPA) via the SciFy Program, and Schmidt Sciences via the AI2050 fellowship. The views and conclusions contained herein are those of the authors and should not be interpreted as necessarily representing the official policies, either expressed or implied, of ODNI, IARPA, DARPA, Schmidt Sciences, or the U.S. Government. The U.S. Government is authorized to reproduce and distribute reprints for governmental purposes notwithstanding any copyright annotation therein.

\newpage
\bibliography{ref}

\begin{thebibliography}{62}
\providecommand{\natexlab}[1]{#1}

\bibitem[{Bloom(1970)}]{bloom1970space}
Burton~H Bloom. 1970.
\newblock Space/time trade-offs in hash coding with allowable errors.
\newblock \emph{Communications of the ACM}, 13(7):422--426.

\bibitem[{Brittain(2023)}]{brittain2023us}
Blake Brittain. 2023.
\newblock Us copyright office says some ai-assisted works may be copyrighted.
\newblock \emph{Reuters. March}, 15.

\bibitem[{Brown et~al.(2020)Brown, Mann, Ryder, Subbiah, Kaplan, Dhariwal, Neelakantan, Shyam, Sastry, Askell et~al.}]{brown2020language}
Tom Brown, Benjamin Mann, Nick Ryder, Melanie Subbiah, Jared~D Kaplan, Prafulla Dhariwal, Arvind Neelakantan, Pranav Shyam, Girish Sastry, Amanda Askell, et~al. 2020.
\newblock Language models are few-shot learners.
\newblock \emph{Advances in neural information processing systems}, 33:1877--1901.

\bibitem[{Cao and Yang(2015)}]{cao2015towards}
Yinzhi Cao and Junfeng Yang. 2015.
\newblock Towards making systems forget with machine unlearning.
\newblock In \emph{2015 IEEE symposium on security and privacy}, pages 463--480. IEEE.

\bibitem[{Carlini et~al.(2022)Carlini, Ippolito, Jagielski, Lee, Tramer, and Zhang}]{carlini2022quantifying}
Nicholas Carlini, Daphne Ippolito, Matthew Jagielski, Katherine Lee, Florian Tramer, and Chiyuan Zhang. 2022.
\newblock Quantifying memorization across neural language models.
\newblock \emph{arXiv preprint arXiv:2202.07646}.

\bibitem[{Carlini et~al.(2021)Carlini, Tramer, Wallace, Jagielski, Herbert-Voss, Lee, Roberts, Brown, Song, Erlingsson et~al.}]{carlini2021extracting}
Nicholas Carlini, Florian Tramer, Eric Wallace, Matthew Jagielski, Ariel Herbert-Voss, Katherine Lee, Adam Roberts, Tom Brown, Dawn Song, Ulfar Erlingsson, et~al. 2021.
\newblock Extracting training data from large language models.
\newblock In \emph{30th USENIX Security Symposium (USENIX Security 21)}, pages 2633--2650.

\bibitem[{Chen and Yang(2023)}]{chen2023unlearn}
Jiaao Chen and Diyi Yang. 2023.
\newblock Unlearn what you want to forget: Efficient unlearning for llms.
\newblock \emph{arXiv preprint arXiv:2310.20150}.

\bibitem[{Chowdhery et~al.(2023)Chowdhery, Narang, Devlin, Bosma, Mishra, Roberts, Barham, Chung, Sutton, Gehrmann et~al.}]{chowdhery2023palm}
Aakanksha Chowdhery, Sharan Narang, Jacob Devlin, Maarten Bosma, Gaurav Mishra, Adam Roberts, Paul Barham, Hyung~Won Chung, Charles Sutton, Sebastian Gehrmann, et~al. 2023.
\newblock Palm: Scaling language modeling with pathways.
\newblock \emph{Journal of Machine Learning Research}, 24(240):1--113.

\bibitem[{Chu et~al.(2024)Chu, Song, and Yang}]{chu2024protect}
Timothy Chu, Zhao Song, and Chiwun Yang. 2024.
\newblock How to protect copyright data in optimization of large language models?
\newblock In \emph{Proceedings of the AAAI Conference on Artificial Intelligence}, volume~38, pages 17871--17879.

\bibitem[{Cooper et~al.(2024)Cooper, Choquette-Choo, Bogen, Jagielski, Filippova, Liu, Chouldechova, Hayes, Huang, Mireshghallah et~al.}]{cooper2024machine}
A~Feder Cooper, Christopher~A Choquette-Choo, Miranda Bogen, Matthew Jagielski, Katja Filippova, Ken~Ziyu Liu, Alexandra Chouldechova, Jamie Hayes, Yangsibo Huang, Niloofar Mireshghallah, et~al. 2024.
\newblock Machine unlearning doesn't do what you think: Lessons for generative ai policy, research, and practice.
\newblock \emph{arXiv preprint arXiv:2412.06966}.

\bibitem[{Dong et~al.(2024)Dong, Lin, Belkin, Huerta, and Vuli{\'c}}]{dong2024unmemorization}
Yijiang~River Dong, Hongzhou Lin, Mikhail Belkin, Ramon Huerta, and Ivan Vuli{\'c}. 2024.
\newblock Unmemorization in large language models via self-distillation and deliberate imagination.
\newblock \emph{arXiv preprint arXiv:2402.10052}.

\bibitem[{Duarte et~al.(2024)Duarte, Zhao, Oliveira, and Li}]{duarte2024cop}
Andr{\'e}~V Duarte, Xuandong Zhao, Arlindo~L Oliveira, and Lei Li. 2024.
\newblock De-cop: Detecting copyrighted content in language models training data.
\newblock \emph{arXiv preprint arXiv:2402.09910}.

\bibitem[{Dubey et~al.(2024)Dubey, Jauhri, Pandey, Kadian, Al-Dahle, Letman, Mathur, Schelten, Yang, Fan et~al.}]{dubey2024llama}
Abhimanyu Dubey, Abhinav Jauhri, Abhinav Pandey, Abhishek Kadian, Ahmad Al-Dahle, Aiesha Letman, Akhil Mathur, Alan Schelten, Amy Yang, Angela Fan, et~al. 2024.
\newblock The llama 3 herd of models.
\newblock \emph{arXiv preprint arXiv:2407.21783}.

\bibitem[{Eldan and Russinovich(2023)}]{eldan2023s}
Ronen Eldan and Mark Russinovich. 2023.
\newblock Who's harry potter? approximate unlearning in llms.
\newblock \emph{arXiv preprint arXiv:2310.02238}.

\bibitem[{Fan et~al.(2023)Fan, Liu, Zhang, Wei, Wong, and Liu}]{fan2023salun}
Chongyu Fan, Jiancheng Liu, Yihua Zhang, Dennis Wei, Eric Wong, and Sijia Liu. 2023.
\newblock Salun: Empowering machine unlearning via gradient-based weight saliency in both image classification and generation.
\newblock \emph{arXiv preprint arXiv:2310.12508}.

\bibitem[{Gerlach and Font-Clos(2020)}]{gerlach2020standardized}
Martin Gerlach and Francesc Font-Clos. 2020.
\newblock A standardized project gutenberg corpus for statistical analysis of natural language and quantitative linguistics.
\newblock \emph{Entropy}, 22(1):126.

\bibitem[{Gupta and Anumanchipalli(2024)}]{gupta2024rebuilding}
Akshat Gupta and Gopala Anumanchipalli. 2024.
\newblock Rebuilding rome: Resolving model collapse during sequential model editing.
\newblock \emph{arXiv preprint arXiv:2403.07175}.

\bibitem[{Gupta et~al.(2024{\natexlab{a}})Gupta, Rao, and Anumanchipalli}]{gupta2024model}
Akshat Gupta, Anurag Rao, and Gopala Anumanchipalli. 2024{\natexlab{a}}.
\newblock Model editing at scale leads to gradual and catastrophic forgetting.
\newblock \emph{arXiv preprint arXiv:2401.07453}.

\bibitem[{Gupta et~al.(2024{\natexlab{b}})Gupta, Sajnani, and Anumanchipalli}]{gupta2024unified}
Akshat Gupta, Dev Sajnani, and Gopala Anumanchipalli. 2024{\natexlab{b}}.
\newblock A unified framework for model editing.
\newblock \emph{arXiv preprint arXiv:2403.14236}.

\bibitem[{Henderson et~al.(2023)Henderson, Li, Jurafsky, Hashimoto, Lemley, and Liang}]{henderson2023foundation}
Peter Henderson, Xuechen Li, Dan Jurafsky, Tatsunori Hashimoto, Mark~A Lemley, and Percy Liang. 2023.
\newblock Foundation models and fair use.
\newblock \emph{Journal of Machine Learning Research}, 24(400):1--79.

\bibitem[{Hendrycks et~al.(2020)Hendrycks, Burns, Basart, Zou, Mazeika, Song, and Steinhardt}]{hendrycks2020measuring}
Dan Hendrycks, Collin Burns, Steven Basart, Andy Zou, Mantas Mazeika, Dawn Song, and Jacob Steinhardt. 2020.
\newblock Measuring massive multitask language understanding.
\newblock \emph{arXiv preprint arXiv:2009.03300}.

\bibitem[{Hewitt et~al.(2024)Hewitt, Chen, Xie, Adams, Liang, and Manning}]{hewitt2024model}
John Hewitt, Sarah Chen, Lanruo~Lora Xie, Edward Adams, Percy Liang, and Christopher~D Manning. 2024.
\newblock Model editing with canonical examples.
\newblock \emph{arXiv preprint arXiv:2402.06155}.

\bibitem[{Holtzman et~al.(2019)Holtzman, Buys, Du, Forbes, and Choi}]{holtzman2019curious}
Ari Holtzman, Jan Buys, Li~Du, Maxwell Forbes, and Yejin Choi. 2019.
\newblock The curious case of neural text degeneration.
\newblock \emph{arXiv preprint arXiv:1904.09751}.

\bibitem[{Hu et~al.(2021)Hu, Shen, Wallis, Allen-Zhu, Li, Wang, Wang, and Chen}]{hu2021lora}
Edward~J Hu, Yelong Shen, Phillip Wallis, Zeyuan Allen-Zhu, Yuanzhi Li, Shean Wang, Lu~Wang, and Weizhu Chen. 2021.
\newblock Lora: Low-rank adaptation of large language models.
\newblock \emph{arXiv preprint arXiv:2106.09685}.

\bibitem[{Huang et~al.(2024)Huang, Yang, and Potts}]{huang2024demystifying}
Jing Huang, Diyi Yang, and Christopher Potts. 2024.
\newblock Demystifying verbatim memorization in large language models.
\newblock \emph{arXiv preprint arXiv:2407.17817}.

\bibitem[{Ilharco et~al.(2022)Ilharco, Ribeiro, Wortsman, Gururangan, Schmidt, Hajishirzi, and Farhadi}]{ilharco2022editing}
Gabriel Ilharco, Marco~Tulio Ribeiro, Mitchell Wortsman, Suchin Gururangan, Ludwig Schmidt, Hannaneh Hajishirzi, and Ali Farhadi. 2022.
\newblock Editing models with task arithmetic.
\newblock \emph{arXiv preprint arXiv:2212.04089}.

\bibitem[{Ippolito et~al.(2023)Ippolito, Tramer, Nasr, Zhang, Jagielski, Lee, Choo, and Carlini}]{ippolito2023preventing}
Daphne Ippolito, Florian Tramer, Milad Nasr, Chiyuan Zhang, Matthew Jagielski, Katherine Lee, Christopher~Choquette Choo, and Nicholas Carlini. 2023.
\newblock Preventing generation of verbatim memorization in language models gives a false sense of privacy.
\newblock In \emph{Proceedings of the 16th International Natural Language Generation Conference}, pages 28--53.

\bibitem[{Ippolito et~al.(2022)Ippolito, Tram{\`e}r, Nasr, Zhang, Jagielski, Lee, Choquette-Choo, and Carlini}]{ippolito2022preventing}
Daphne Ippolito, Florian Tram{\`e}r, Milad Nasr, Chiyuan Zhang, Matthew Jagielski, Katherine Lee, Christopher~A Choquette-Choo, and Nicholas Carlini. 2022.
\newblock Preventing verbatim memorization in language models gives a false sense of privacy.
\newblock \emph{arXiv preprint arXiv:2210.17546}.

\bibitem[{Jang et~al.(2022)Jang, Yoon, Yang, Cha, Lee, Logeswaran, and Seo}]{jang2022knowledge}
Joel Jang, Dongkeun Yoon, Sohee Yang, Sungmin Cha, Moontae Lee, Lajanugen Logeswaran, and Minjoon Seo. 2022.
\newblock Knowledge unlearning for mitigating privacy risks in language models.
\newblock \emph{arXiv preprint arXiv:2210.01504}.

\bibitem[{Jang et~al.(2023)Jang, Yoon, Yang, Cha, Lee, Logeswaran, and Seo}]{jang2023knowledge}
Joel Jang, Dongkeun Yoon, Sohee Yang, Sungmin Cha, Moontae Lee, Lajanugen Logeswaran, and Minjoon Seo. 2023.
\newblock Knowledge unlearning for mitigating privacy risks in language models.
\newblock In \emph{Proceedings of the 61st Annual Meeting of the Association for Computational Linguistics (Volume 1: Long Papers)}, pages 14389--14408.

\bibitem[{Jiang et~al.(2023)Jiang, Sablayrolles, Mensch, Bamford, Chaplot, Casas, Bressand, Lengyel, Lample, Saulnier et~al.}]{jiang2023mistral}
Albert~Q Jiang, Alexandre Sablayrolles, Arthur Mensch, Chris Bamford, Devendra~Singh Chaplot, Diego de~las Casas, Florian Bressand, Gianna Lengyel, Guillaume Lample, Lucile Saulnier, et~al. 2023.
\newblock Mistral 7b.
\newblock \emph{arXiv preprint arXiv:2310.06825}.

\bibitem[{Karamolegkou et~al.(2023)Karamolegkou, Li, Zhou, and S{\o}gaard}]{karamolegkou2023copyright}
Antonia Karamolegkou, Jiaang Li, Li~Zhou, and Anders S{\o}gaard. 2023.
\newblock Copyright violations and large language models.
\newblock \emph{arXiv preprint arXiv:2310.13771}.

\bibitem[{Li et~al.(2024)Li, Pan, Gopal, Yue, Berrios, Gatti, Li, Dombrowski, Goel, Phan, Mukobi, Helm-Burger, Lababidi, Justen, Liu, Chen, Barrass, Zhang, Zhu, Tamirisa, Bharathi, Khoja, Zhao, Herbert-Voss, Breuer, Marks, Patel, Zou, Mazeika, Wang, Oswal, Liu, Hunt, Tienken-Harder, Shih, Talley, Guan, Kaplan, Steneker, Campbell, Jokubaitis, Levinson, Wang, Qian, Karmakar, Basart, Fitz, Levine, Kumaraguru, Tupakula, Varadharajan, Shoshitaishvili, Ba, Esvelt, Wang, and Hendrycks}]{li2024wmdp}
Nathaniel Li, Alexander Pan, Anjali Gopal, Summer Yue, Daniel Berrios, Alice Gatti, Justin~D. Li, Ann-Kathrin Dombrowski, Shashwat Goel, Long Phan, Gabriel Mukobi, Nathan Helm-Burger, Rassin Lababidi, Lennart Justen, Andrew~B. Liu, Michael Chen, Isabelle Barrass, Oliver Zhang, Xiaoyuan Zhu, Rishub Tamirisa, Bhrugu Bharathi, Adam Khoja, Zhenqi Zhao, Ariel Herbert-Voss, Cort~B. Breuer, Samuel Marks, Oam Patel, Andy Zou, Mantas Mazeika, Zifan Wang, Palash Oswal, Weiran Liu, Adam~A. Hunt, Justin Tienken-Harder, Kevin~Y. Shih, Kemper Talley, John Guan, Russell Kaplan, Ian Steneker, David Campbell, Brad Jokubaitis, Alex Levinson, Jean Wang, William Qian, Kallol~Krishna Karmakar, Steven Basart, Stephen Fitz, Mindy Levine, Ponnurangam Kumaraguru, Uday Tupakula, Vijay Varadharajan, Yan Shoshitaishvili, Jimmy Ba, Kevin~M. Esvelt, Alexandr Wang, and Dan Hendrycks. 2024.
\newblock \href {https://arxiv.org/abs/2403.03218} {The wmdp benchmark: Measuring and reducing malicious use with unlearning}.
\newblock \emph{Preprint}, arXiv:2403.03218.

\bibitem[{Lin(2004)}]{lin2004rouge}
Chin-Yew Lin. 2004.
\newblock Rouge: A package for automatic evaluation of summaries.
\newblock In \emph{Text summarization branches out}, pages 74--81.

\bibitem[{Liu et~al.(2024{\natexlab{a}})Liu, Min, Zettlemoyer, Choi, and Hajishirzi}]{liu2024infini}
Jiacheng Liu, Sewon Min, Luke Zettlemoyer, Yejin Choi, and Hannaneh Hajishirzi. 2024{\natexlab{a}}.
\newblock Infini-gram: Scaling unbounded n-gram language models to a trillion tokens.
\newblock \emph{arXiv preprint arXiv:2401.17377}.

\bibitem[{Liu et~al.(2024{\natexlab{b}})Liu, Sun, Xu, Wu, Wang, Wang, and Gao}]{liu2024shield}
Xiaoze Liu, Ting Sun, Tianyang Xu, Feijie Wu, Cunxiang Wang, Xiaoqian Wang, and Jing Gao. 2024{\natexlab{b}}.
\newblock Shield: Evaluation and defense strategies for copyright compliance in llm text generation.
\newblock \emph{arXiv preprint arXiv:2406.12975}.

\bibitem[{Liu et~al.(2024{\natexlab{c}})Liu, Dou, Chien, Zhang, Tian, and Zhu}]{liu2024breaking}
Zheyuan Liu, Guangyao Dou, Eli Chien, Chunhui Zhang, Yijun Tian, and Ziwei Zhu. 2024{\natexlab{c}}.
\newblock Breaking the trilemma of privacy, utility, and efficiency via controllable machine unlearning.
\newblock In \emph{Proceedings of the ACM on Web Conference 2024}, pages 1260--1271.

\bibitem[{Liu et~al.(2024{\natexlab{d}})Liu, Dou, Tan, Tian, and Jiang}]{liu2024machine}
Zheyuan Liu, Guangyao Dou, Zhaoxuan Tan, Yijun Tian, and Meng Jiang. 2024{\natexlab{d}}.
\newblock Machine unlearning in generative ai: A survey.
\newblock \emph{arXiv preprint arXiv:2407.20516}.

\bibitem[{Liu et~al.(2024{\natexlab{e}})Liu, Dou, Tan, Tian, and Jiang}]{liu2024towards}
Zheyuan Liu, Guangyao Dou, Zhaoxuan Tan, Yijun Tian, and Meng Jiang. 2024{\natexlab{e}}.
\newblock Towards safer large language models through machine unlearning.
\newblock \emph{arXiv preprint arXiv:2402.10058}.

\bibitem[{{\L}ucki et~al.(2024){\L}ucki, Wei, Huang, Henderson, Tram{\`e}r, and Rando}]{lucki2024adversarial}
Jakub {\L}ucki, Boyi Wei, Yangsibo Huang, Peter Henderson, Florian Tram{\`e}r, and Javier Rando. 2024.
\newblock An adversarial perspective on machine unlearning for ai safety.
\newblock \emph{arXiv preprint arXiv:2409.18025}.

\bibitem[{Maini et~al.(2024)Maini, Feng, Schwarzschild, Lipton, and Kolter}]{maini2024tofu}
Pratyush Maini, Zhili Feng, Avi Schwarzschild, Zachary~C Lipton, and J~Zico Kolter. 2024.
\newblock Tofu: A task of fictitious unlearning for llms.
\newblock \emph{arXiv preprint arXiv:2401.06121}.

\bibitem[{Meeus et~al.(2024)Meeus, Shilov, Faysse, and de~Montjoye}]{meeus2024copyright}
Matthieu Meeus, Igor Shilov, Manuel Faysse, and Yves-Alexandre de~Montjoye. 2024.
\newblock Copyright traps for large language models.
\newblock \emph{arXiv preprint arXiv:2402.09363}.

\bibitem[{Meng et~al.(2022)Meng, Sharma, Andonian, Belinkov, and Bau}]{meng2022mass}
Kevin Meng, Arnab~Sen Sharma, Alex Andonian, Yonatan Belinkov, and David Bau. 2022.
\newblock Mass-editing memory in a transformer.
\newblock \emph{arXiv preprint arXiv:2210.07229}.

\bibitem[{Min et~al.(2023)Min, Gururangan, Wallace, Hajishirzi, Smith, and Zettlemoyer}]{min2023silo}
Sewon Min, Suchin Gururangan, Eric Wallace, Hannaneh Hajishirzi, Noah~A Smith, and Luke Zettlemoyer. 2023.
\newblock Silo language models: Isolating legal risk in a nonparametric datastore.
\newblock \emph{arXiv preprint arXiv:2308.04430}.

\bibitem[{Miyato et~al.(2016)Miyato, Dai, and Goodfellow}]{miyato2016adversarial}
Takeru Miyato, Andrew~M Dai, and Ian Goodfellow. 2016.
\newblock Adversarial training methods for semi-supervised text classification.
\newblock \emph{arXiv preprint arXiv:1605.07725}.

\bibitem[{Nasr et~al.(2023)Nasr, Carlini, Hayase, Jagielski, Cooper, Ippolito, Choquette-Choo, Wallace, Tram{\`e}r, and Lee}]{nasr2023scalable}
Milad Nasr, Nicholas Carlini, Jonathan Hayase, Matthew Jagielski, A~Feder Cooper, Daphne Ippolito, Christopher~A Choquette-Choo, Eric Wallace, Florian Tram{\`e}r, and Katherine Lee. 2023.
\newblock Scalable extraction of training data from (production) language models.
\newblock \emph{arXiv preprint arXiv:2311.17035}.

\bibitem[{Neelakantan et~al.(2015)Neelakantan, Vilnis, Le, Sutskever, Kaiser, Kurach, and Martens}]{neelakantan2015adding}
Arvind Neelakantan, Luke Vilnis, Quoc~V Le, Ilya Sutskever, Lukasz Kaiser, Karol Kurach, and James Martens. 2015.
\newblock Adding gradient noise improves learning for very deep networks.
\newblock \emph{arXiv preprint arXiv:1511.06807}.

\bibitem[{Rahman and Santacana(2023)}]{rahman2023beyond}
Noorjahan Rahman and Eduardo Santacana. 2023.
\newblock Beyond fair use: Legal risk evaluation for training llms on copyrighted text.
\newblock In \emph{ICML Workshop on Generative AI and Law}.

\bibitem[{Srivastava et~al.(2014)Srivastava, Hinton, Krizhevsky, Sutskever, and Salakhutdinov}]{srivastava2014dropout}
Nitish Srivastava, Geoffrey Hinton, Alex Krizhevsky, Ilya Sutskever, and Ruslan Salakhutdinov. 2014.
\newblock Dropout: a simple way to prevent neural networks from overfitting.
\newblock \emph{The journal of machine learning research}, 15(1):1929--1958.

\bibitem[{Team et~al.(2024)}]{mosaic2024introducing}
Mosaic AI~Research Team et~al. 2024.
\newblock Introducing dbrx: a new state-of-the-art open llm.
\newblock \emph{Mosaic AI Res. Available online at: https://www. databricks. com/blog/introducing-dbrx-new-state-art-open-llm (accessed June 4, 2024)}.

\bibitem[{Touvron et~al.(2023)Touvron, Martin, Stone, Albert, Almahairi, Babaei, Bashlykov, Batra, Bhargava, Bhosale et~al.}]{touvron2023llama}
Hugo Touvron, Louis Martin, Kevin Stone, Peter Albert, Amjad Almahairi, Yasmine Babaei, Nikolay Bashlykov, Soumya Batra, Prajjwal Bhargava, Shruti Bhosale, et~al. 2023.
\newblock Llama 2: Open foundation and fine-tuned chat models.
\newblock \emph{arXiv preprint arXiv:2307.09288}.

\bibitem[{Wei et~al.(2024)Wei, Shi, Huang, Smith, Zhang, Zettlemoyer, Li, and Henderson}]{wei2024evaluating}
Boyi Wei, Weijia Shi, Yangsibo Huang, Noah~A Smith, Chiyuan Zhang, Luke Zettlemoyer, Kai Li, and Peter Henderson. 2024.
\newblock Evaluating copyright takedown methods for language models.
\newblock \emph{arXiv preprint arXiv:2406.18664}.

\bibitem[{Yang et~al.(2024{\natexlab{a}})Yang, Sun, Ma, Liu, Yin, and Cheng}]{yang2024butterfly}
Wanli Yang, Fei Sun, Xinyu Ma, Xun Liu, Dawei Yin, and Xueqi Cheng. 2024{\natexlab{a}}.
\newblock The butterfly effect of model editing: Few edits can trigger large language models collapse.
\newblock \emph{arXiv preprint arXiv:2402.09656}.

\bibitem[{Yang et~al.(2024{\natexlab{b}})Yang, Jones, Mozer, and Ren}]{yang2024reawakening}
Yanlai Yang, Matt Jones, Michael~C Mozer, and Mengye Ren. 2024{\natexlab{b}}.
\newblock Reawakening knowledge: Anticipatory recovery from catastrophic interference via structured training.
\newblock \emph{arXiv preprint arXiv:2403.09613}.

\bibitem[{Yao et~al.(2024)Yao, Chien, Du, Niu, Wang, Cheng, and Yue}]{yao2024machine}
Jin Yao, Eli Chien, Minxin Du, Xinyao Niu, Tianhao Wang, Zezhou Cheng, and Xiang Yue. 2024.
\newblock Machine unlearning of pre-trained large language models.
\newblock \emph{arXiv preprint arXiv:2402.15159}.

\bibitem[{Yao et~al.(2023)Yao, Xu, and Liu}]{yao2023large}
Yuanshun Yao, Xiaojun Xu, and Yang Liu. 2023.
\newblock Large language model unlearning.
\newblock \emph{arXiv preprint arXiv:2310.10683}.

\bibitem[{Yu et~al.(2023)Yu, Pang, Liu, Du, Kang, Huang, Lin, and Yan}]{yu2023bag}
Weichen Yu, Tianyu Pang, Qian Liu, Chao Du, Bingyi Kang, Yan Huang, Min Lin, and Shuicheng Yan. 2023.
\newblock Bag of tricks for training data extraction from language models.
\newblock In \emph{International Conference on Machine Learning}, pages 40306--40320. PMLR.

\bibitem[{Zhang et~al.(2023)Zhang, Ippolito, Lee, Jagielski, Tram{\`e}r, and Carlini}]{zhang2023counterfactual}
Chiyuan Zhang, Daphne Ippolito, Katherine Lee, Matthew Jagielski, Florian Tram{\`e}r, and Nicholas Carlini. 2023.
\newblock Counterfactual memorization in neural language models.
\newblock \emph{Advances in Neural Information Processing Systems}, 36:39321--39362.

\bibitem[{Zhang et~al.(2024)Zhang, Lin, Bai, and Mei}]{zhang2024negative}
Ruiqi Zhang, Licong Lin, Yu~Bai, and Song Mei. 2024.
\newblock Negative preference optimization: From catastrophic collapse to effective unlearning.
\newblock \emph{arXiv preprint arXiv:2404.05868}.

\bibitem[{Zhao et~al.(2024)Zhao, Hu, Li, Deng, Zhao, Qin, and Chua}]{zhao2024towards}
Weixiang Zhao, Yulin Hu, Zhuojun Li, Yang Deng, Yanyan Zhao, Bing Qin, and Tat-Seng Chua. 2024.
\newblock Towards comprehensive and efficient post safety alignment of large language models via safety patching.
\newblock \emph{arXiv preprint arXiv:2405.13820}.

\bibitem[{Zheng et~al.(2023)Zheng, Chiang, Sheng, Zhuang, Wu, Zhuang, Lin, Li, Li, Xing, Zhang, Gonzalez, and Stoica}]{zheng2023judging}
Lianmin Zheng, Wei-Lin Chiang, Ying Sheng, Siyuan Zhuang, Zhanghao Wu, Yonghao Zhuang, Zi~Lin, Zhuohan Li, Dacheng Li, Eric.~P Xing, Hao Zhang, Joseph~E. Gonzalez, and Ion Stoica. 2023.
\newblock \href {https://arxiv.org/abs/2306.05685} {Judging llm-as-a-judge with mt-bench and chatbot arena}.
\newblock \emph{Preprint}, arXiv:2306.05685.

\bibitem[{Zhou et~al.(2023)Zhou, Lu, Ma, Gui, Zhang, and Huang}]{zhou2023making}
Xin Zhou, Yi~Lu, Ruotian Ma, Tao Gui, Qi~Zhang, and Xuanjing Huang. 2023.
\newblock Making harmful behaviors unlearnable for large language models.
\newblock \emph{arXiv preprint arXiv:2311.02105}.

\end{thebibliography}

\newpage
\appendix
\label{sec:appendix}

\section{Experiment Details}
\subsection{Experiment Settings}
\label{sec:appendix-experiment_details:experiment_settings}
To evaluate the effectiveness of sequential unlearning, we conduct experiments on several copyrighted books. Our process involves the following steps:

First, each book is split into many chunks of 200 tokens. For each chunk, the initial 100 tokens are used as a prompt, which is fed into the LLM. The remaining 100 tokens serve as the answer or continuation from the original book. This setup allows us to assess how well the model can generate text that follows the given prompt.

In addition to the prompt from the book, we use a system prompt and a instruction prompt to guide the model in generating the completion. 

The default system prompt is 

\begin{quote} "You are a helpful, respectful and honest assistant." \end{quote}

and the default instruction prompt is 

\begin{quote} "Please complete the rest of the following paragraph based on the context." \end{quote}

For each prompt, the model generates a completion using a nucleus sampling by setting the temperature to 0.4 and $\eta = 0.6$. This follows bags of tricks to extract training data suggested by \citet{yu2023bag}.

To evaluate the generated completions and its risk of copyright infringement, we use Rouge-1 and ROUGE-L score. These metrics allow us to compare the LLM's completions with the original text and assess the model's ability to unlearn specific content.

Specifically, we evaluate the scores on the following sets of books:
\begin{itemize}
\item Books to be forgotten ($D_f$)
\item Books that are previously unlearned when time step is greater than one ($D_{prev}$)
\item Books that not to be forgotten ($D_{nor}$) 

\end{itemize}
 In subsequent sections, we refer to the performance on  $D_{nor}$ as knowledge retention.

Additionally, following the setup of \citet{wei2024evaluating}, we evaluate the model's performance on general reasoning tasks to assess its capability retention. The downstream tasks considered include 5-shot Massive Multitask Language Understanding (MMLU) \cite{hendrycks2020measuring}, and MT-Bench \cite{zheng2023judging}. 

\subsection{Evaluation Metrics}
\label{sec:appendix-experiment_details:eval_metrics}

\subsubsection{Rouge-1} Recall-Oriented Understudy for Gisting Evaluation (Rouge) includes Rouge-1, which measures the overlap of unigram (single word) occurrences between the LLM's completion and the original books. A unigram is any individual word that appears in both the completion (hypothesis text) and the original book (reference text).

First, we define the recall as the ratio of the number of overlapping unigrams between the hypothesis and reference text to the total number of unigrams in the reference text: \begin{equation*} Recall = \frac{\text{overlapping unigrams}}{\text{total unigrams in the reference}}. \end{equation*}

Similarly, precision is defined as the ratio of the number of overlapping unigrams to the total number of unigrams in the hypothesis text: \begin{equation*} Precision = \frac{\text{overlapping unigrams}}{\text{total unigrams in the hypothesis}}. \end{equation*}

Lastly, the Rouge-1 score used in our experiments is calculated as the F1 score, which combines precision and recall: \begin{equation*} F1 = 2 \cdot \frac{Precision \cdot Recall}{Precision + Recall} \end{equation*}

\subsubsection{Rouge-L}
Rouge-L measures the longest common subsequence (LCS) between the LLM's completion and original books. In detail, LCS is a sequence that appears in both the completion (hypothesis text) and original book (reference text) in the same order but not necessarily contiguously. 

Next, we define the recall as the ratio of the length of the LCS to the total length of the reference text:
\begin{equation*}
Recall = \frac{LCS}{\text{length of the reference text}}.
\end{equation*}

We also define the precision as the ratio of the length of the LCS to the total length of the hypothesis text:
\begin{equation*}
Precision = \frac{LCS}{\text{length of the hypothesis text}}.
\end{equation*}

Lastly, the Rouge-L score we used in our experiments is defined as:
\begin{equation*}
F1 = 2 \cdot \frac{Precision \cdot Recall}{Precision + Recall}
\end{equation*}

\subsection{Datasets}
\label{sec:appendix-experiment_details:datasets}
This section provides detailed information about the books used in the experiment. We crawled all available books from Project Gutenberg\footnote{\href{https://www.gutenberg.org/}{gutenberg.org}} and pre-processed them following the methodology of \citet{gerlach2020standardized}, in which we remove all headers and boiler plate text. 

\subsubsection{Books to Forget}
At time step 1, we unlearn The Adventures of Sherlock Holmes by Arthur Conan Doyle. At time step 2, we unlearn Flowers of the Sky by Richard A. Proctor, Pagan Papers by Kenneth Grahame at time step 3, Diary of Samuel Pepys by Samuel Pepys at time step 4, Pride and Prejudice by Jane Austen at time step 5, They Call Me Carpenter: A Tale of the Second Coming by Upton Sinclair at time step 6, Memoirs of the Court of Louis XIV. and of the Regency — Complete by Orléans at time step 7, Alice's Adventures in Wonderland by Lewis Carroll at time step 8, The Wonderful Adventures of Nils by Selma Lagerlöf at time step 9, and Starr, of the Desert by B. M. Bower at time step 10. 

For the books in $D_f$, the entire texts are split into chunks of 200 tokens, and the dataset is formatted as question-answer pairs, where the first 100 tokens represent the Question, and the subsequent 100 tokens represent the Answer. All texts from each book are included and formatted into JSON files.

At each time step greater than one, we form $D_{prev}$ by collecting all the books that we have unlearned since the first time step. Specifically, we then split all of the books that are used in previous unlearning steps into chunks of 200 tokens to form many QA pairs, and then randomly sampled 200 such pairs to form $D_{prev}$.

\subsubsection{Books in $D_{nor}$} To evaluate the model's knowledge retention after unlearning, we constructed a set of books that we do not want the model to forget. Specifically, we randomly sampled 100 books from Project Gutenberg that are not included in $D_f$. Following the same preprocessing steps used for $D_f$, we split these books into chunks of 200 tokens, and randomly sampled 200 chunks to form the dataset $D_{nor}$. Since the books in $D_{nor}$ are not part of $D_f$ at any time step, the content of $D_{nor}$ remains unchanged throughout the experiment.

\subsection{Baseline Methods}
\label{sec:appendix-experiment_details:baseline_methods}

\subsubsection{Generic Prevention using System Prompt}
The default system prompts being used for all the unlearning and MemFree decoding methods have been described in Appendix \ref{sec:appendix-experiment_details:experiment_settings}. Here, we present two copyright takedown methods using well crafted system prompts. 

First, we denote our baseline method Prompt (a) as using the system prompt: 
\begin{quote} "You are a helpful, respectful and honest assistant. When generating your response, please do not generate the copyrighted content." \end{quote}

which adds a brief instruction to not generate copyrighted content after the default system prompt. 

We also included Prompt (dbrx), which is a system prompt being used by the DBRX model, released by Databricks:

\begin{quote} "You are a helpful, respectful and honest assistant. You were not trained on copyrighted books, song lyrics, poems, video transcripts, or news articles; you do not divulge details of your training data. You do not provide song lyrics, poems, or news articles and instead refer the user to find them online or in a store." \end{quote}

\subsubsection{MemFree Decoding} To prevent the model from emitting memorized content, we employed MemFree decoding \cite{ippolito2022preventing}. This method checks each n-gram during text generation to ensure it does not match any sequences from the training set. If a match is detected, the token is resampled, thereby avoiding verbatim reproduction of training data. The process is optimized through the use of Bloom filters, which allow for efficient real-time memorization checks. Although MemFree effectively stops exact memorization, it does not fully eliminate the risk of paraphrased or approximate memorization. We implemented MemFree based on \citet{wei2024evaluating}.

\subsubsection{Unlearning via Gradient Difference}
\label{sec:appendix-experiment_details:baseline_methods-GA}
In this work, we use the method proposed by ~\cite{yao2023large} as one of the baseline methods. Here we present the case of performing gradient difference unlearning. 

Specifically, let $\theta$ to be the current LLM, let $DD_f$ to be the dataset representing the book we want to forget, and $D_{add}$ to a set of book corpora that does not contain the book to be forgotten, nor the books in $D_{nor}$. Moreover, we define $h_\theta(x, y_{y<i}) = \mathbb{P}(y_i | (x, y_{<i}); \theta)$, which is the probability of the token $y_i$ conditioned on the prompt $x$ and the already generated tokens $y_{<i} = [y_1, y_2, ..., y_{i-1}]$. Next, we define the LLM's loss on y as: \begin{equation*}
    L(x, y; \theta) := \sum_{i=1}^{|y|} \ell (h_\theta(x, y_{<i}), y_i)
\end{equation*}

The Gradient Difference has three loss terms, defined as follows:
\begin{equation*}
    \mathcal{L}_{\text{fgt}} = - \sum_{(x_{\text{fgt}}, y_{\text{fgt}}) \in D_f} L(x_{\text{fgt}}, y_{\text{fgt}}, \theta_t)
\end{equation*}
\begin{equation*}
    \mathcal{L}_{\text{rnd}} := \sum_{(x_{\text{fgt}}, ) \in D_{f}} \frac{1}{|Y_{\text{rnd}}|} \sum_{(,y_{\text{rnd})} \in Y_{\text{rnd}}} L(x_{\text{fgt}}, y_{\text{rnd}}, \theta_t)
\end{equation*}
\begin{equation*}
    \phi_\theta = h_{\theta} (x_{\text{nor}}, y_{\text{nor}<i})
\end{equation*}
\begin{equation*}
    \mathcal{L}_{\text{add}} := \sum_{(x_{\text{add}}, y_{\text{add}}) \in D_{\text{add}}} \sum_{i=1}^{|y_{\text{add}}|} \text{KL}(\phi_{\theta_o} \parallel \phi_{\theta_t}).
\end{equation*}
in which $Y_{\text{rnd}}$ is a set of responses irrelevant to responses of $D_f$, sampled from $D_{add}$. 

Lastly, the GA approach is trying to minimize the following loss to obtain the unlearned model:
\begin{equation*}
    L = \epsilon_1 \mathcal{L}_{\text{fgt}} + \epsilon_2  \mathcal{L}_{\text{rnd}} + \epsilon_3  \mathcal{L}_{\text{add}}
\end{equation*}
\begin{equation*}
    \theta_{t + 1} \leftarrow \theta_t - \nabla L.
\end{equation*}
in which $\mathcal{L}_{\text{fgt}}$ is a gradient ascent loss on $D_f$, which tries to make the model perform poorly on the $D_f$. Next, $\mathcal{L}_{\text{rnd}}$ tries to randomly mismatch the labels from non-relevant dataset to the inputs of the dataset we want to forget. Lastly, $\mathcal{L}_{\text{add}}$ tries to maintain the performance on the normal dataset. In our experiment, we set $\epsilon_1 = 1$, $\epsilon_2 = \epsilon_3 = 0.5$ across all time steps and all models.

\subsubsection{Unlearning via Task Vector}
We also use the task vector method as one of the baseline approaches, which typically involves a two-stage process. Considering the case of $t = 1$, we denote $\theta_o$ as the original model weights. We intentionally fine-tune the model on $D_f$ to obtain $\theta_{ft}^1$. This fine-tuning process is defined as follows:

\begin{equation*}
    \mathcal{L}_{\text{fgt}} = \sum_{(x_{\text{fgt}}, y_{\text{fgt}}) \in D_f} L(x_{\text{fgt}}, y_{\text{fgt}}, \theta_t)
\end{equation*}
\begin{equation*}
    \theta_{t+1} \leftarrow \theta_t - \epsilon \nabla_{\theta_t} \mathcal{L}_{\text{fgt}}
\end{equation*}

Next, we define the task vector $\tau$ as the element-wise difference between $\theta_{ft}$ and $\theta_o$:

\begin{equation*}
    \tau = \theta_{ft}^1 - \theta_o
\end{equation*}

Finally, the unlearned model $\theta_u$ at time step $t$ is obtained by:

\begin{equation*}
    \theta_u^1 = \theta_o - \tau
\end{equation*}

The general intuition behind this method is to first obtain a model that is specialized in the dataset we aim to forget. The task vector $\tau$ represents the changes in weights required to acquire this specific knowledge. By subtracting these "knowledge" weights from the original model, we effectively unlearn the targeted information.

\subsubsection{Unlearning via NPO}
\label{sec:appendix-experiment_details:unlearning_NPO}

In this work, we utilize the Negative Preference Optimization (NPO) method for unlearning undesirable data, aiming to overcome the catastrophic collapse often observed with gradient ascent methods. NPO builds on the framework of preference optimization, specifically focusing on negative samples from the dataset to be unlearned.

The NPO loss function is defined as follows:
\begin{equation*}
    \mathcal{L}_{\text{NPO}} = \frac{2}{\beta} \mathbb{E}_{(x, y) \in D_{\text{f}}} \left[ \log \left(1 + \left(\frac{\pi_\theta(y|x)}{\pi_{\text{ref}}(y|x)}\right)^\beta \right) \right]
\end{equation*}
where \( \pi_\theta(y|x) \) represents the prediction probability of the current model for token \( y \) given the input \( x \), and \( \pi_{\text{ref}}(y|x) \) is the prediction probability from the reference model trained on the entire dataset. The parameter \( \beta \) controls the smoothness of the optimization, and as \( \beta \to 0 \), the NPO loss converges to the standard gradient ascent loss. 

Minimizing this loss helps reduce the model’s reliance on the forget set, ensuring that the unlearning process remains stable and avoids the rapid deterioration seen in gradient ascent approaches. In our experiment, we set $\beta = 0.4$, and we obtain $\pi_{\text{ref}}$ by optimizing off-the-shelf LLMs on $D_{f} \bigcup D_{nor}$. 

\begin{figure*}
\centering
         \begin{subfigure}[b]{\textwidth}
            \centering
            \includegraphics[width=0.9\textwidth]{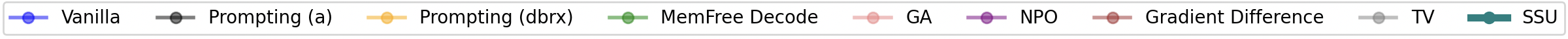}
          \end{subfigure}
        \begin{subfigure}{0.43\textwidth}
        \includegraphics[width=\textwidth]{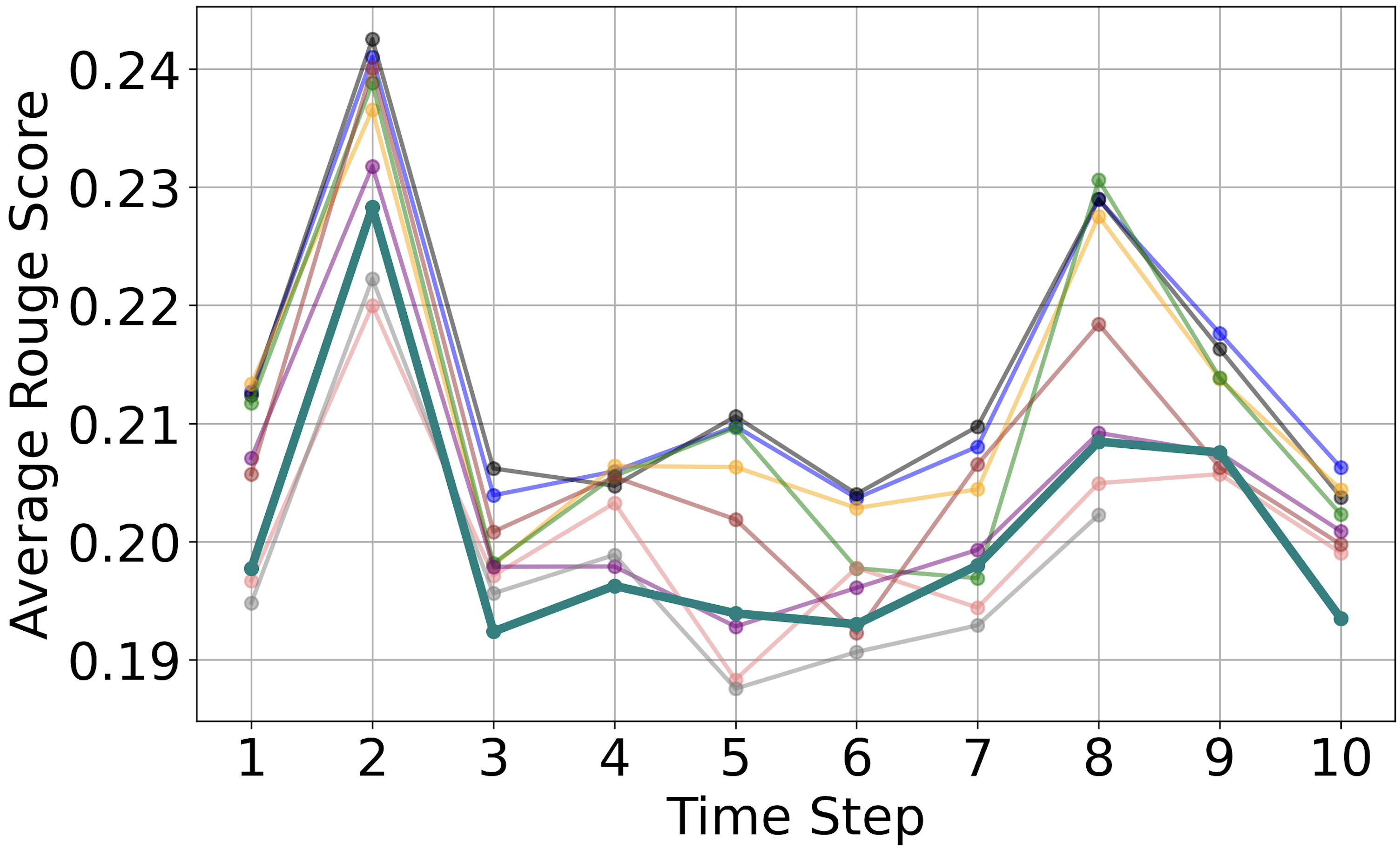}
        \subcaption{Avg Rouge Score on $D_f$}
        \label{fig:main_book_forget_unlearn}
        \end{subfigure}
        \begin{subfigure}{0.41\textwidth}
        \includegraphics[width=\textwidth]{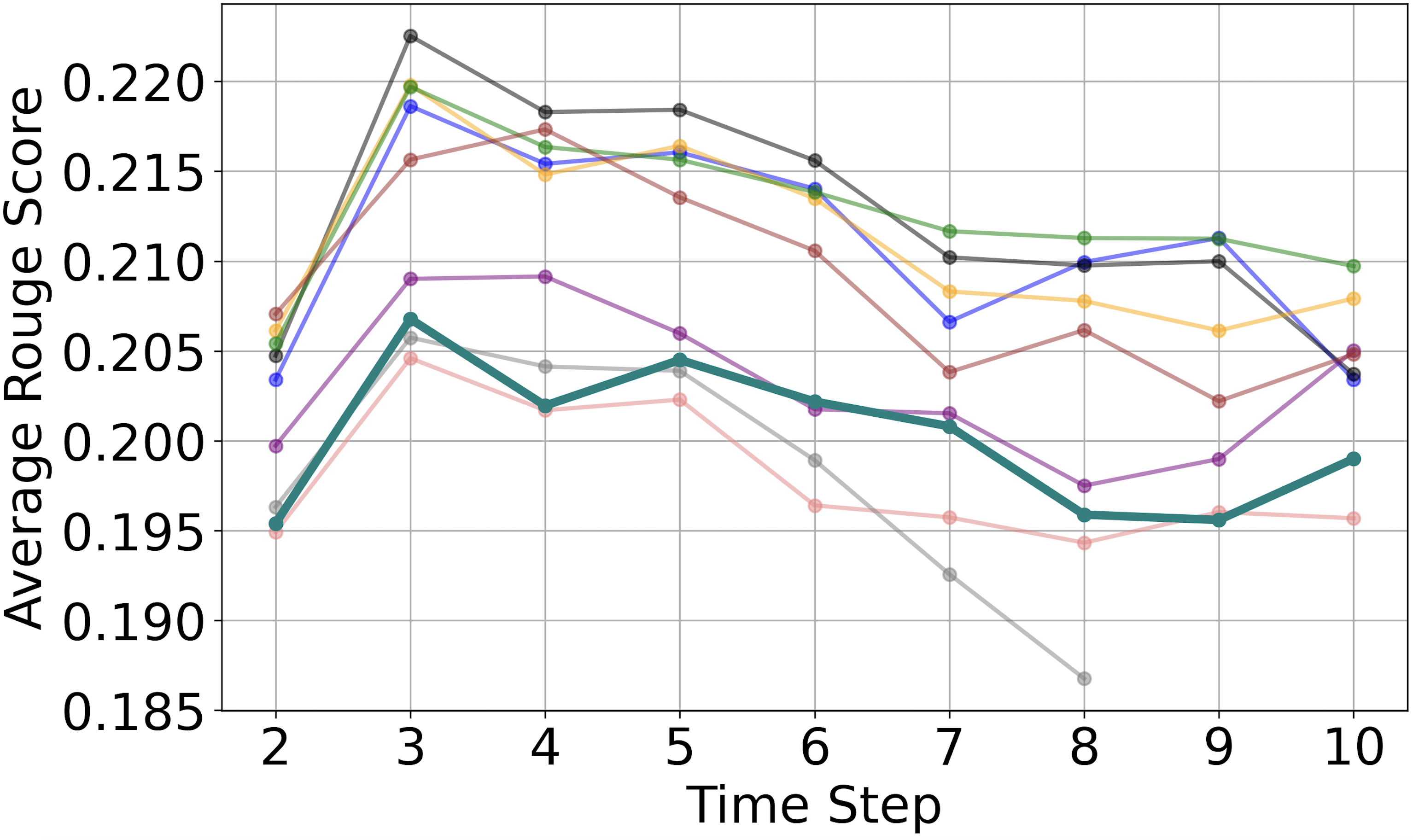}
        \subcaption{Avg Rouge Score on $D_{prev}$}
        \label{fig:main_book_forget_previous}
        \end{subfigure} 
        
      \begin{subfigure}{0.43\textwidth}
        \includegraphics[width=\textwidth]{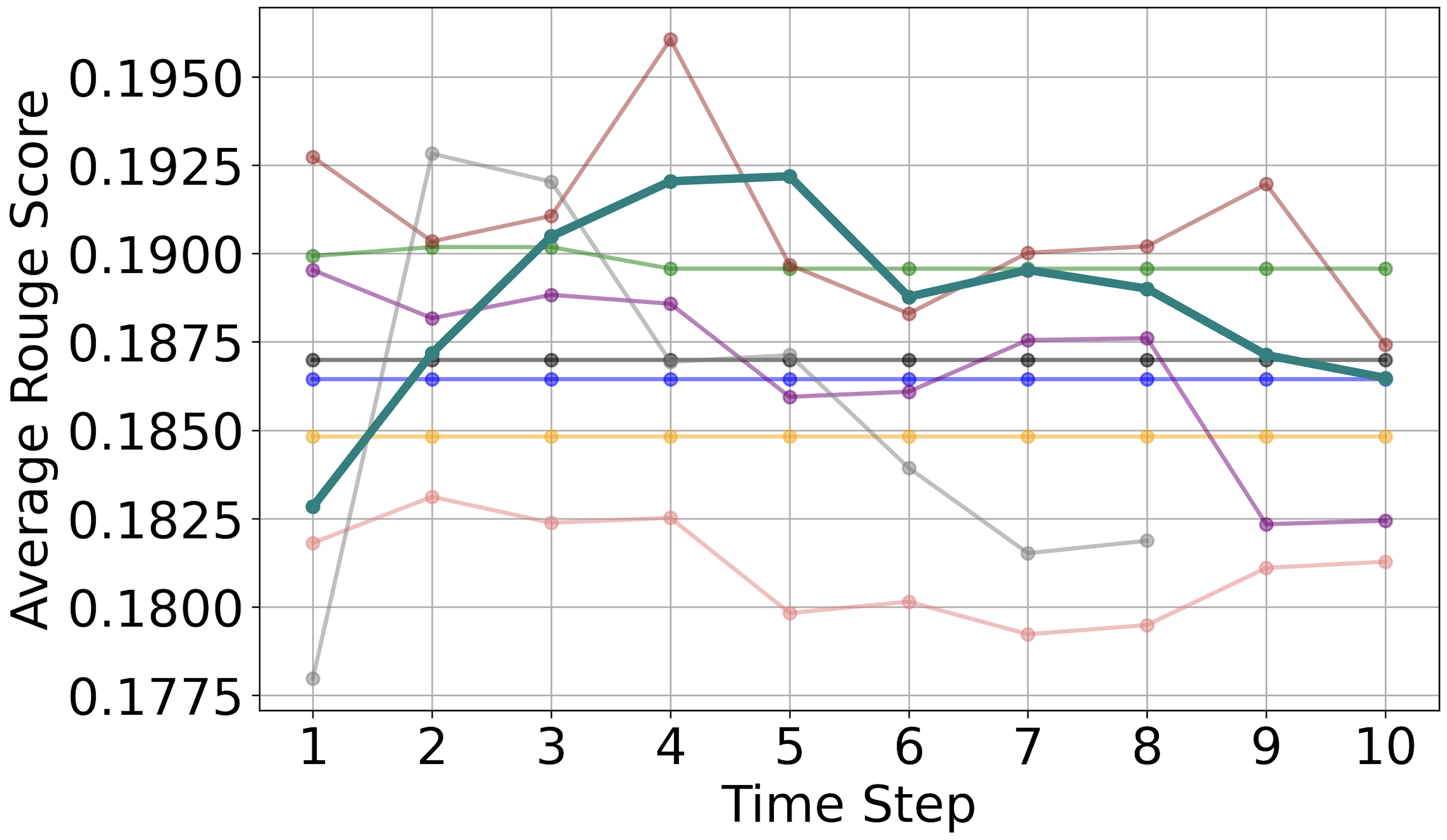}
        \subcaption{Avg Rouge Score on $D_n$}
        \label{fig:main_book_forget_normal}
        \end{subfigure}
        \begin{subfigure}{0.41\textwidth}
        \includegraphics[width=\textwidth]{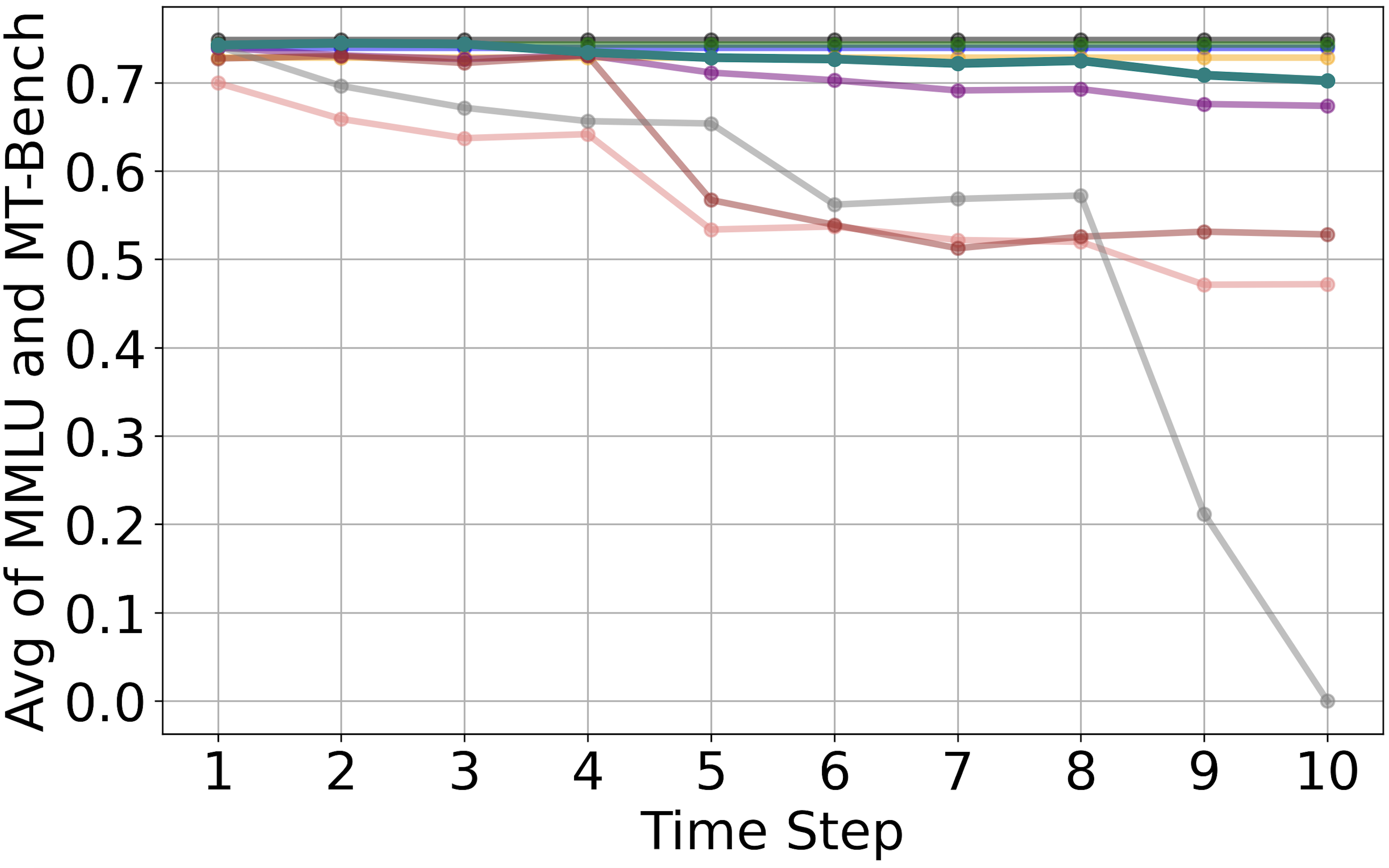}
        \subcaption{Avg MMLU and MT-Bench score}
\label{fig:main_book_forget_general_llama3.1}
        \end{subfigure}
\caption{The average of Rouge-1 and Rouge-l and benchmark scores for Llama3.1: (a) books to forget $D_f$ ($\downarrow$); (b) previously unlearned books $D_{prev}$ ($\downarrow$); (c) $D_{nor}$ ($\uparrow$). and (d) averaged normalized MMLU and MT-Bench scores ($\uparrow$). The results for TV after time step 8 are omitted due to collapse. Lower Rouge scores for $D_f$ and $D_{prev}$ indicate better unlearning, while higher scores for $D_{nor}$ and benchmarks reflect better performance.}
\label{fig:main_book_forget_all_appendix}
\end{figure*}

\subsection{Implementation Details}
The experiments are conducted on eight RTX A6000 GPUs. For all unlearning algorithms, at each time step, we only unlearn the model for 1 epoch, with a batch size set to 2. During all the fine-tuning process, we used Lora \cite{hu2021lora}, and we did not quantize the model because quantization leads to inaccurate element-wise subtraction for TV-based methods. 

For the Llama-3.1-8B-Instruct model, we set the learning rate to be 1e-5 for the first five time steps, and decrease the learning rate to 1e-6 for the rest of the time steps. For the Mistral-7B-Instruct-v0.3, we set the learning rate to be 1e-5 for the first three time steps, and the learning rate to be 1e-6 for other time steps. For SSU, we set $\epsilon_1 = 1$,a and $\epsilon_2 = 0.5$ for all the time steps and models. We set $\gamma$ to be 1 standard deviation away from the  mean of the gradient vector $\nabla_{\theta} L_f (\theta_t)$

\section{Additional Analysis}

\subsection{Impact of $\epsilon_1$ and $\epsilon_2$}
\label{sec:appendix-epsilon_results}

In this section, we analyze the impact of the hyperparameters $\epsilon_1$ and $\epsilon_2$ in Equation \ref{equation:ssu_main_no_weight_saliency}. Specifically, we denote $SSU_{1.5,1}$ as the method with $\epsilon_1 = 1.5$ and $\epsilon_2 = 1$ on Llama 3.1, and $SSU_{1,2}$ as the method with $\epsilon_1 = 1$ and $\epsilon_2 = 2$. All other hyperparameters remain consistent with the settings used in our main experiments.

As shown in Table \ref{tab:time_steps_epsilon_results}, SSU maintains strong performance on MMLU and MT-Bench, indicating its stability across different parameter values. Unlike gradient ascent and gradient difference methods, SSU does not lead to model collapse. However, in our main experiments, we tuned $\epsilon_1$ and $\epsilon_2$ to achieve the best trade-off between unlearning efficacy and general-purpose language abilities.

\begin{table*}[t]
    \centering
    \renewcommand{\arraystretch}{1.2} 
    \setlength{\tabcolsep}{6pt} 
    \resizebox{\textwidth}{!}{ 
    \begin{tabular}{cccccccccc}
        \toprule
        Time Step & Method & $D_f$  & $D_f$  & $D_{prev}$  & $D_{prev}$ & $D_{nor}$  & $D_{nor}$  & MMLU & MT-Bench \\
        \midrule
        1 & $SSU_{1.5,1}$ & 0.2513 & 0.1424 & N/A & N/A & 0.2307 & 0.1335 & 0.6631 & 8.1180 \\
        1 & $SSU_{1,2}$   & 0.2515 & 0.1417 & N/A & N/A & 0.2316 & 0.1335 & 0.6624 & 7.8058 \\
        \midrule
        2 & $SSU_{1.5,1}$ & 0.2856 & 0.1688 & 0.2558 & 0.1433 & 0.2402 & 0.1376 & 0.6463 & 8.1850 \\
        2 & $SSU_{1,2}$   & 0.2852 & 0.1689 & 0.2567 & 0.1433 & 0.2393 & 0.1367 & 0.6470 & 8.1675 \\
        \midrule
        3 & $SSU_{1.5,1}$ & 0.2415 & 0.1400 & 0.2699 & 0.1539 & 0.2403 & 0.1378 & 0.6463 & 7.9870 \\
        3 & $SSU_{1,2}$   & 0.2414 & 0.1386 & 0.2704 & 0.1543 & 0.2394 & 0.1370 & 0.6466 & 8.1283 \\
        \midrule
        4 & $SSU_{1.5,1}$ & 0.2492 & 0.1390 & 0.2576 & 0.1517 & 0.2458 & 0.1389 & 0.6431 & 8.0349 \\
        4 & $SSU_{1,2}$   & 0.2420 & 0.1403 & 0.2565 & 0.1511 & 0.2439 & 0.1391 & 0.6431 & 8.0266 \\
        \midrule
        5 & $SSU_{1.5,1}$ & 0.2551 & 0.1365 & 0.2650 & 0.1527 & 0.2441 & 0.1385 & 0.6298 & 7.9812 \\
        5 & $SSU_{1,2}$   & 0.2545 & 0.1369 & 0.2640 & 0.1517 & 0.2429 & 0.1378 & 0.6319 & 8.0578 \\
        \midrule
        6 & $SSU_{1.5,1}$ & 0.2494 & 0.1412 & 0.2621 & 0.1472 & 0.2435 & 0.1392 & 0.6298 & 7.9812 \\
        6 & $SSU_{1,2}$   & 0.2476 & 0.1408 & 0.2632 & 0.1467 & 0.2433 & 0.1380 & 0.6263 & 8.0484 \\
        \bottomrule
    \end{tabular}
    } 
    \caption{Results across time steps with different $\epsilon$. All performance metrics on $D$ are reported as Rouge-L scores.}
    \label{tab:time_steps_epsilon_results}
\end{table*}

\subsection{Re-Emergence in Sequential Unlearning}
\label{sec:appendix-sequential_unlearning_challenge} 

In the main text, we use $D_{prev}$ -the set of all previously unlearned books—to assess whether they remain unlearned throughout sequential unlearning. Here, instead of evaluating all unlearned books collectively, we examine how knowledge of each previously unlearned book evolves as new books are unlearned at later time steps. 

\begin{figure*}
\centering
         \begin{subfigure}[b]{\textwidth}
            \centering
            \includegraphics[width=0.8\textwidth]{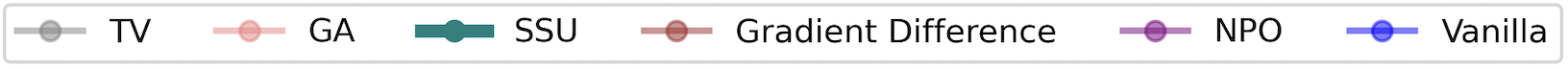}
          \end{subfigure}
        \begin{subfigure}{0.48\textwidth}
        \includegraphics[width=\textwidth]{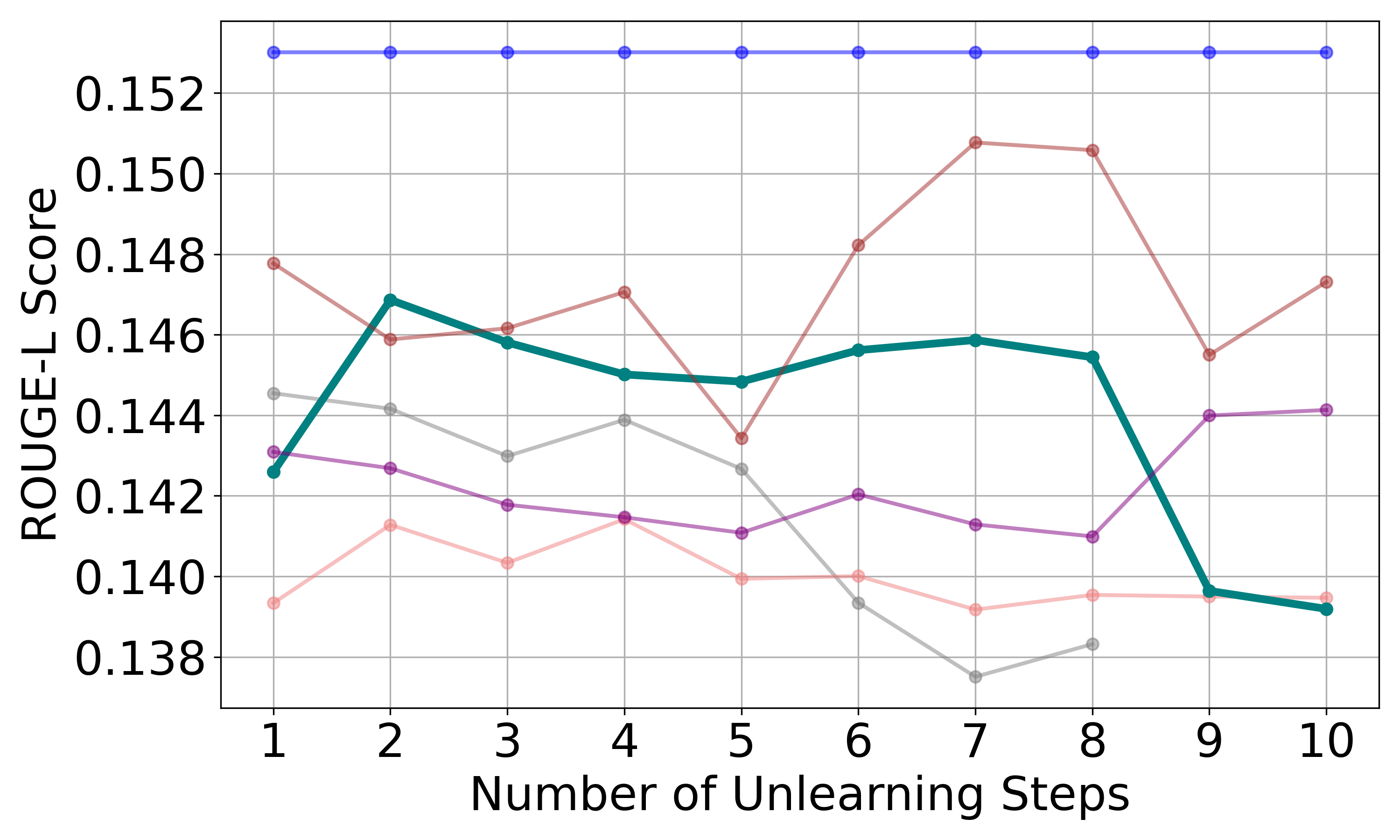}
        \subcaption{ROUGE-L Score for Book 1}
        \label{fig:rouge_l_book_1_individual}
        \end{subfigure}
        \begin{subfigure}{0.48\textwidth}
        \includegraphics[width=\textwidth]{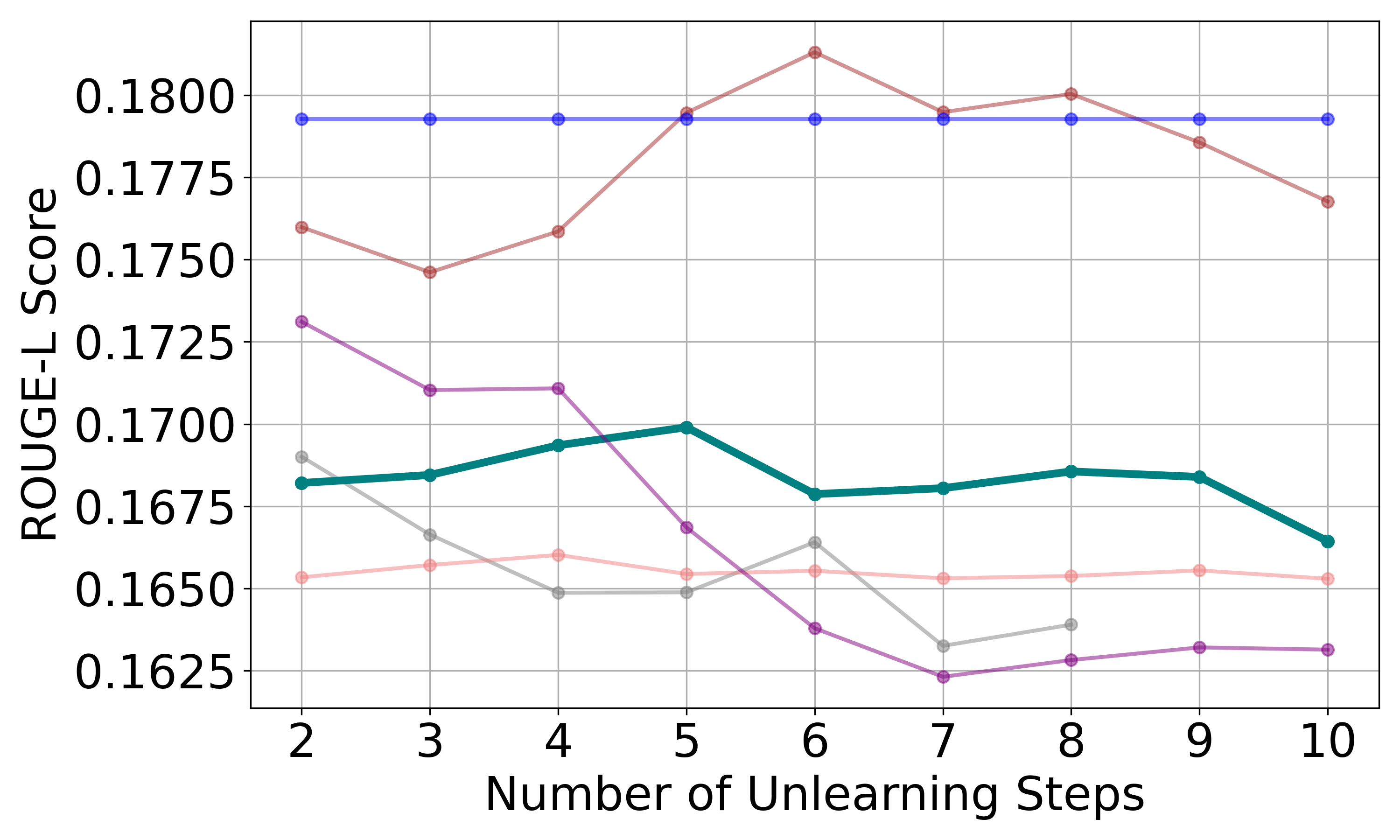}
        \subcaption{ROUGE-L Score for Book 2}
        \label{fig:rouge_l_book_2_individual}
        \end{subfigure}

       \begin{subfigure}{0.48\textwidth}
        \includegraphics[width=\textwidth]{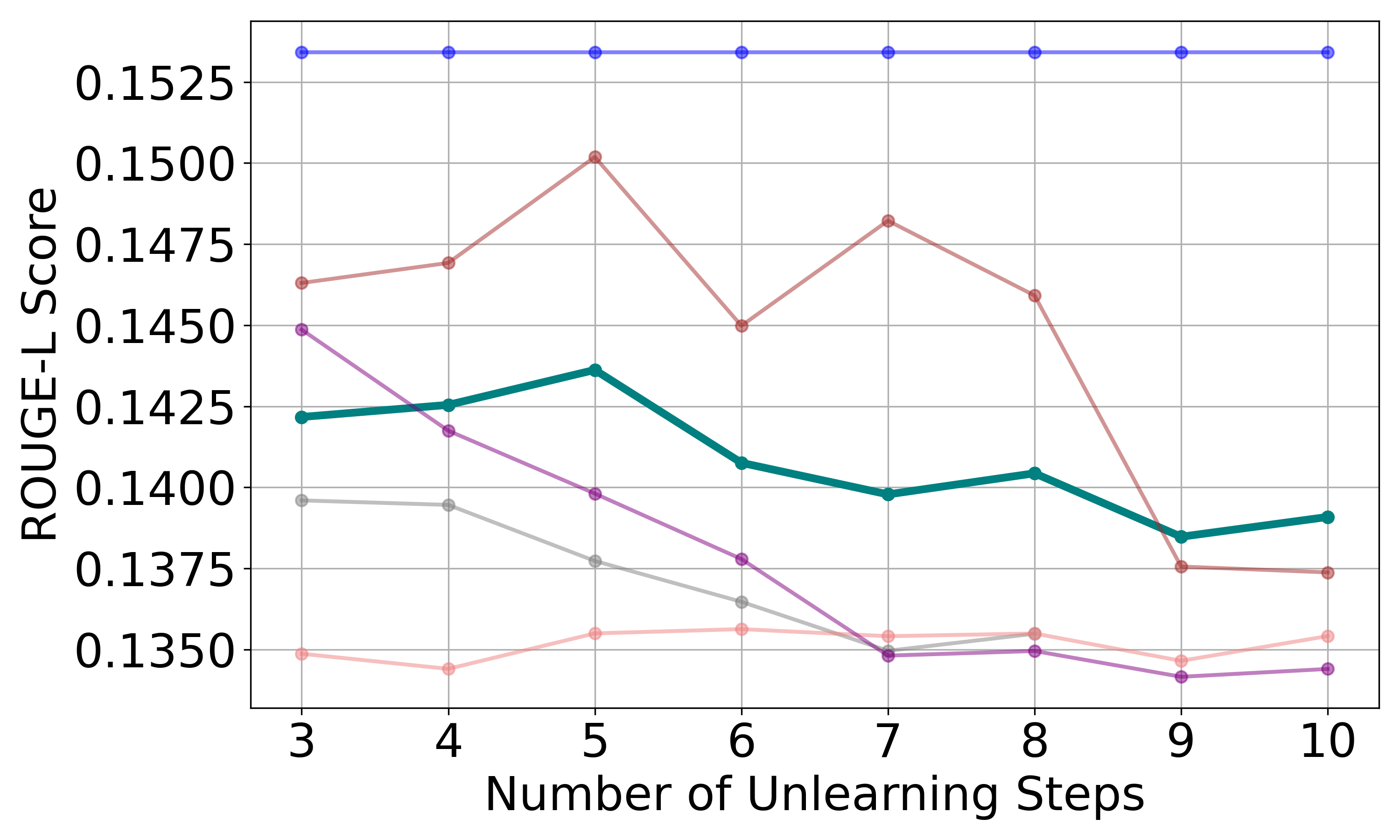}
        \subcaption{ROUGE-L Score for Book 3}
        \label{fig:rouge_l_book_3_individual}
        \end{subfigure} 
        \begin{subfigure}{0.48\textwidth}
        \includegraphics[width=\textwidth]{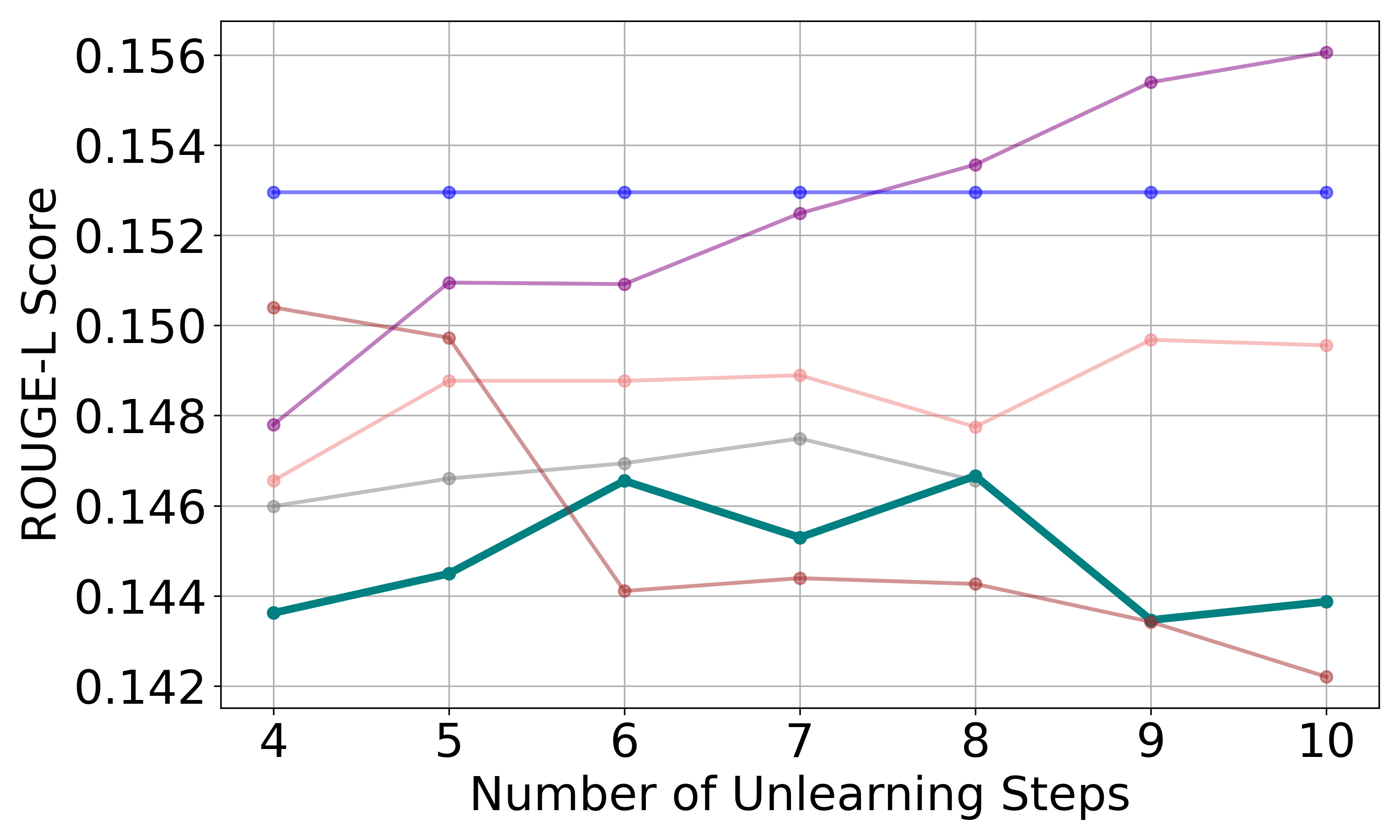}
        \subcaption{ROUGE-L Score for Book 4}
        \label{fig:rouge_l_book_4_individual}
        \end{subfigure} 
        
      \begin{subfigure}{0.48\textwidth}
        \includegraphics[width=\textwidth]{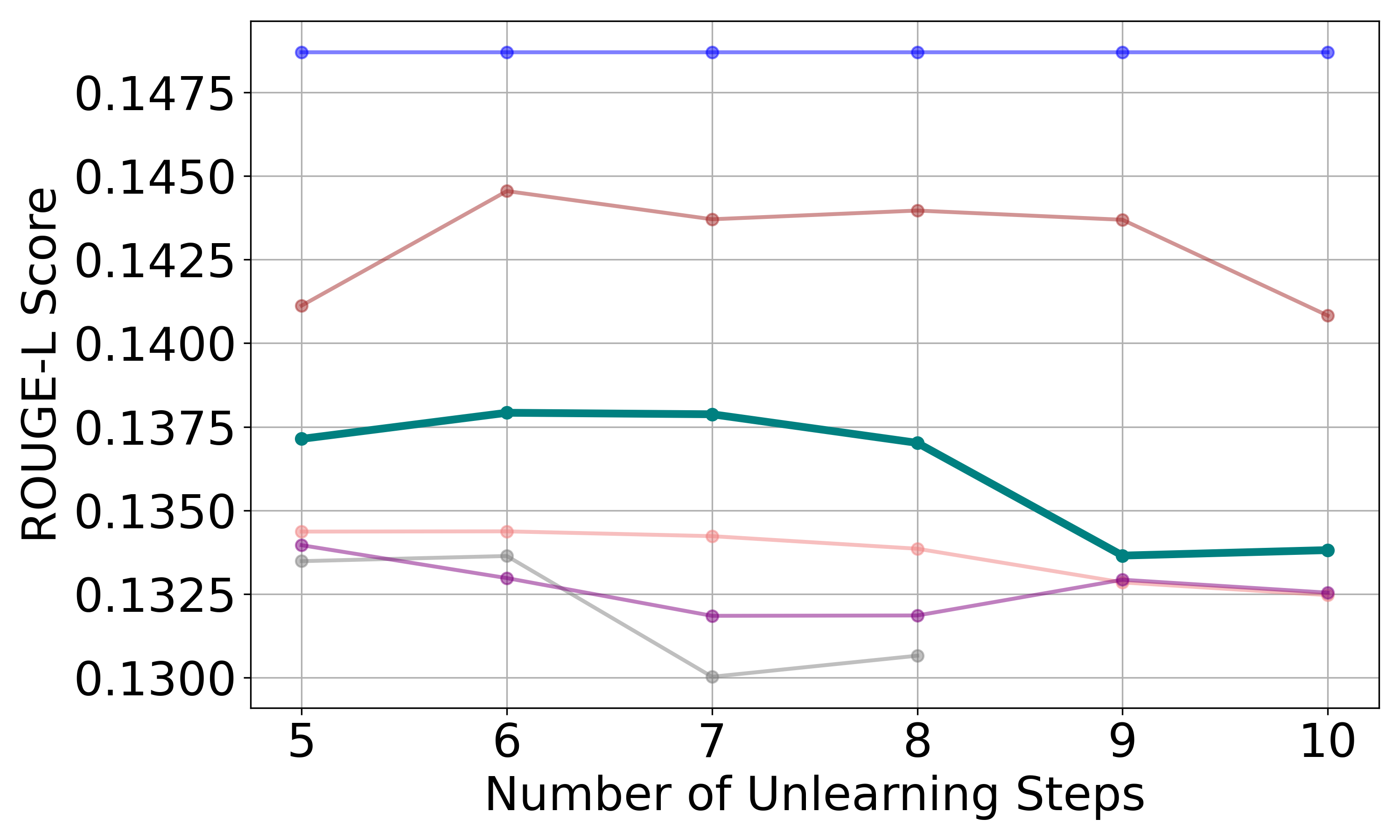}
        \subcaption{ROUGE-L Score for Book 5}
        \label{fig:rouge_l_book_5_individual}
        \end{subfigure} 
        \begin{subfigure}{0.48\textwidth}
        \includegraphics[width=\textwidth]{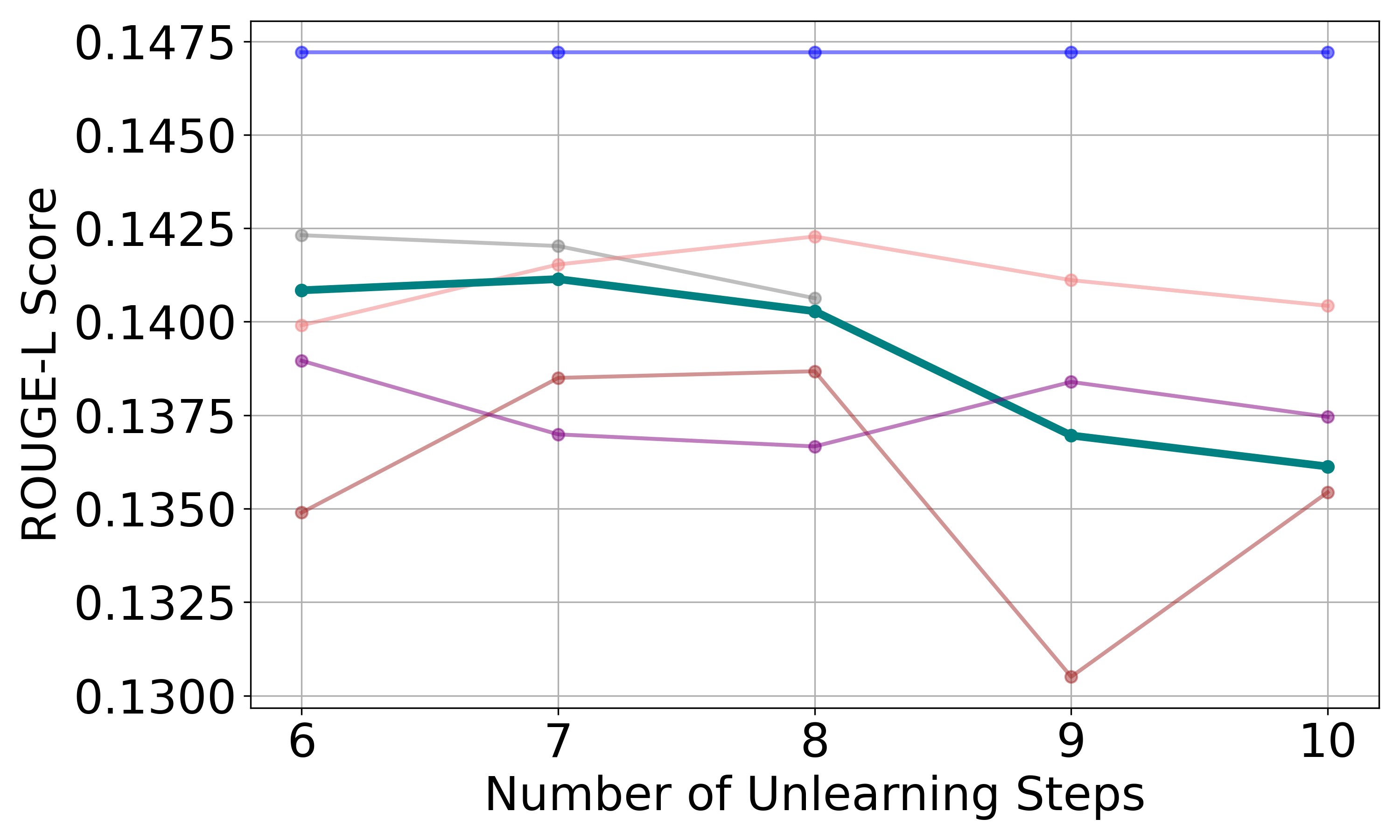}
        \subcaption{ROUGE-L Score for Book 6}
        \label{fig:rouge_l_book_6_individual}
        \end{subfigure} 

     \begin{subfigure}{0.48\textwidth}
        \includegraphics[width=\textwidth]{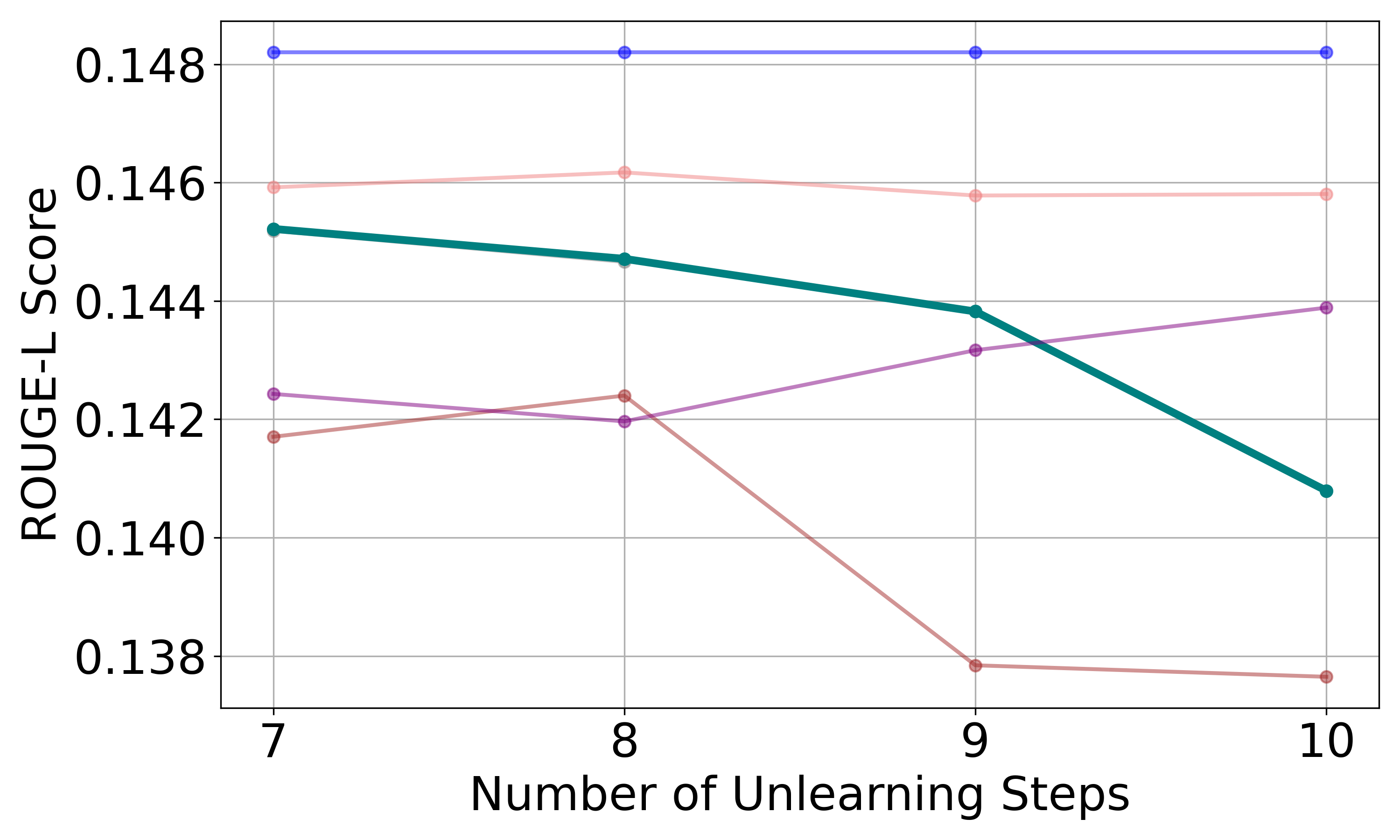}
        \subcaption{ROUGE-L Score for Book 7}
        \label{fig:rouge_l_book_7_individual}
        \end{subfigure} 
        \begin{subfigure}{0.48\textwidth}
        \includegraphics[width=\textwidth]{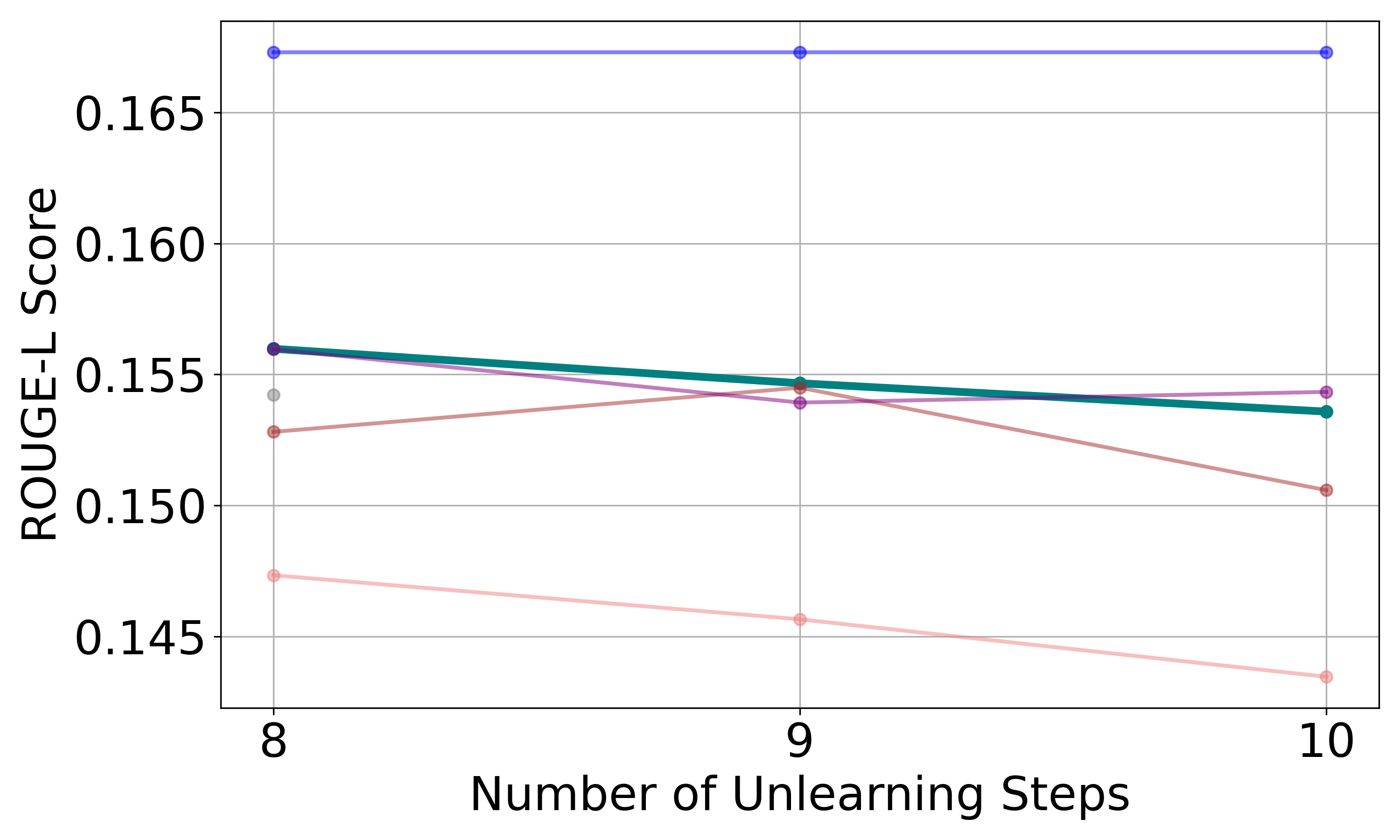}
        \subcaption{ROUGE-L Score for Book 8}
        \label{fig:rouge_l_book_8_individual}
        \end{subfigure} 
\caption{ROUGE-L score for each previously unlearned book, evaluated using various unlearning methods on the Llama 3.1. The x-axis represents the number of time steps (i.e., the number of sequential unlearning operations performed), while the y-axis indicates the ROUGE-L score. Different unlearning methods are compared, with ‘Vanilla’ serving as the baseline model without unlearning. We omit certain time steps for some methods because of the catastrophic collapse.}
\label{fig:seperate_unlearning_eval_llama}
\end{figure*}

\begin{figure*}
\centering
         \begin{subfigure}[b]{\textwidth}
            \centering
            \includegraphics[width=0.8\textwidth]{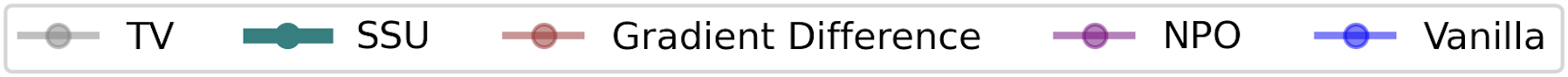}
          \end{subfigure}
        \begin{subfigure}{0.48\textwidth}
        \includegraphics[width=\textwidth]{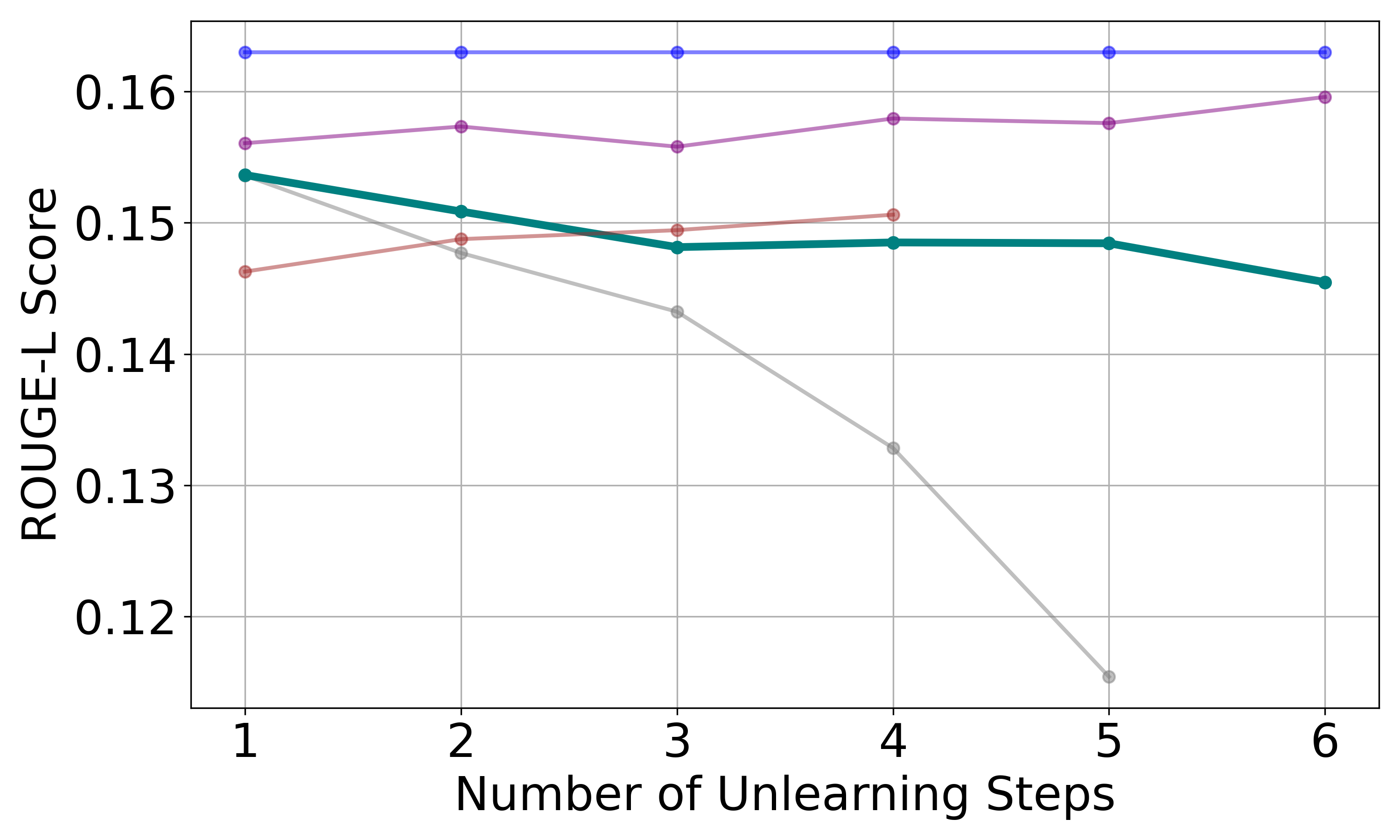}
        \subcaption{ROUGE-L Score for Book 1}
        \label{fig:rouge_l_book_1_individual_mistral}
        \end{subfigure}
        \begin{subfigure}{0.48\textwidth}
        \includegraphics[width=\textwidth]{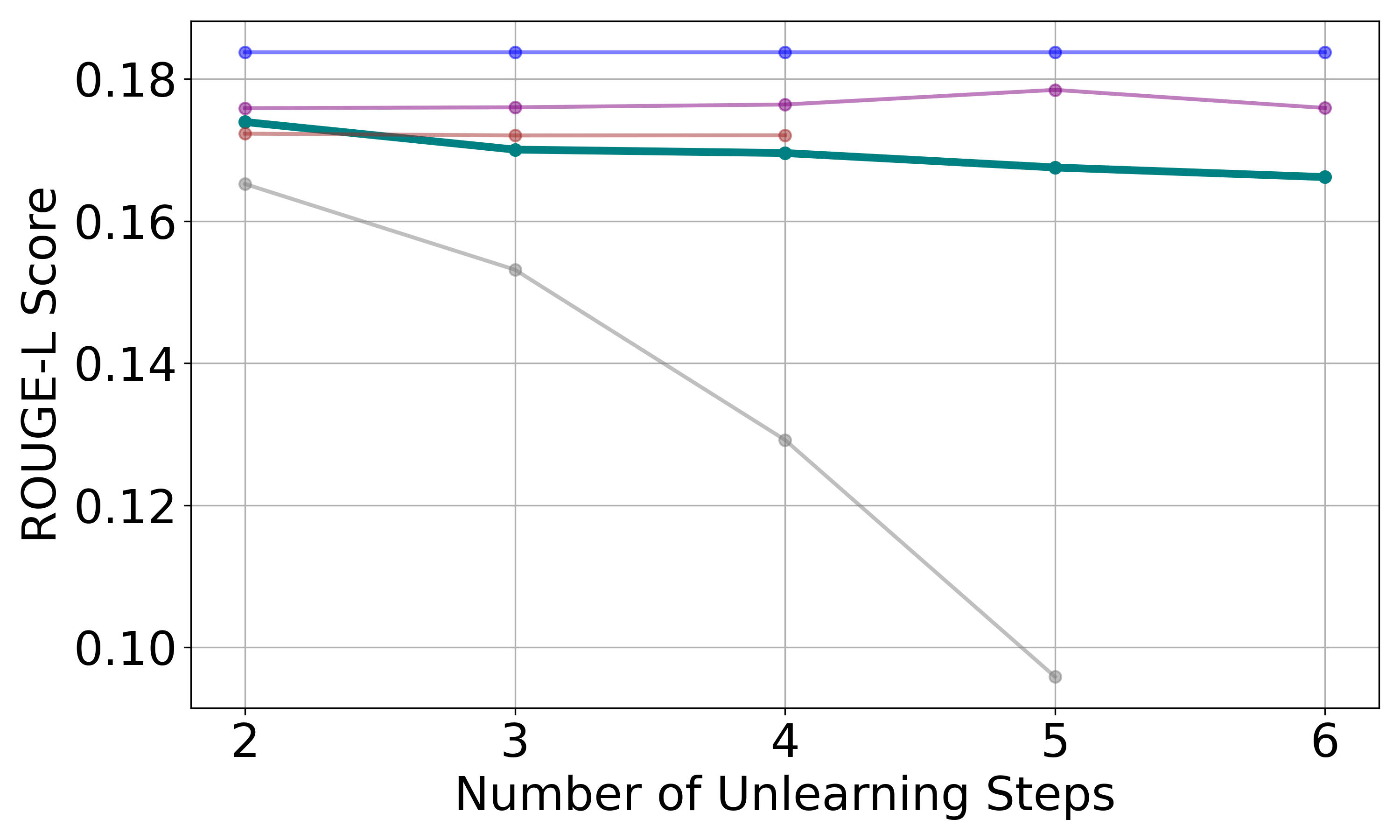}
        \subcaption{ROUGE-L Score for Book 2}
        \label{fig:rouge_l_book_2_individual_mistral}
        \end{subfigure}

       \begin{subfigure}{0.48\textwidth}
        \includegraphics[width=\textwidth]{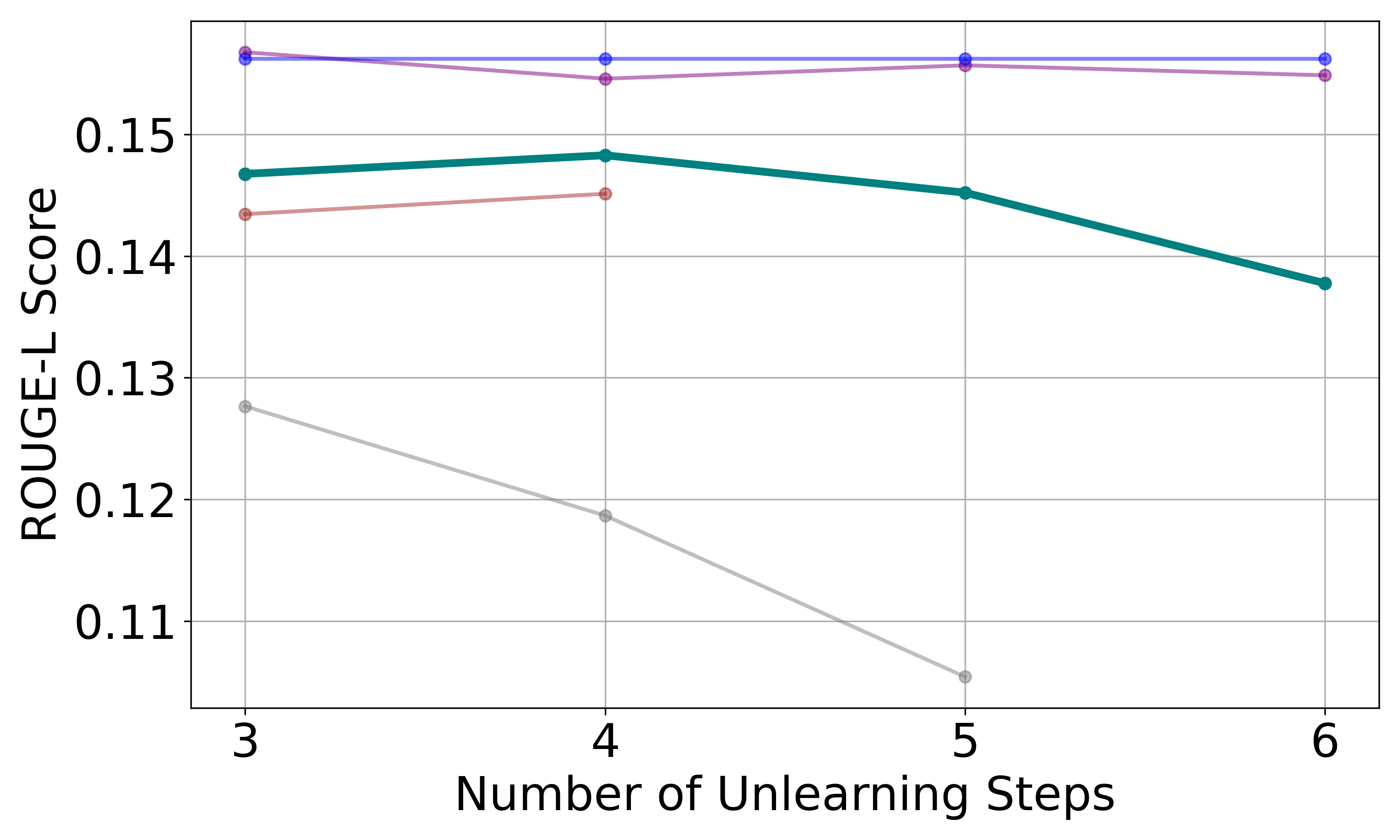}
        \subcaption{ROUGE-L Score for Book 3}
        \label{fig:rouge_l_book_3_individual_mistral}
        \end{subfigure} 
        \begin{subfigure}{0.48\textwidth}
        \includegraphics[width=\textwidth]{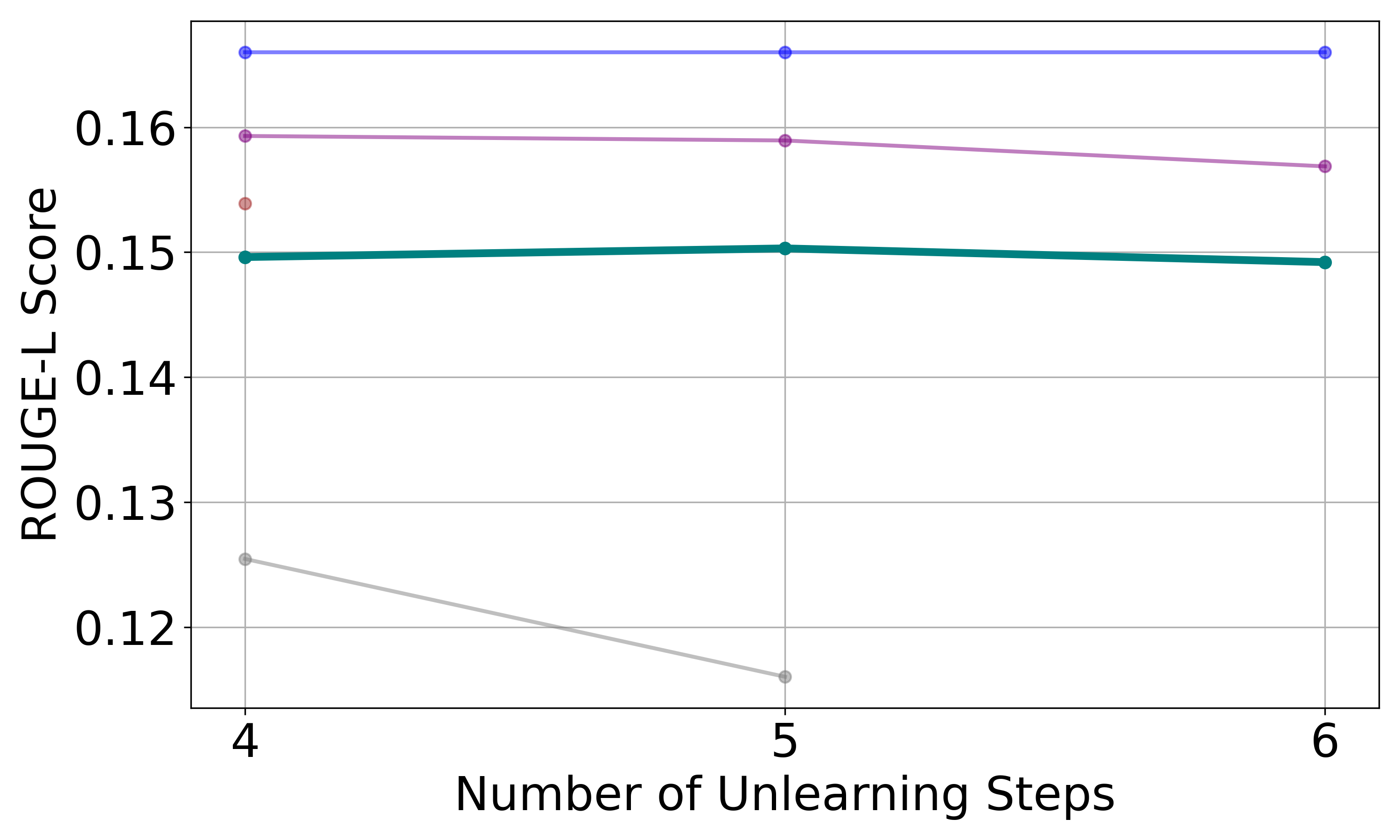}
        \subcaption{ROUGE-L Score for Book 4}
        \label{fig:rouge_l_book_4_individual_mistral}
        \end{subfigure} 
        
      \begin{subfigure}{0.48\textwidth}
        \includegraphics[width=\textwidth]{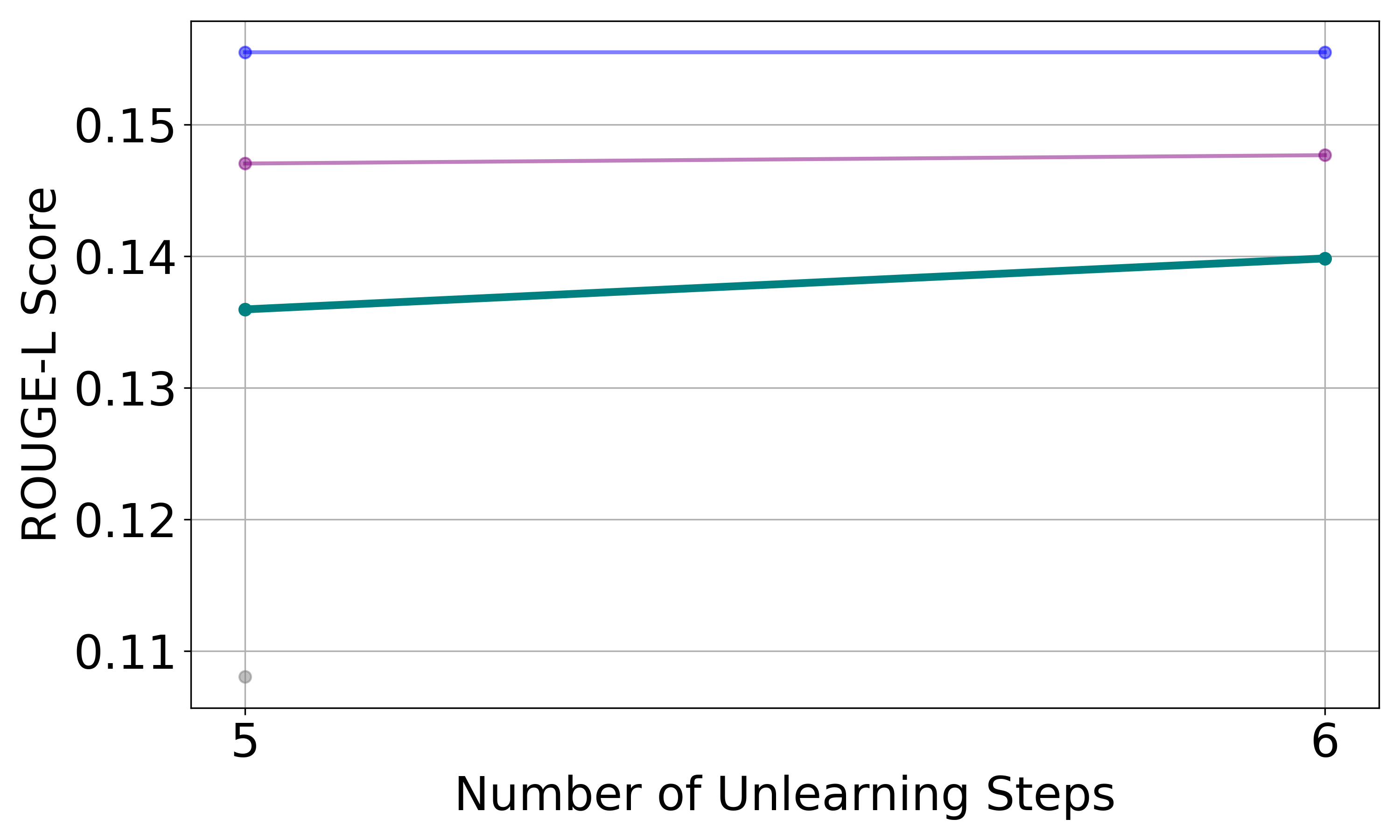}
        \subcaption{ROUGE-L Score for Book 5}
        \label{fig:rouge_l_book_5_individual_mistral}
        \end{subfigure} 
        \begin{subfigure}{0.48\textwidth}
        \includegraphics[width=\textwidth]{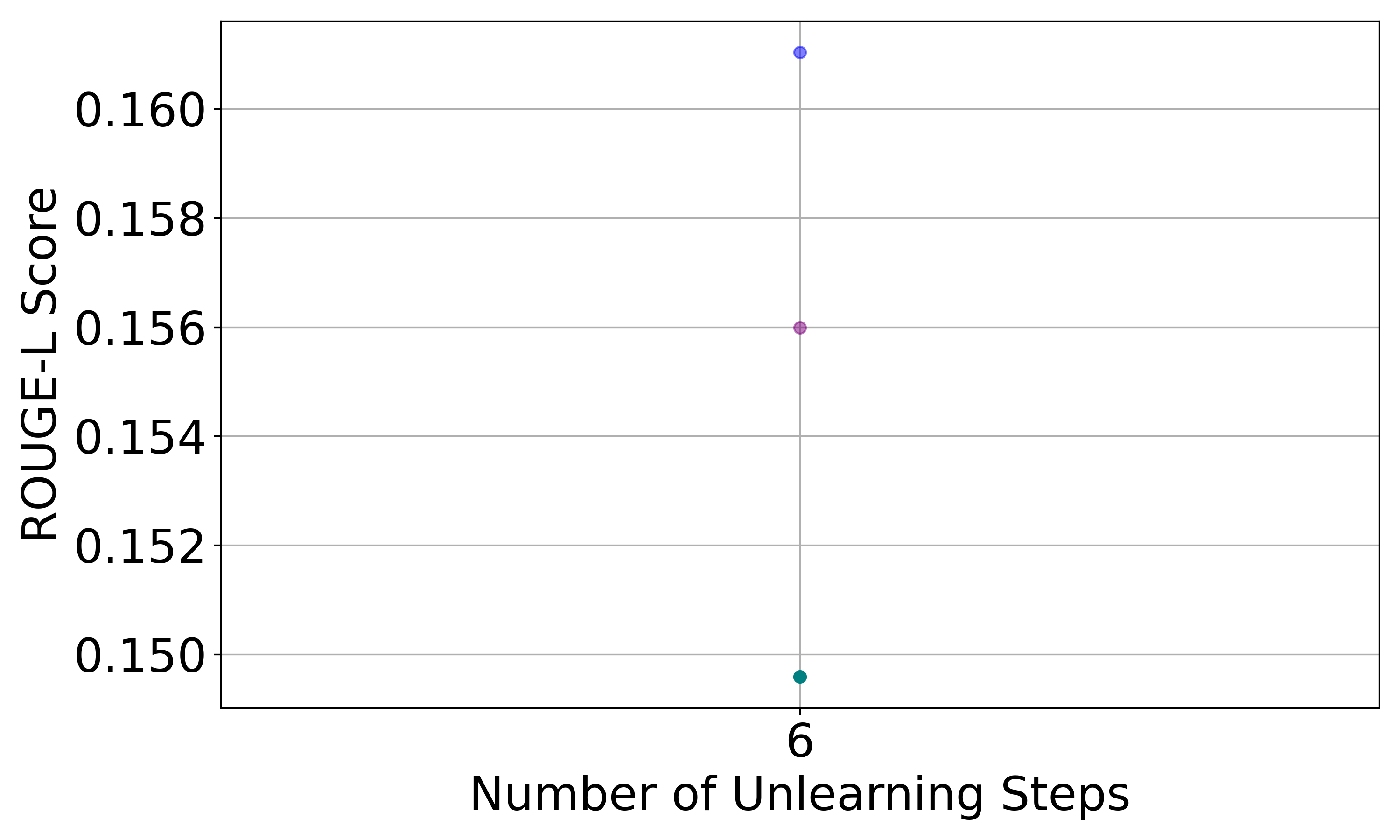}
        \subcaption{ROUGE-L Score for Book 6}
        \label{fig:rouge_l_book_6_individual_mistral}
        \end{subfigure} 

\caption{ROUGE-L score for each previously unlearned book, evaluated using various unlearning methods on the Mistral model. The x-axis represents the number of time steps (i.e., the number of sequential unlearning operations performed), while the y-axis indicates the ROUGE-L score. Different unlearning methods are compared, with ‘Vanilla’ serving as the baseline model without unlearning. We omit certain time steps for some methods because of the catastrophic collapse.}
\label{fig:seperate_unlearning_eval_mistral}
\end{figure*}

Figures \ref{fig:seperate_unlearning_eval_llama} and \ref{fig:seperate_unlearning_eval_mistral} present the results for Llama 3.1 and Mistral-7B, respectively. The $x$-axis denotes the number of unlearning steps performed, while the $y$-axis represents the ROUGE-L score for each individual book. Notably, in certain future time steps, the ROUGE-L score of previously unlearned books increases. For instance, as shown in Figure \ref{fig:rouge_l_book_4_individual}, the ROUGE-L score of the fourth book under NPO rises as the model unlearns additional books. Similar trends appear in Figure \ref{fig:rouge_l_book_1_individual}, where NPO’s score for the first book increases at time step 9, and in Figure \ref{fig:rouge_l_book_1_individual_mistral}, where the Mistral model exhibits a steady increase in NPO’s score as it unlearns other books over time.

We also observe instability in the gradient difference, particularly in Figure \ref{fig:seperate_unlearning_eval_llama}, where performance on previously unlearned books fluctuates significantly across future time steps. Additionally, \method exhibits some degree of reappearance of previously unlearned knowledge. This aligns with \citet{lucki2024adversarial}, in which it pointed out that fine-tuning on data unrelated to $D_f$ can inadvertently recover unlearned knowledge. Since all baselines in our experiments rely on fine-tuning with unlearning objectives, sequential unlearning fine-tunes the model on unrelated data—books unlearned at later time steps are independent of those unlearned earlier. This highlights a key challenge in sequential unlearning: continual model modifications may lead to knowledge re-emergence, highlighting the need for more robust unlearning algorithms in sequential settings.

\section{Full Experiment Results }
\label{sec:appendix-full-mistral-llama-results}

\begin{figure*}
\centering
         \begin{subfigure}[b]{\textwidth}
            \centering
            \includegraphics[width=1\textwidth]{Figure/save_figure/legend_plot.png}
          \end{subfigure}
        \begin{subfigure}{0.48\textwidth}
        \includegraphics[width=\textwidth]{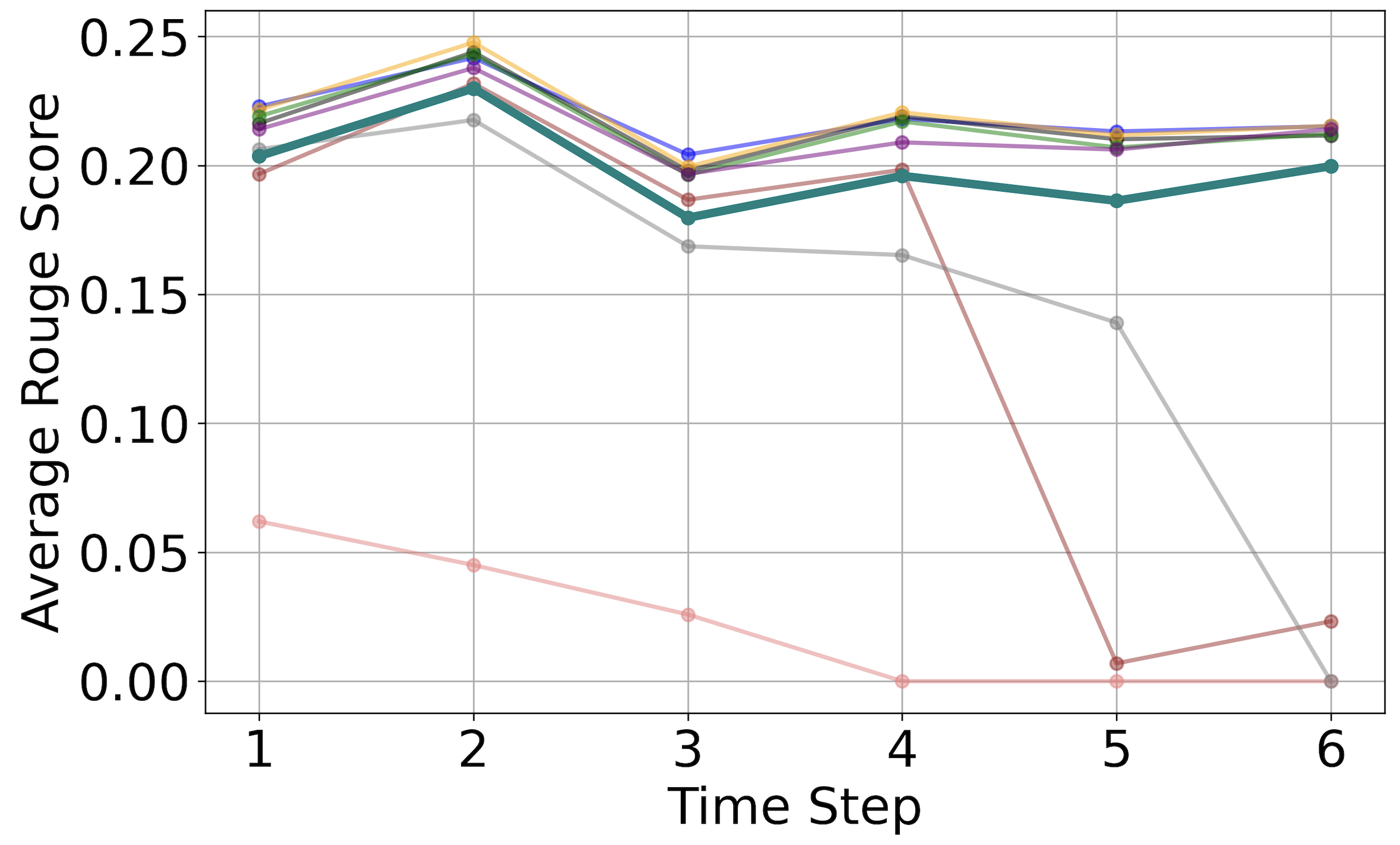}
        \subcaption{Avg Rouge Score on $D_f$}
        \label{fig:main_book_forget_unlearn_mistral}
        \end{subfigure}
        \begin{subfigure}{0.48\textwidth}
        \includegraphics[width=\textwidth]{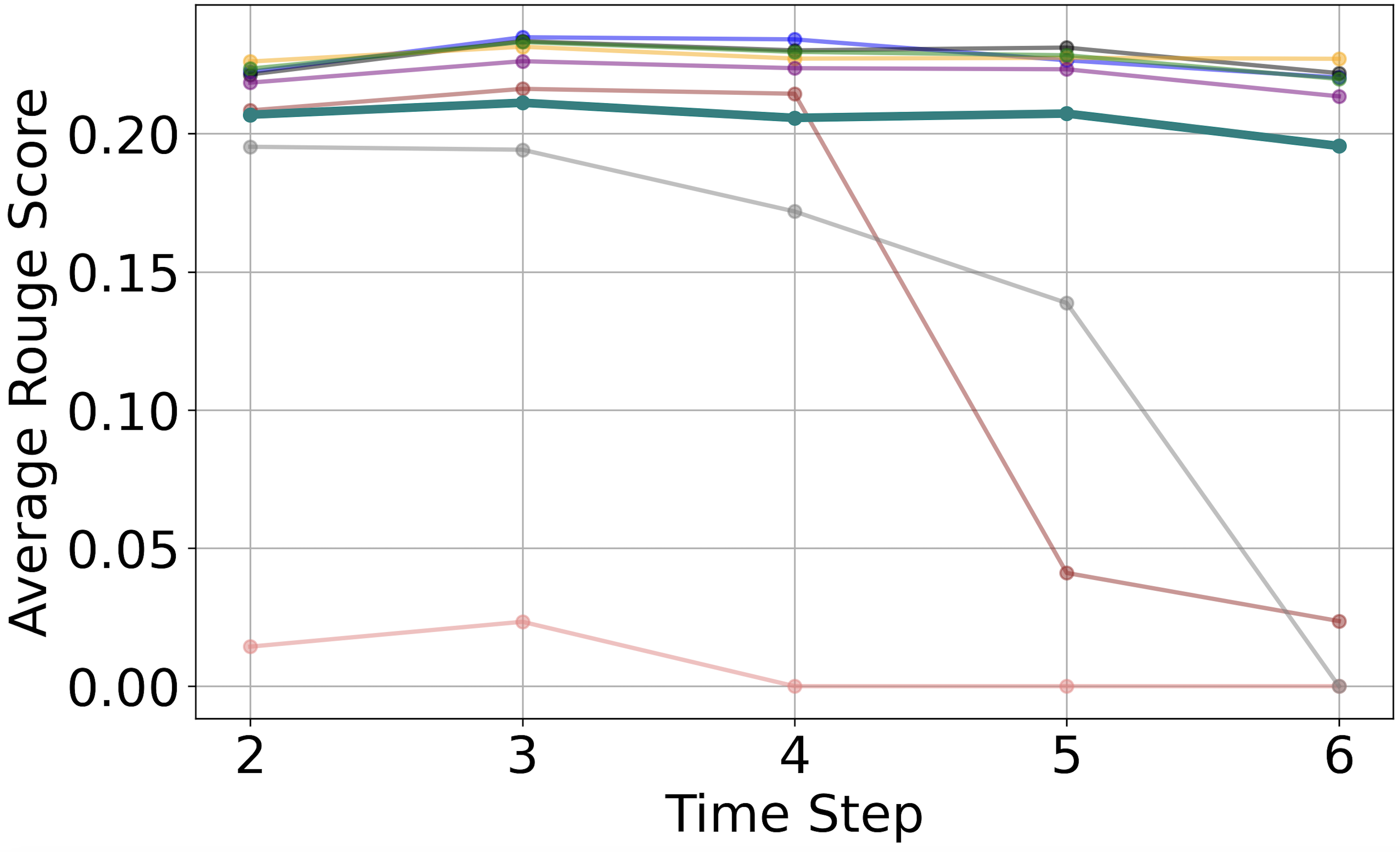}
        \subcaption{Avg Rouge Score on $D_{prev}$}
        \label{fig:main_book_forget_previous_mistral}
        \end{subfigure} 

        \begin{subfigure}{0.49\textwidth}
        \includegraphics[width=\textwidth]{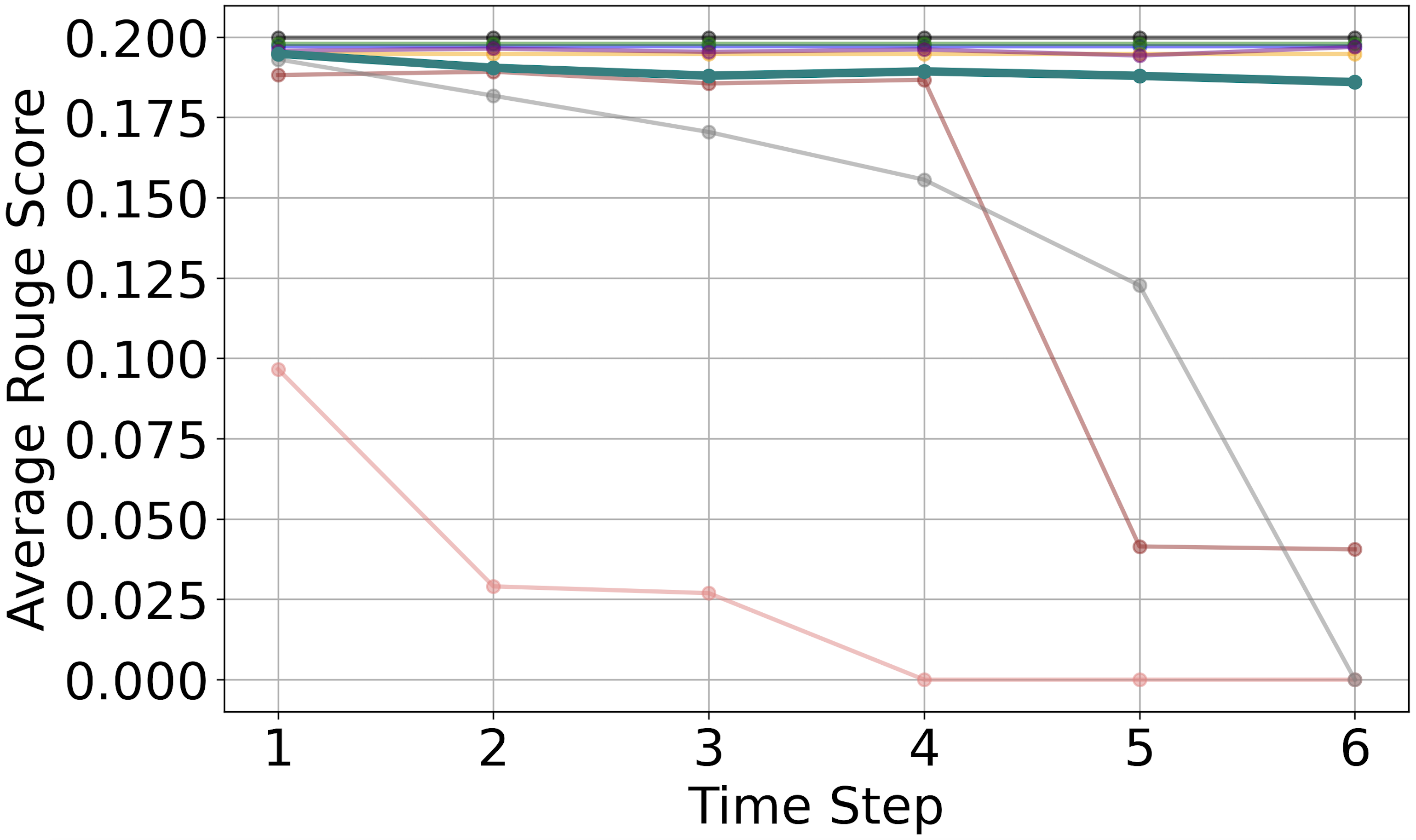}
        \subcaption{Avg Rouge Score on $D_n$}
        \label{fig:main_book_forget_normal_mistral}
        \end{subfigure}
        \begin{subfigure}{0.47\textwidth}
        \includegraphics[width=\textwidth]{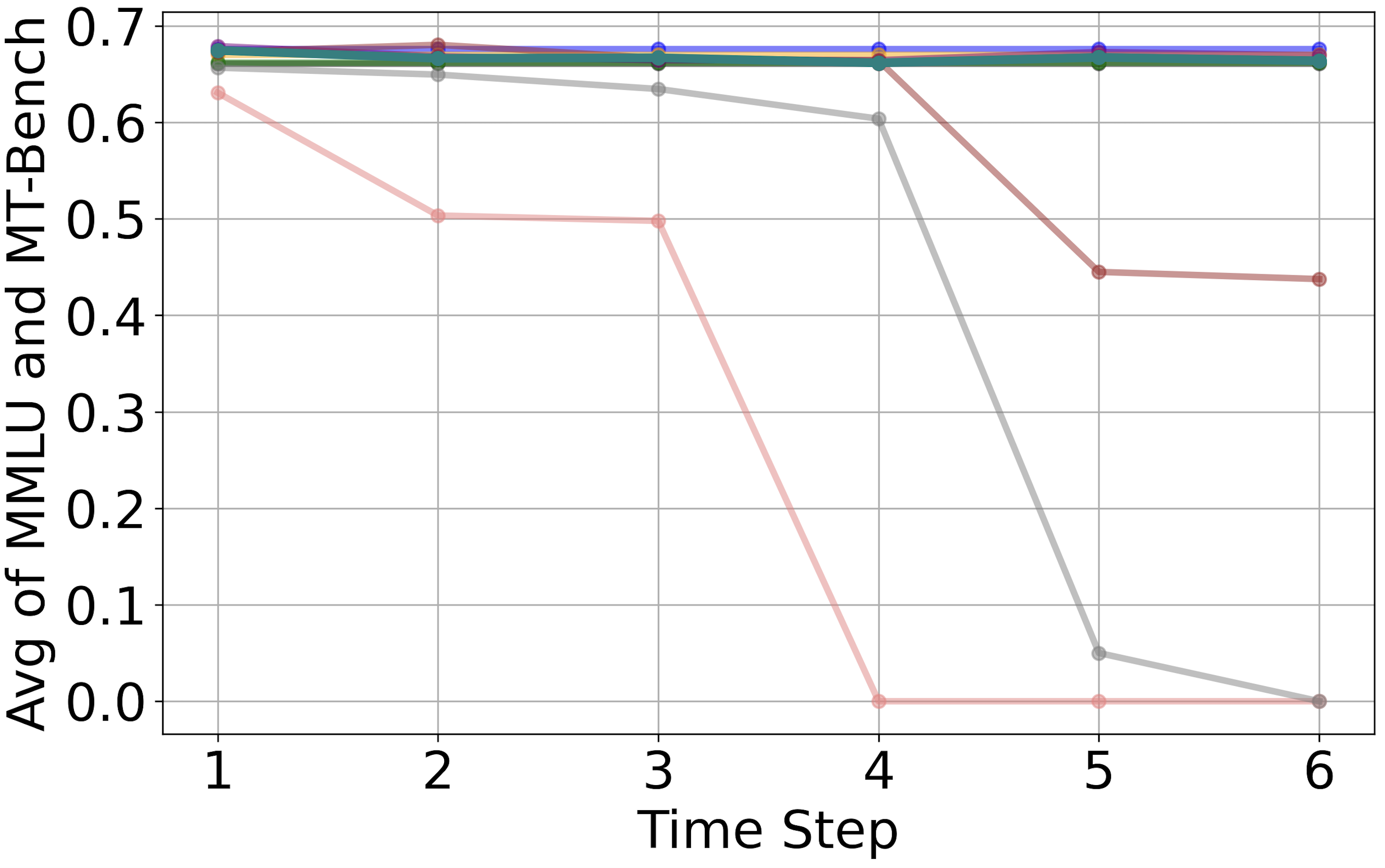}
        \subcaption{Avg MMLU and MT-Bench score}
        \label{fig:main_book_forget_general_mistral}
        \end{subfigure}
        
\caption{The average of Rouge-1 and Rouge-l score and reasoning abilities for Mistral-7B-Instruct: (a) books to forget $D_f$ ($\downarrow$); (b) previously unlearned books $D_{prev}$ ($\downarrow$); (c) $D_{nor}$ ($\uparrow$). and (d) averaged normalized MMLU and MT-Bench scores ($\uparrow$). The results for TV after time step 8 are omitted due to collapse. Lower Rouge scores for $D_f$ and $D_{prev}$ indicate better unlearning, while higher scores for $D_{nor}$ and benchmarks reflect better performance.}
\label{fig:main_book_forget_all_mistral}
\vspace{-0.1in}
\end{figure*}

\begin{figure*}
\centering
         \begin{subfigure}[b]{\textwidth}
            \centering
            \includegraphics[width=1\textwidth]{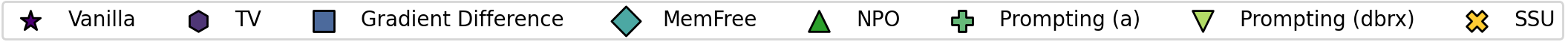}
          \end{subfigure}
        \begin{subfigure}{0.44\textwidth}
        \includegraphics[width=\textwidth]{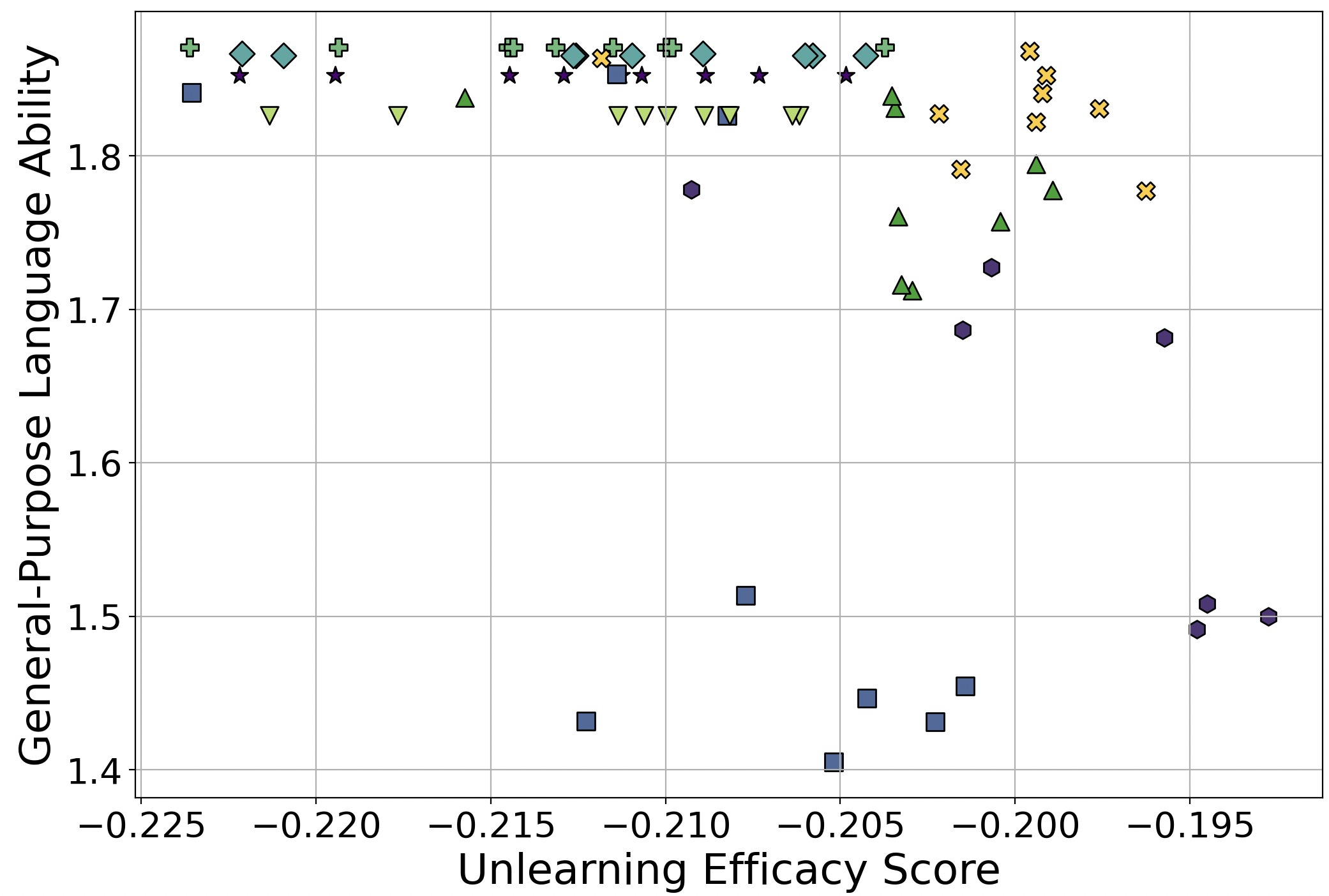}
        \subcaption{Llama3.1-8B-Instruct Trade-off}
        \label{fig:trade_off_llama3.1}
        \end{subfigure}
        \begin{subfigure}{0.45\textwidth}
        \includegraphics[width=\textwidth]{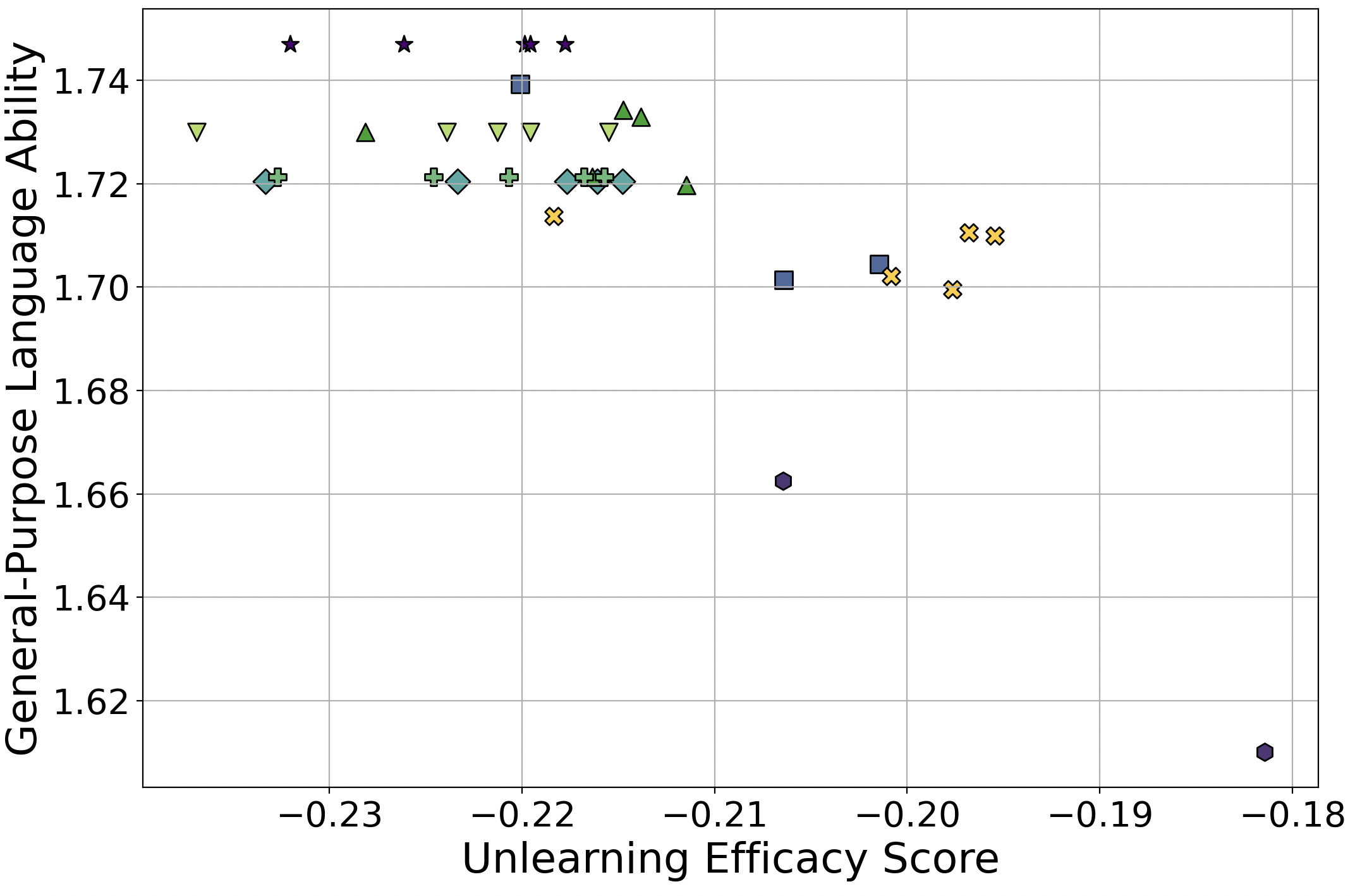}
        \subcaption{Mistral-7B-Instruct Trade-off}
        \label{fig:trade_off_mistral}
        \end{subfigure} 
\caption{Trade-off between general-purpose language abilities and unlearning efficacy for Llama3.1 and Mistral-7B, including all methods, except TV beyond time step 9 (Llama3.1) and time step 3 (Mistral-7B), and Gradient Difference beyond time step 4 (Mistral-7B), as they all collapsed during these time steps. General-purpose abilities are represented by the average of MMLU and MT-Bench scores, normalized. Unlearning efficacy is measured as the average of Rouge-1 and Rouge-L scores on $D_f$ and $D_{prev}$, where lower Rouge scores indicate better unlearning performance; thus, values were negated for clarity. The ideal performance is positioned in the top-right corner. The plots capture the performance of all methods at every time step greater than 1.}
\vspace{-0.2in}
\label{fig:trade_off_appendix}
\end{figure*}

\begin{figure*}[btp]
\centering
         \begin{subfigure}[b]{\textwidth}
            \centering
            \includegraphics[width=0.45\textwidth]{Figure/ablation/legend_ablation_plog.png}
          \end{subfigure}
        \begin{subfigure}{0.24\textwidth}
        \includegraphics[width=\textwidth]{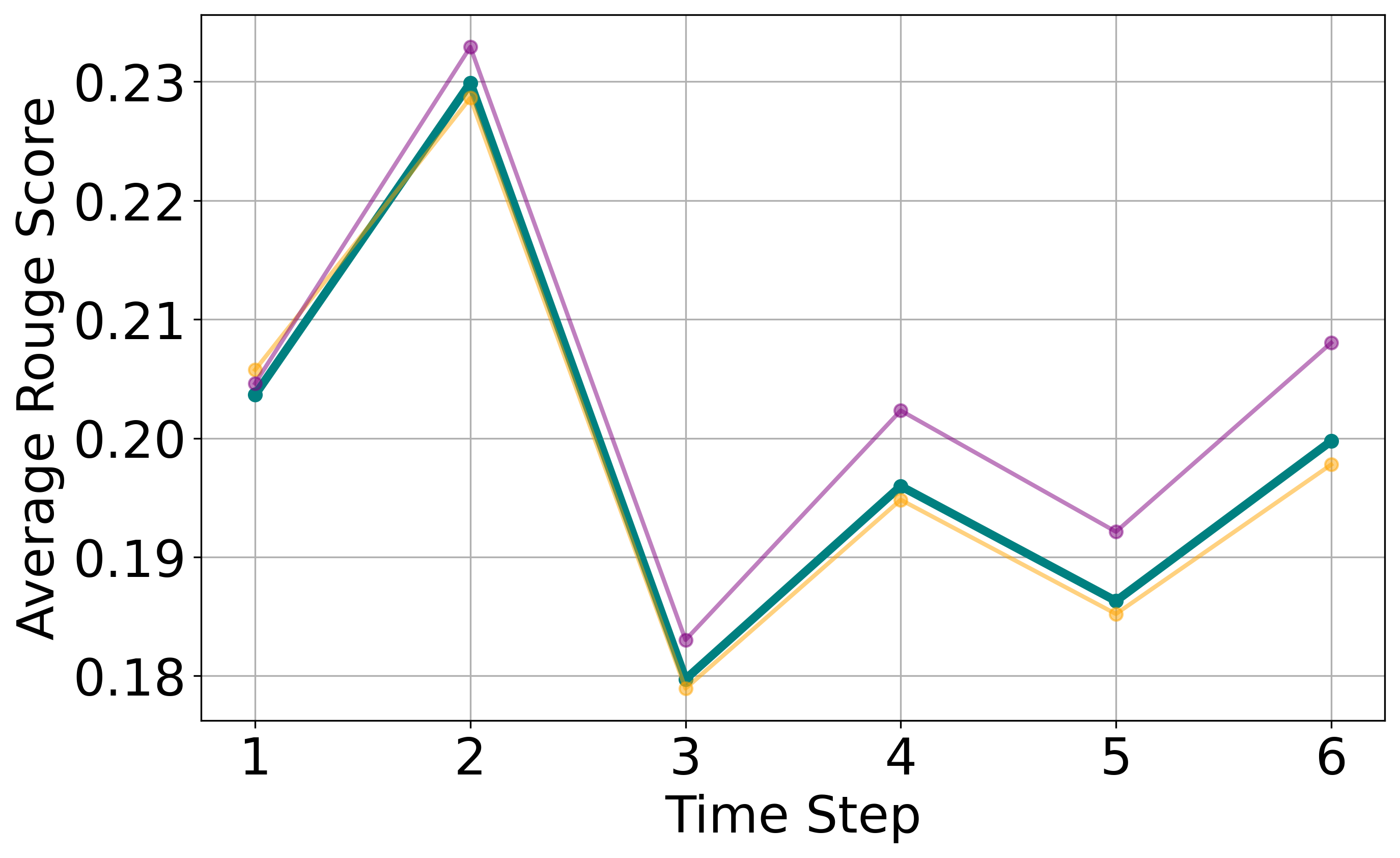}
        \subcaption{Avg Rouge on $D_f$}
        \label{fig:ablation_book_forget_mistral}
        \end{subfigure}
        \begin{subfigure}{0.24\textwidth}
        \includegraphics[width=\textwidth]{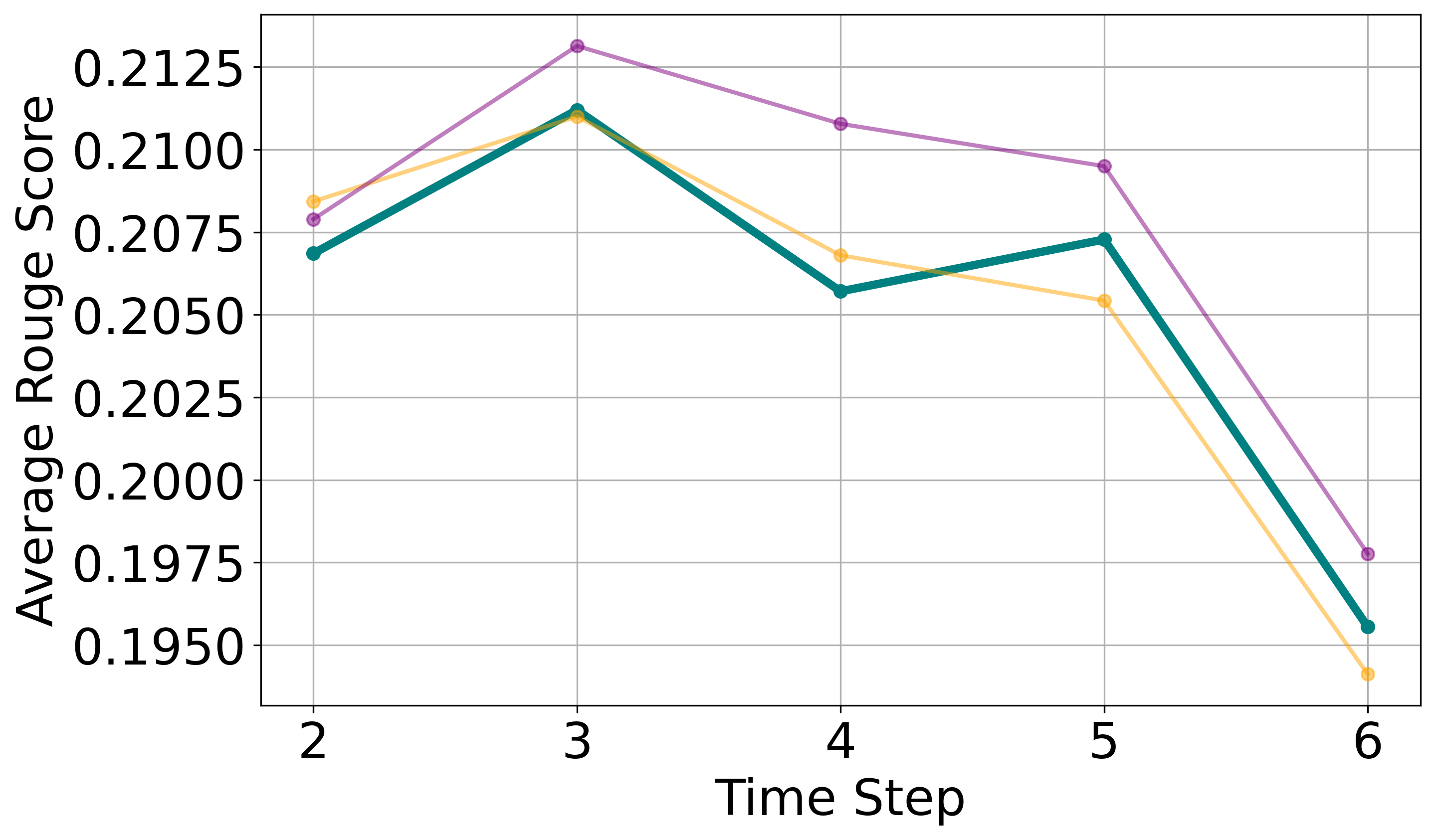}
        \subcaption{Avg Rouge Score on $D_{prev}$}
        \label{fig:ablation_book_prev_mistral}
        \end{subfigure}
        \begin{subfigure}{0.24\textwidth}
        \includegraphics[width=\textwidth]{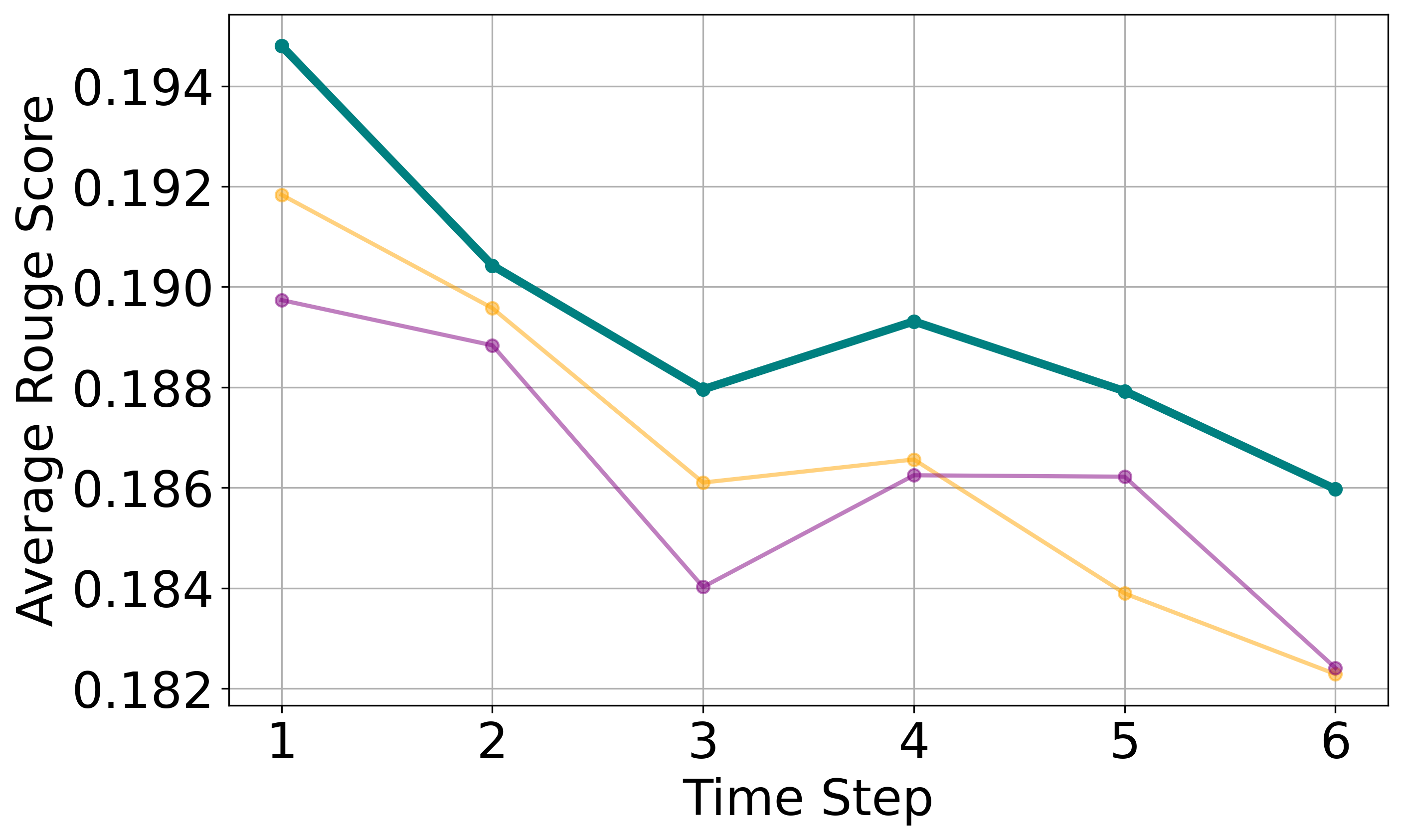}
        \subcaption{Avg Rouge Score on $D_n$}
        \label{fig:ablation_book_norm_mistral)}
        \end{subfigure}
        \begin{subfigure}{0.24\textwidth}
        \includegraphics[width=\textwidth]{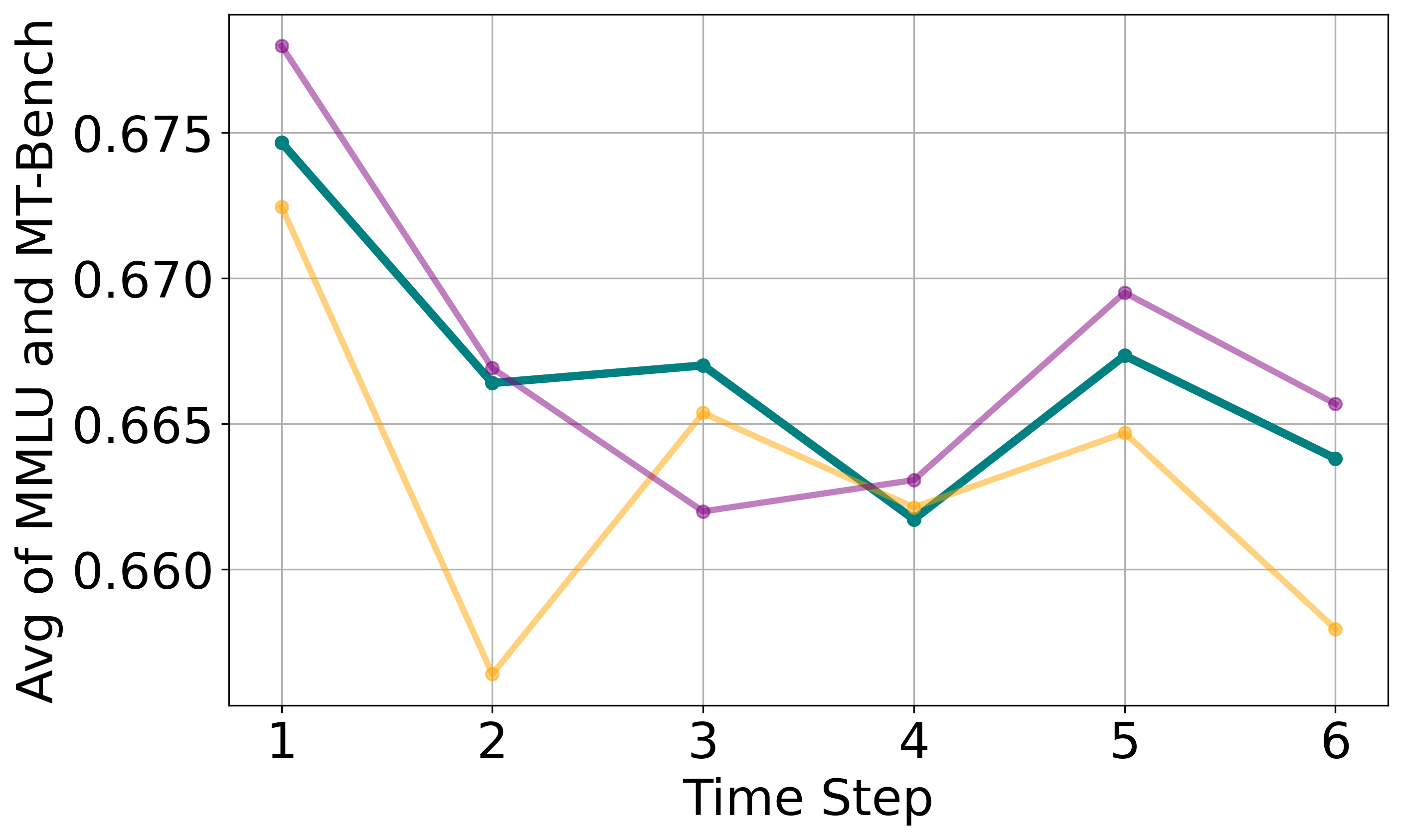}
        \subcaption{General Performance}
        \label{fig:ablation_general_mistral)}
        \end{subfigure}
\caption{Ablation study of \method for Mistral-7B-Instruct-v0.3. The orange line represents unlearning without the weight saliency map, while the purple line shows the effect of removing the random labeling loss.}
\label{fig:ablation_mistral}
\end{figure*}

Here we present the performance of \method compared to all other baseline methods, using Llama3.1-8B-Instruct in Figure \ref{fig:main_book_forget_all_appendix} and Mistral-7B-Instruct in Figure \ref{fig:main_book_forget_all_mistral}.

Additionally, we present trade-off comparisons of all the baseline methods except GA (GA is collapsed at most time steps in Mistral-7B-Instruct) in Figure \ref{fig:trade_off_appendix}. 
Moreover, the x-axis used in Figure \ref{fig:trade_off_appendix} is defined as:

\begin{equation*}
\text{Unlearning Efficacy} = U_f + U_{\text{prev}}
\end{equation*}
where 
\begin{equation*}
    U_f = -\frac{1}{2} \left( R_1^f + R_L^f \right)
\end{equation*}
 
  \begin{equation*}
 U_{\text{prev}} = -\frac{1}{2} \left( R_1^{\text{prev}} + R_L^{\text{prev}} \right).
 \end{equation*}
Here, $R_1^f$ and $R_L^f$ represent the Rouge-1 and Rouge-L scores for $D_f$, and $R_1^{\text{prev}}$ and $R_L^{\text{prev}}$ represent the Rouge-1 and Rouge-L scores for $D_{prev}$. The y-axis (General-purpose Language Ability) is defined as:
\vspace{-0.15in}
\begin{equation*}
 \frac{1}{2} \left( M + \frac{B}{10} \right)
\end{equation*}
where $M$ and $B$ represent the MMLU and MT-Bench scores, respectively.
We also present the ablation study resuls for Mistral-7B-Instruct-v0.3 in Figure \ref{fig:ablation_mistral}. 

\section{Numeric Experiment Results}
\label{sec:appendix-full_experiment_numbers}

In this section, we present our experimental results numerically. Tables \ref{tab:main_results_time_step_1} to \ref{tab:main_results_time_step_10} display the unlearning results of Llama3.1 across all ten time steps. Tables \ref{tab:main_results_mistral_time_step_1} to \ref{tab:main_results_mistral_time_step_6} show the results for Mistral-7B unlearning across all six time steps. In sections \ref{section:appendix-collapse-llama3.1} and \ref{section:appendix-collapse-mistral7B} we illustrate how GA, TV, and Gradient Difference encounter catastrophic collapse. 

\subsection{Catastrophic Collapse of Llama3.1}
\label{section:appendix-collapse-llama3.1}

\textbf{GA, TV, and Gradient Difference experience varying levels of model collapse during the sequential unlearning process.} 
As shown in table \ref{tab:main_results_time_step_1}, GA begins with an MMLU of 0.5821. By time step 5, GA's MMLU has dropped to 0.3102 \ref{tab:main_results_time_step_5}, demonstrating a rapid degradation in general reasoning. Similarly, TV starts with an MMLU of 0.6621 at time step 1 and undergoes a steep decline in reasoning ability, dropping to 0.4887 by time step 5, and reaching an MMLU of 0 at time step 10 (as shown in tables \ref{tab:main_results_time_step_5} and \ref{tab:main_results_time_step_10}). Additionally, Gradient Difference also faces catastrophic collapse at time step 5. Specifically, its MT-Bench score falls from 8.034 at time step 4 (table \ref{tab:main_results_time_step_4}) to 4.9438 at time step 5, and further declines to 4.48 at time step 10, though its MMLU score remains stable. Lastly, NPO and \method exhibit a gradual decline in general-purpose language abilities, but \method consistently outperforms NPO across all the time steps. 

\subsection{Catastrophic Collapse of Mistral-7B}
\label{section:appendix-collapse-mistral7B}
\textbf{GA and TV experience rapid model collapse, while Gradient Difference still suffers loss of conversational ability.} 
As shown in table \ref{tab:main_results_mistral_time_step_1}, GA collapses at the first time step. TV initially declines gradually but undergoes a sudden reasoning degradation at time step 5 (table \ref{tab:main_results_mistral_time_step_4}), where both MMLU and MT-Bench scores drop to 0. Gradient Difference experiences a sharp decrease in MT-Bench score from 7.2375 at time step 4 to 2.8375 at time step 5, eventually dropping to 2.6815 at time step 6. Similar to the Llama3.1 case, Gradient Difference maintains stable performance on MMLU. Lastly, NPO maintains competitive general-purpose abilities scores compared to \method, Figures \ref{fig:main_book_forget_all_mistral} and \ref{fig:trade_off_appendix} show that \method demonstrates superior unlearning efficacy and achieves a better trade-off. 

\begin{table*}[h]
\centering
\resizebox{1\textwidth}{!}{
\begin{tabular}{l||ll||ll||ll||ll}
\hline
\multicolumn{1}{c||}{\multirow{2}{*}{}} & 
\multicolumn{2}{c||}{\textbf{$\mathbf{D_f}$}} & 
\multicolumn{2}{c||}{\textbf{$\mathbf{D_{prev}}$}} & 
\multicolumn{2}{c||}{\textbf{$\mathbf{D_{nor}}$}} &
\multicolumn{2}{c}{\textbf{Benchmark}} \\ \cline{2-9}

\multicolumn{1}{c||}{} &
\begin{tabular}[c]{@{}l@{}}Rouge-1 \\ \end{tabular} &
\begin{tabular}[c]{@{}l@{}}Rouge-L\\ \end{tabular} &
\begin{tabular}[c]{@{}l@{}}Rouge-1\\ \end{tabular} &
\begin{tabular}[c]{@{}l@{}}Rouge-L \\  \end{tabular} &
\begin{tabular}[c]{@{}l@{}}Rouge-1 \\ \end{tabular} &
\begin{tabular}[c]{@{}l@{}}Rouge-L \\  \end{tabular} &
\begin{tabular}[c]{@{}l@{}}MMLU \\ \end{tabular} &
\begin{tabular}[c]{@{}l@{}}MT-Bench \\ \end{tabular} 
\\ \hline
Vanilla &  0.2724 & 0.1530 & 0 & 0 & 0.2349 & 0.1380 & 0.6618 & 8.1808 \\
Prompting (a) & 0.2707 & 0.1541 & 0 & 0 & 0.2376 & 0.1364 & 0.6635 & 8.3344 \\
Prompting (dbrx) & 0.2730 & 0.1535 & 0 & 0 & 0.2333 & 0.1364 & 0.6611 & 7.9563 \\
MemFree Decode & 0.2711 & 0.1524 & 0 & 0 & 0.2392 & 0.1407 & 0.6618 & 8.2453 \\
GA & 0.2504 & 0.1430 & 0 & 0 & 0.2282 & 0.1354 & 0.5821 & 8.1719 \\
NPO & 0.2655 & 0.1487 & 0 & 0 & 0.2380 & 0.1411 & 0.6600 & 8.1938 \\
Gradiet Difference & 0.2619 & 0.1496 & 0 & 0 & 0.2433 & 0.1241 & 0.6544 & 8.0031 \\
TV & 0.2463 & 0.1433 & 0 & 0 & 0.2228 & 0.1331 & 0.6621 & 8.2038 \\
SSU & 0.2523 & 0.1432 &0 & 0 & 0.2299 & 0.1357 & 0.6625 & 8.2250 \\
\hline
\end{tabular}}
\caption{Overall results of Llama3.1 at time step 1, compared with several baselines for $D_f$, $D_{prev}$, and $D_{nor}$. Benchmark performance includes MMLU and MT-Bench scores.}
\vspace{-0.15in}
\label{tab:main_results_time_step_1}
\end{table*}

\begin{table*}[h]
\centering
\resizebox{1\textwidth}{!}{
\begin{tabular}{l||ll||ll||ll||ll}
\hline
\multicolumn{1}{c||}{\multirow{2}{*}{}} & 
\multicolumn{2}{c||}{\textbf{$\mathbf{D_f}$}} & 
\multicolumn{2}{c||}{\textbf{$\mathbf{D_{prev}}$}} & 
\multicolumn{2}{c||}{\textbf{$\mathbf{D_{nor}}$}} &
\multicolumn{2}{c}{\textbf{Benchmark}} \\ \cline{2-9}

\multicolumn{1}{c||}{} &
\begin{tabular}[c]{@{}l@{}}Rouge-1 \\ \end{tabular} &
\begin{tabular}[c]{@{}l@{}}Rouge-L\\ \end{tabular} &
\begin{tabular}[c]{@{}l@{}}Rouge-1\\ \end{tabular} &
\begin{tabular}[c]{@{}l@{}}Rouge-L \\  \end{tabular} &
\begin{tabular}[c]{@{}l@{}}Rouge-1 \\ \end{tabular} &
\begin{tabular}[c]{@{}l@{}}Rouge-L \\  \end{tabular} &
\begin{tabular}[c]{@{}l@{}}MMLU \\ \end{tabular} &
\begin{tabular}[c]{@{}l@{}}MT-Bench \\ \end{tabular} 
\\ \hline
Vanilla &  0.3027 & 0.1793 & 0.2586 & 0.1478 & 0.2349 & 0.1380 & 0.6618 & 8.1808 \\
Prompting (a) & 0.3036 & 0.1814 & 0.2605 & 0.1489 & 0.2376 & 0.1364 & 0.6635 & 8.3344 \\
Prompting (dbrx) & 0.2971 & 0.1759 & 0.2638 & 0.1485 & 0.2333 & 0.1364 & 0.6611 & 7.9563 \\
MemFree Decode & 0.2995 & 0.1781 & 0.2604 & 0.1501 & 0.2392 & 0.1407 & 0.6618 & 8.2453 \\
GA & 0.2755 & 0.1644 & 0.2489 & 0.1409 & 0.2287 & 0.1375 & 0.5157 & 8.1719 \\
NPO & 0.2910 & 0.1725 & 0.2556 & 0.1439 & 0.2368 & 0.1395 & 0.6512 & 8.1063 \\
Gradient Difference & 0.2996 & 0.1806 & 0.2605 & 0.1494 & 0.2401 & 0.1405 & 0.6544 & 8.0627 \\
TV & 0.2774 & 0.1674 & 0.2492 & 0.1433 & 0.2425 & 0.1431 & 0.5845 & 8.0808 \\
SSU & 0.2863 & 0.1702 & 0.2494 & 0.1414 & 0.2354 & 0.1390 & 0.6519 & 8.3769 \\
\hline
\end{tabular}}
\caption{Overall results of Llama3.1 at time step 2, compared with several baselines for $D_f$, $D_{prev}$, and $D_{nor}$. Benchmark performance includes MMLU and MT-Bench scores.}
\vspace{-0.15in}
\label{tab:main_results_time_step_2}
\end{table*}

\begin{table*}[h]
\centering
\resizebox{1\textwidth}{!}{
\begin{tabular}{l||ll||ll||ll||ll}
\hline
\multicolumn{1}{c||}{\multirow{2}{*}{}} & 
\multicolumn{2}{c||}{\textbf{$\mathbf{D_f}$}} & 
\multicolumn{2}{c||}{\textbf{$\mathbf{D_{prev}}$}} & 
\multicolumn{2}{c||}{\textbf{$\mathbf{D_{nor}}$}} &
\multicolumn{2}{c}{\textbf{Benchmark}} \\ \cline{2-9}

\multicolumn{1}{c||}{} &
\begin{tabular}[c]{@{}l@{}}Rouge-1 \\ \end{tabular} &
\begin{tabular}[c]{@{}l@{}}Rouge-L\\ \end{tabular} &
\begin{tabular}[c]{@{}l@{}}Rouge-1\\ \end{tabular} &
\begin{tabular}[c]{@{}l@{}}Rouge-L \\  \end{tabular} &
\begin{tabular}[c]{@{}l@{}}Rouge-1 \\ \end{tabular} &
\begin{tabular}[c]{@{}l@{}}Rouge-L \\  \end{tabular} &
\begin{tabular}[c]{@{}l@{}}MMLU \\ \end{tabular} &
\begin{tabular}[c]{@{}l@{}}MT-Bench \\ \end{tabular} 
\\ \hline
Vanilla & 0.2544 & 0.1534 & 0.2756 & 0.1617 & 0.2349 & 0.1380 & 0.6618 & 8.1808 \\
Prompting (a) & 0.2568 & 0.1556 & 0.2812 & 0.1638 & 0.2376 & 0.1364 & 0.6635 & 8.3344 \\
Prompting (dbrx) & 0.2464 & 0.1496 & 0.2778 & 0.1618 & 0.2333 & 0.1364 & 0.6611 & 7.9563 \\
MemFree Decode & 0.2456 & 0.1508 & 0.2773 & 0.1621 & 0.2392 & 0.1407 & 0.6618 & 8.2453 \\
GA & 0.2459 & 0.1483 & 0.2569 & 0.1527 & 0.2294 & 0.1353 & 0.4940 & 7.8031 \\
NPO & 0.2466 & 0.1492 & 0.2653 & 0.1527 & 0.2375 & 0.1401 & 0.6481 & 8.0547 \\
Gradient Difference & 0.2493 & 0.1523 & 0.2737 & 0.1577 & 0.2409 & 0.1417 & 0.6544 & 7.9727 \\
TV & 0.2439 & 0.1473 & 0.2587 & 0.1527 & 0.2414 & 0.1430 & 0.5316 & 8.1163 \\
SSU & 0.2391 & 0.1458 & 0.2627 & 0.1509 & 0.2398 & 0.1413 & 0.6481 & 8.3938 \\
\hline
\end{tabular}}
\caption{Overall results of Llama3.1 at time step 3, compared with several baselines for $D_f$, $D_{prev}$, and $D_{nor}$. Benchmark performance includes MMLU and MT-Bench scores.}
\vspace{-0.15in}
\label{tab:main_results_time_step_3}
\end{table*}

\begin{table*}[h]
\centering
\resizebox{1\textwidth}{!}{
\begin{tabular}{l||ll||ll||ll||ll}
\hline
\multicolumn{1}{c||}{\multirow{2}{*}{}} & 
\multicolumn{2}{c||}{\textbf{$\mathbf{D_f}$}} & 
\multicolumn{2}{c||}{\textbf{$\mathbf{D_{prev}}$}} & 
\multicolumn{2}{c||}{\textbf{$\mathbf{D_{nor}}$}} &
\multicolumn{2}{c}{\textbf{Benchmark}} \\ \cline{2-9}

\multicolumn{1}{c||}{} &
\begin{tabular}[c]{@{}l@{}}Rouge-1 \\ \end{tabular} &
\begin{tabular}[c]{@{}l@{}}Rouge-L\\ \end{tabular} &
\begin{tabular}[c]{@{}l@{}}Rouge-1\\ \end{tabular} &
\begin{tabular}[c]{@{}l@{}}Rouge-L \\  \end{tabular} &
\begin{tabular}[c]{@{}l@{}}Rouge-1 \\ \end{tabular} &
\begin{tabular}[c]{@{}l@{}}Rouge-L \\  \end{tabular} &
\begin{tabular}[c]{@{}l@{}}MMLU \\ \end{tabular} &
\begin{tabular}[c]{@{}l@{}}MT-Bench \\ \end{tabular} 
\\ \hline
Vanilla & 0.2589 & 0.1530 & 0.2721 & 0.1588 & 0.2349 & 0.1380 & 0.6618 & 8.1808 \\
Prompting (a) & 0.2572 & 0.1522 & 0.2756 & 0.1609 & 0.2376 & 0.1364 & 0.6635 & 8.3344 \\
Prompting (dbrx) & 0.2597 & 0.1531 & 0.2713 & 0.1583 & 0.2333 & 0.1364 & 0.6611 & 7.9563 \\
MemFree Decode & 0.2591 & 0.1520 & 0.2722 & 0.1605 & 0.2383 & 0.1408 & 0.6617 & 8.2453 \\
GA & 0.2599 & 0.1466 & 0.2533 & 0.1501 & 0.2295 & 0.1355 & 0.4853 & 7.9813 \\
NPO & 0.2480 & 0.1478 & 0.2641 & 0.1542 & 0.2364 & 0.1407 & 0.6537 & 8.0821 \\
Gradient Difference & 0.2606 & 0.1504 & 0.2748 & 0.1598 & 0.2464 & 0.1457 & 0.6579 & 8.0344 \\
TV & 0.2518 & 0.1500 & 0.2564 & 0.1519 & 0.2342 & 0.1396 & 0.4982 & 8.1456 \\
SSU & 0.2489 & 0.1436 & 0.2522 & 0.1516 & 0.2417 & 0.1424 & 0.6432 & 8.2547 \\
\hline
\end{tabular}}
\caption{Overall results of Llama3.1 at time step 4, compared with several baselines for $D_f$, $D_{prev}$, and $D_{nor}$. Benchmark performance includes MMLU and MT-Bench scores.}
\vspace{-0.15in}
\label{tab:main_results_time_step_4}
\end{table*}

\begin{table*}[h]
\centering
\resizebox{1\textwidth}{!}{
\begin{tabular}{l||ll||ll||ll||ll}
\hline
\multicolumn{1}{c||}{\multirow{2}{*}{}} & 
\multicolumn{2}{c||}{\textbf{$\mathbf{D_f}$}} & 
\multicolumn{2}{c||}{\textbf{$\mathbf{D_{prev}}$}} & 
\multicolumn{2}{c||}{\textbf{$\mathbf{D_{nor}}$}} &
\multicolumn{2}{c}{\textbf{Benchmark}} \\ \cline{2-9}

\multicolumn{1}{c||}{} &
\begin{tabular}[c]{@{}l@{}}Rouge-1 \\ \end{tabular} &
\begin{tabular}[c]{@{}l@{}}Rouge-L\\ \end{tabular} &
\begin{tabular}[c]{@{}l@{}}Rouge-1\\ \end{tabular} &
\begin{tabular}[c]{@{}l@{}}Rouge-L \\  \end{tabular} &
\begin{tabular}[c]{@{}l@{}}Rouge-1 \\ \end{tabular} &
\begin{tabular}[c]{@{}l@{}}Rouge-L \\  \end{tabular} &
\begin{tabular}[c]{@{}l@{}}MMLU \\ \end{tabular} &
\begin{tabular}[c]{@{}l@{}}MT-Bench \\ \end{tabular} 
\\ \hline
Vanilla & 0.2708 & 0.1487 & 0.2726 & 0.1595 & 0.2349 & 0.1380 & 0.6618 & 8.1808 \\
Prompting (a) & 0.2720 & 0.1492 & 0.2751 & 0.1617 & 0.2376 & 0.1364 & 0.6635 & 8.3344 \\
Prompting (dbrx) & 0.2677 & 0.1449 & 0.2714 & 0.1615 & 0.2333 & 0.1364 & 0.6611 & 7.9563 \\
MemFree Decode & 0.2724 & 0.1468 & 0.2714 & 0.1599 & 0.2383 & 0.1408 & 0.6617 & 8.2453 \\
GA & 0.2423 & 0.1342 & 0.2544 & 0.1502 & 0.2276 & 0.1320 & 0.3102 & 7.5719 \\
NPO & 0.2489 & 0.1367 & 0.2611 & 0.1508 & 0.2335 & 0.1384 & 0.6196 & 8.0313 \\
Gradient Difference & 0.2617 & 0.1425 & 0.2689 & 0.1582 & 0.2374 & 0.1419 & 0.6399 & 4.9438 \\
TV & 0.2394 & 0.1357 & 0.2571 & 0.1507 & 0.2339 & 0.1403 & 0.4887 & 8.1875 \\
SSU & 0.2515 & 0.1364 & 0.2582 & 0.1508 & 0.2409 & 0.1423 & 0.6425 & 8.1415 \\
\hline
\end{tabular}}
\caption{Overall results of Llama3.1 at time step 5, compared with several baselines for $D_f$, $D_{prev}$, and $D_{nor}$. Benchmark performance includes MMLU and MT-Bench scores.}
\vspace{-0.15in}
\label{tab:main_results_time_step_5}
\end{table*}

\begin{table*}[h]
\centering
\resizebox{1\textwidth}{!}{
\begin{tabular}{l||ll||ll||ll||ll}
\hline
\multicolumn{1}{c||}{\multirow{2}{*}{}} & 
\multicolumn{2}{c||}{\textbf{$\mathbf{D_f}$}} & 
\multicolumn{2}{c||}{\textbf{$\mathbf{D_{prev}}$}} & 
\multicolumn{2}{c||}{\textbf{$\mathbf{D_{nor}}$}} &
\multicolumn{2}{c}{\textbf{Benchmark}} \\ \cline{2-9}

\multicolumn{1}{c||}{} &
\begin{tabular}[c]{@{}l@{}}Rouge-1 \\ \end{tabular} &
\begin{tabular}[c]{@{}l@{}}Rouge-L\\ \end{tabular} &
\begin{tabular}[c]{@{}l@{}}Rouge-1\\ \end{tabular} &
\begin{tabular}[c]{@{}l@{}}Rouge-L \\  \end{tabular} &
\begin{tabular}[c]{@{}l@{}}Rouge-1 \\ \end{tabular} &
\begin{tabular}[c]{@{}l@{}}Rouge-L \\  \end{tabular} &
\begin{tabular}[c]{@{}l@{}}MMLU \\ \end{tabular} &
\begin{tabular}[c]{@{}l@{}}MT-Bench \\ \end{tabular} 
\\ \hline
Vanilla & 0.2602 & 0.1472 & 0.2723 & 0.1558 & 0.2349 & 0.1380 & 0.6618 & 8.1808 \\
Prompting (a) & 0.2603 & 0.1478 & 0.2747 & 0.1565 & 0.2376 & 0.1364 & 0.6635 & 8.3344 \\
Prompting (dbrx) & 0.2567 & 0.1489 & 0.2717 & 0.1552 & 0.2333 & 0.1364 & 0.6611 & 7.9563 \\
MemFree Decode & 0.2503 & 0.1452 & 0.2726 & 0.1551 & 0.2383 & 0.1408 & 0.6617 & 8.2453 \\
GA & 0.2535 & 0.1420 & 0.2499 & 0.1430 & 0.2278 & 0.1323 & 0.3082 & 7.6594 \\
NPO & 0.2502 & 0.1419 & 0.2563 & 0.1472 & 0.2342 & 0.1379 & 0.6018 & 8.0375 \\
Gradient Difference & 0.2455 & 0.1391 & 0.2686 & 0.1525 & 0.2369 & 0.1397 & 0.6232 & 4.5500 \\
TV & 0.2408 & 0.1405 & 0.2518 & 0.1459 & 0.2287 & 0.1399 & 0.3116 & 8.1219 \\
SSU & 0.2462 & 0.1398 & 0.2581 & 0.1463 & 0.2374 & 0.1401 & 0.6298 & 8.2359 \\
\hline
\end{tabular}}
\caption{Overall results of Llama3.1 at time step 6, compared with several baselines for $D_f$, $D_{prev}$, and $D_{nor}$. Benchmark performance includes MMLU and MT-Bench scores.}
\vspace{-0.15in}
\label{tab:main_results_time_step_6}
\end{table*}

\begin{table*}[h]
\centering
\resizebox{1\textwidth}{!}{
\begin{tabular}{l||ll||ll||ll||ll}
\hline
\multicolumn{1}{c||}{\multirow{2}{*}{}} & 
\multicolumn{2}{c||}{\textbf{$\mathbf{D_f}$}} & 
\multicolumn{2}{c||}{\textbf{$\mathbf{D_{prev}}$}} & 
\multicolumn{2}{c||}{\textbf{$\mathbf{D_{nor}}$}} &
\multicolumn{2}{c}{\textbf{Benchmark}} \\ \cline{2-9}

\multicolumn{1}{c||}{} &
\begin{tabular}[c]{@{}l@{}}Rouge-1 \\ \end{tabular} &
\begin{tabular}[c]{@{}l@{}}Rouge-L\\ \end{tabular} &
\begin{tabular}[c]{@{}l@{}}Rouge-1\\ \end{tabular} &
\begin{tabular}[c]{@{}l@{}}Rouge-L \\  \end{tabular} &
\begin{tabular}[c]{@{}l@{}}Rouge-1 \\ \end{tabular} &
\begin{tabular}[c]{@{}l@{}}Rouge-L \\  \end{tabular} &
\begin{tabular}[c]{@{}l@{}}MMLU \\ \end{tabular} &
\begin{tabular}[c]{@{}l@{}}MT-Bench \\ \end{tabular} 
\\ \hline
Vanilla & 0.2678 & 0.1482 & 0.2625 & 0.1507 & 0.2349 & 0.1380 & 0.6618 & 8.1808 \\
Prompting (a) & 0.2674 & 0.1521 & 0.2676 & 0.1528 & 0.2376 & 0.1364 & 0.6635 & 8.3344 \\
Prompting (dbrx) & 0.2602 & 0.1489 & 0.2632 & 0.1535 & 0.2333 & 0.1364 & 0.6611 & 7.9563 \\
MemFree Decode & 0.2488 & 0.1451 & 0.2675 & 0.1559 & 0.2383 & 0.1408 & 0.6617 & 8.2453 \\
GA & 0.2485 & 0.1403 & 0.2473 & 0.1441 & 0.2277 & 0.1324 & 0.2729 & 7.7063 \\
NPO & 0.2534 & 0.1452 & 0.2567 & 0.1465 & 0.2369 & 0.1382 & 0.5786 & 8.0375 \\
Gradient Difference & 0.2637 & 0.1494 & 0.2588 & 0.1492 & 0.2386 & 0.1414 & 0.6112 & 4.1384 \\
TV & 0.2453 & 0.1406 & 0.2433 & 0.1419 & 0.2266 & 0.1365 & 0.3477 & 7.8899 \\
SSU & 0.2532 & 0.1428 & 0.2559 & 0.1457 & 0.2391 & 0.1398 & 0.6291 & 8.1406 \\
\hline
\end{tabular}}
\caption{Overall results of Llama3.1 at time step 7, compared with several baselines for $D_f$, $D_{prev}$, and $D_{nor}$. Benchmark performance includes MMLU and MT-Bench scores.}
\vspace{-0.15in}
\label{tab:main_results_time_step_7}
\end{table*}

\begin{table*}[h]
\centering
\resizebox{1\textwidth}{!}{
\begin{tabular}{l||ll||ll||ll||ll}
\hline
\multicolumn{1}{c||}{\multirow{2}{*}{}} & 
\multicolumn{2}{c||}{\textbf{$\mathbf{D_f}$}} & 
\multicolumn{2}{c||}{\textbf{$\mathbf{D_{prev}}$}} & 
\multicolumn{2}{c||}{\textbf{$\mathbf{D_{nor}}$}} &
\multicolumn{2}{c}{\textbf{Benchmark}} \\ \cline{2-9}

\multicolumn{1}{c||}{} &
\begin{tabular}[c]{@{}l@{}}Rouge-1 \\ \end{tabular} &
\begin{tabular}[c]{@{}l@{}}Rouge-L\\ \end{tabular} &
\begin{tabular}[c]{@{}l@{}}Rouge-1\\ \end{tabular} &
\begin{tabular}[c]{@{}l@{}}Rouge-L \\  \end{tabular} &
\begin{tabular}[c]{@{}l@{}}Rouge-1 \\ \end{tabular} &
\begin{tabular}[c]{@{}l@{}}Rouge-L \\  \end{tabular} &
\begin{tabular}[c]{@{}l@{}}MMLU \\ \end{tabular} &
\begin{tabular}[c]{@{}l@{}}MT-Bench \\ \end{tabular} 
\\ \hline
Vanilla & 0.2906 & 0.1673 & 0.2685 & 0.1514 & 0.2349 & 0.1380 & 0.6618 & 8.1808 \\
Prompting (a) & 0.2912 & 0.1668 & 0.2656 & 0.1538 & 0.2376 & 0.1364 & 0.6635 & 8.3344 \\
Prompting (dbrx) & 0.2902 & 0.1648 & 0.2637 & 0.1521 & 0.2333 & 0.1364 & 0.6611 & 7.9563 \\
MemFree Decode & 0.2922 & 0.1690 & 0.2683 & 0.1543 & 0.2383 & 0.1408 & 0.6617 & 8.2453 \\
GA & 0.2623 & 0.1476 & 0.2471 & 0.1451 & 0.2266 & 0.1325 & 0.2674 & 7.7219 \\
NPO & 0.2668 & 0.1516 & 0.2515 & 0.1434 & 0.2368 & 0.1383 & 0.5783 & 8.0719 \\
Gradient Difference & 0.2786 & 0.1578 & 0.2609 & 0.1513 & 0.2393 & 0.1414 & 0.6112 & 4.4000 \\
TV & 0.2539 & 0.1505 & 0.2347 & 0.1388 & 0.2259 & 0.1377 & 0.3516 & 7.9281 \\
SSU & 0.2676 & 0.1493 & 0.2479 & 0.1438 & 0.2386 & 0.1393 & 0.6263 & 8.2344 \\
\hline
\end{tabular}}
\caption{Overall results of Llama3.1 at time step 8, compared with several baselines for $D_f$, $D_{prev}$, and $D_{nor}$. Benchmark performance includes MMLU and MT-Bench scores.}
\vspace{-0.15in}
\label{tab:main_results_time_step_8}
\end{table*}

\begin{table*}[h]
\centering
\resizebox{1\textwidth}{!}{
\begin{tabular}{l||ll||ll||ll||ll}
\hline
\multicolumn{1}{c||}{\multirow{2}{*}{}} & 
\multicolumn{2}{c||}{\textbf{$\mathbf{D_f}$}} & 
\multicolumn{2}{c||}{\textbf{$\mathbf{D_{prev}}$}} & 
\multicolumn{2}{c||}{\textbf{$\mathbf{D_{nor}}$}} &
\multicolumn{2}{c}{\textbf{Benchmark}} \\ \cline{2-9}

\multicolumn{1}{c||}{} &
\begin{tabular}[c]{@{}l@{}}Rouge-1 \\ \end{tabular} &
\begin{tabular}[c]{@{}l@{}}Rouge-L\\ \end{tabular} &
\begin{tabular}[c]{@{}l@{}}Rouge-1\\ \end{tabular} &
\begin{tabular}[c]{@{}l@{}}Rouge-L \\  \end{tabular} &
\begin{tabular}[c]{@{}l@{}}Rouge-1 \\ \end{tabular} &
\begin{tabular}[c]{@{}l@{}}Rouge-L \\  \end{tabular} &
\begin{tabular}[c]{@{}l@{}}MMLU \\ \end{tabular} &
\begin{tabular}[c]{@{}l@{}}MT-Bench \\ \end{tabular} 
\\ \hline
Vanilla & 0.2725 & 0.1628 & 0.2662 & 0.1564 & 0.2349 & 0.1380 & 0.6618 & 8.1808 \\
Prompting (a) & 0.2726 & 0.1601 & 0.2670 & 0.1528 & 0.2376 & 0.1364 & 0.6635 & 8.3344 \\
Prompting (dbrx) & 0.2697 & 0.1579 & 0.2604 & 0.1520 & 0.2333 & 0.1364 & 0.6611 & 7.9563 \\
MemFree Decode & 0.2695 & 0.1583 & 0.2674 & 0.1552 & 0.2383 & 0.1408 & 0.6617 & 8.2453 \\
GA & 0.2608 & 0.1505 & 0.2480 & 0.1441 & 0.2279 & 0.1343 & 0.1996 & 7.4281 \\
NPO & 0.2619 & 0.1530 & 0.2532 & 0.1448 & 0.2299 & 0.1348 & 0.5488 & 8.0281 \\
Gradient Difference & 0.2594 & 0.1531 & 0.2562 & 0.1482 & 0.2422 & 0.1417 & 0.6305 & 4.3208 \\
TV & 0.0001 & 0.0001 & 0.0005 & 0.0005 & 0.0002 & 0.0002 & 0 & 4.2373 \\
SSU & 0.2629 & 0.1522 & 0.2484 & 0.1427 & 0.2360 & 0.1382 & 0.6049 & 8.1219 \\
\hline
\end{tabular}}
\caption{Overall results of Llama3.1 at time step 9, compared with several baselines for $D_f$, $D_{prev}$, and $D_{nor}$. Benchmark performance includes MMLU and MT-Bench scores.}
\vspace{-0.15in}
\label{tab:main_results_time_step_9}
\end{table*}

\begin{table*}[h]
\centering
\resizebox{1\textwidth}{!}{
\begin{tabular}{l||ll||ll||ll||ll}
\hline
\multicolumn{1}{c||}{\multirow{2}{*}{}} & 
\multicolumn{2}{c||}{\textbf{$\mathbf{D_f}$}} & 
\multicolumn{2}{c||}{\textbf{$\mathbf{D_{prev}}$}} & 
\multicolumn{2}{c||}{\textbf{$\mathbf{D_{nor}}$}} &
\multicolumn{2}{c}{\textbf{Benchmark}} \\ \cline{2-9}

\multicolumn{1}{c||}{} &
\begin{tabular}[c]{@{}l@{}}Rouge-1 \\ \end{tabular} &
\begin{tabular}[c]{@{}l@{}}Rouge-L\\ \end{tabular} &
\begin{tabular}[c]{@{}l@{}}Rouge-1\\ \end{tabular} &
\begin{tabular}[c]{@{}l@{}}Rouge-L \\  \end{tabular} &
\begin{tabular}[c]{@{}l@{}}Rouge-1 \\ \end{tabular} &
\begin{tabular}[c]{@{}l@{}}Rouge-L \\  \end{tabular} &
\begin{tabular}[c]{@{}l@{}}MMLU \\ \end{tabular} &
\begin{tabular}[c]{@{}l@{}}MT-Bench \\ \end{tabular} 
\\ \hline
Vanilla & 0.2667 & 0.1458 & 0.2602 & 0.1467 & 0.2349 & 0.1380 & 0.6618 & 8.1808 \\
Prompting (a) & 0.2605 & 0.1469 & 0.2597 & 0.1477 & 0.2376 & 0.1364 & 0.6635 & 8.3344 \\
Prompting (dbrx) & 0.2622 & 0.1467 & 0.2648 & 0.1510 & 0.2333 & 0.1364 & 0.6611 & 7.9563 \\
MemFree Decode & 0.2596 & 0.1450 & 0.2672 & 0.1522 & 0.2383 & 0.1408 & 0.6617 & 8.2453 \\
GA & 0.2559 & 0.1422 & 0.2491 & 0.1422 & 0.2289 & 0.1335 & 0.2004 & 7.4344 \\
NPO & 0.2583 & 0.1435 & 0.2602 & 0.1498 & 0.2306 & 0.1342 & 0.5477 & 7.9969 \\
Gradient Difference & 0.2542 & 0.1453 & 0.2601 & 0.1496 & 0.2365 & 0.1383 & 0.6079 & 4.4843 \\
TV & 0 & 0 & 0.0005 & 0.0005 & 0.0002 & 0.0002 & 0 & 3.915 \\
SSU & 0.2481 & 0.1389 & 0.2541 & 0.1439 & 0.2333 & 0.1396 & 0.6023 & 8.0206 \\
\hline
\end{tabular}}
\caption{Overall results of Llama3.1 at time step 10, compared with several baselines for $D_f$, $D_{prev}$, and $D_{nor}$. Benchmark performance includes MMLU and MT-Bench scores.}
\vspace{-0.15in}
\label{tab:main_results_time_step_10}
\end{table*}

\begin{table*}[h]
\centering
\resizebox{1\textwidth}{!}{
\begin{tabular}{l||ll||ll||ll||ll}
\hline
\multicolumn{1}{c||}{\multirow{2}{*}{}} & 
\multicolumn{2}{c||}{\textbf{$\mathbf{D_f}$}} & 
\multicolumn{2}{c||}{\textbf{$\mathbf{D_{prev}}$}} & 
\multicolumn{2}{c||}{\textbf{$\mathbf{D_{nor}}$}} &
\multicolumn{2}{c}{\textbf{Benchmark}} \\ \cline{2-9}

\multicolumn{1}{c||}{} &
\begin{tabular}[c]{@{}l@{}}Rouge-1 \\ \end{tabular} &
\begin{tabular}[c]{@{}l@{}}Rouge-L\\ \end{tabular} &
\begin{tabular}[c]{@{}l@{}}Rouge-1\\ \end{tabular} &
\begin{tabular}[c]{@{}l@{}}Rouge-L \\  \end{tabular} &
\begin{tabular}[c]{@{}l@{}}Rouge-1 \\ \end{tabular} &
\begin{tabular}[c]{@{}l@{}}Rouge-L \\  \end{tabular} &
\begin{tabular}[c]{@{}l@{}}MMLU \\ \end{tabular} &
\begin{tabular}[c]{@{}l@{}}MT-Bench \\ \end{tabular} 
\\ \hline
Vanilla & 0.2828 & 0.1629 & 0 & 0 & 0.2456 & 0.1487 & 0.6070 & 7.4563 \\
Prompting (a) & 0.2731 & 0.1596 & 0 & 0 & 0.2481 & 0.1515 & 0.6074 & 7.1438 \\
Prompting (dbrx) & 0.2783 & 0.1649 & 0 & 0 & 0.2412 & 0.1482 & 0.6053 & 7.3531 \\
MemFree Decode & 0.2742 & 0.1640 & 0 & 0 & 0.2458 & 0.1502 & 0.6074 & 7.1719 \\
GA & 0.0739 & 0.0501 & 0 & 0 & 0.1183 & 0.0750 & 0.6028 & 6.5875 \\
NPO & 0.2721 & 0.1561 & 0 & 0 & 0.2447 & 0.1466 & 0.6028 & 7.5547 \\
Gradient Difference & 0.2472 & 0.1462 & 0 & 0 & 0.2351 & 0.1413 & 0.6074 & 7.3875 \\
TV & 0.2591 & 0.1539 & 0 & 0 & 0.2391 & 0.1469 & 0.6021 & 7.1125 \\
SSU & 0.2536 & 0.1536 & 0 & 0 & 0.2401 & 0.1495 & 0.6052 & 7.4406 \\
\hline
\end{tabular}}
\caption{Overall results of Mistral-7B at time step 1, compared with several baselines for $D_f$, $D_{prev}$, and $D_{nor}$. Benchmark performance includes MMLU and MT-Bench scores.}
\vspace{-0.15in}
\label{tab:main_results_mistral_time_step_1}
\end{table*}

\begin{table*}[h]
\centering
\resizebox{1\textwidth}{!}{
\begin{tabular}{l||ll||ll||ll||ll}
\hline
\multicolumn{1}{c||}{\multirow{2}{*}{}} & 
\multicolumn{2}{c||}{\textbf{$\mathbf{D_f}$}} & 
\multicolumn{2}{c||}{\textbf{$\mathbf{D_{prev}}$}} & 
\multicolumn{2}{c||}{\textbf{$\mathbf{D_{nor}}$}} &
\multicolumn{2}{c}{\textbf{Benchmark}} \\ \cline{2-9}

\multicolumn{1}{c||}{} &
\begin{tabular}[c]{@{}l@{}}Rouge-1 \\ \end{tabular} &
\begin{tabular}[c]{@{}l@{}}Rouge-L\\ \end{tabular} &
\begin{tabular}[c]{@{}l@{}}Rouge-1\\ \end{tabular} &
\begin{tabular}[c]{@{}l@{}}Rouge-L \\  \end{tabular} &
\begin{tabular}[c]{@{}l@{}}Rouge-1 \\ \end{tabular} &
\begin{tabular}[c]{@{}l@{}}Rouge-L \\  \end{tabular} &
\begin{tabular}[c]{@{}l@{}}MMLU \\ \end{tabular} &
\begin{tabular}[c]{@{}l@{}}MT-Bench \\ \end{tabular} 
\\ \hline
Vanilla & 0.2997 & 0.1838 & 0.2797 & 0.1650 & 0.2456 & 0.1487 & 0.6070 & 7.4563 \\
Prompting (a) & 0.3038 & 0.1842 & 0.2776 & 0.1653 & 0.2481 & 0.1515 & 0.6074 & 7.1438 \\
Prompting (dbrx) & 0.3087 & 0.1867 & 0.2850 & 0.1672 & 0.2412 & 0.1482 & 0.6053 & 7.3531 \\
MemFree Decode & 0.3021 & 0.1841 & 0.2809 & 0.1661 & 0.2458 & 0.1502 & 0.6074 & 7.1719 \\
GA & 0.0544 & 0.0357 & 0.0177 & 0.0118 & 0.0347 & 0.0233 & 0.6039 & 4.0313 \\
NPO & 0.2977 & 0.1779 & 0.2764 & 0.1605 & 0.2453 & 0.1474 & 0.6028 & 7.3438 \\
Gradient Difference & 0.2908 & 0.1728 & 0.2668 & 0.1505 & 0.2373 & 0.1412 & 0.6077 & 7.5313 \\
TV & 0.2682 & 0.1672 & 0.2454 & 0.1451 & 0.2258 & 0.1377 & 0.5944 & 7.0469 \\
SSU & 0.2871 & 0.1726 & 0.2600 & 0.1537 & 0.2357 & 0.1451 & 0.6028 & 7.3000 \\
\hline
\end{tabular}}
\caption{Overall results of Mistral-7B at time step 2, compared with several baselines for $D_f$, $D_{prev}$, and $D_{nor}$. Benchmark performance includes MMLU and MT-Bench scores.}
\vspace{-0.15in}
\label{tab:main_results_mistral_time_step_2}
\end{table*}

\begin{table*}[h]
\centering
\resizebox{1\textwidth}{!}{
\begin{tabular}{l||ll||ll||ll||ll}
\hline
\multicolumn{1}{c||}{\multirow{2}{*}{}} & 
\multicolumn{2}{c||}{\textbf{$\mathbf{D_f}$}} & 
\multicolumn{2}{c||}{\textbf{$\mathbf{D_{prev}}$}} & 
\multicolumn{2}{c||}{\textbf{$\mathbf{D_{nor}}$}} &
\multicolumn{2}{c}{\textbf{Benchmark}} \\ \cline{2-9}

\multicolumn{1}{c||}{} &
\begin{tabular}[c]{@{}l@{}}Rouge-1 \\ \end{tabular} &
\begin{tabular}[c]{@{}l@{}}Rouge-L\\ \end{tabular} &
\begin{tabular}[c]{@{}l@{}}Rouge-1\\ \end{tabular} &
\begin{tabular}[c]{@{}l@{}}Rouge-L \\  \end{tabular} &
\begin{tabular}[c]{@{}l@{}}Rouge-1 \\ \end{tabular} &
\begin{tabular}[c]{@{}l@{}}Rouge-L \\  \end{tabular} &
\begin{tabular}[c]{@{}l@{}}MMLU \\ \end{tabular} &
\begin{tabular}[c]{@{}l@{}}MT-Bench \\ \end{tabular} 
\\ \hline
Vanilla & 0.2523 & 0.1562 & 0.2943 & 0.1755 & 0.2456 & 0.1487 & 0.6070 & 7.4563 \\
Prompting (a) & 0.2440 & 0.1518 & 0.2928 & 0.1742 & 0.2481 & 0.1515 & 0.6074 & 7.1438 \\
Prompting (dbrx) & 0.2429 & 0.1561 & 0.2883 & 0.1746 & 0.2412 & 0.1482 & 0.6053 & 7.3531 \\
MemFree Decode & 0.2401 & 0.1527 & 0.2906 & 0.1758 & 0.2458 & 0.1502 & 0.6074 & 7.1719 \\
GA & 0.0313 & 0.0205 & 0.0274 & 0.0193 & 0.0323 & 0.0216 & 0.6053 & 3.9056 \\
NPO & 0.2448 & 0.1486 & 0.2847 & 0.1677 & 0.2434 & 0.1474 & 0.6046 & 7.2438 \\
Gradient Difference & 0.2392 & 0.1405 & 0.2719 & 0.1505 & 0.2327 & 0.1383 & 0.6049 & 7.2844 \\
TV & 0.2054 & 0.1319 & 0.2404 & 0.1479 & 0.2095 & 0.1314 & 0.5846 & 6.8469 \\
SSU & 0.2204 & 0.1390 & 0.2623 & 0.1601 & 0.2333 & 0.1426 & 0.5996 & 7.3437 \\
\hline
\end{tabular}}
\caption{Overall results of Mistral-7B at time step 3, compared with several baselines for $D_f$, $D_{prev}$, and $D_{nor}$. Benchmark performance includes MMLU and MT-Bench scores.}
\vspace{-0.15in}
\label{tab:main_results_mistral_time_step_3}
\end{table*}

\begin{table*}[h]
\centering
\resizebox{1\textwidth}{!}{
\begin{tabular}{l||ll||ll||ll||ll}
\hline
\multicolumn{1}{c||}{\multirow{2}{*}{}} & 
\multicolumn{2}{c||}{\textbf{$\mathbf{D_f}$}} & 
\multicolumn{2}{c||}{\textbf{$\mathbf{D_{prev}}$}} & 
\multicolumn{2}{c||}{\textbf{$\mathbf{D_{nor}}$}} &
\multicolumn{2}{c}{\textbf{Benchmark}} \\ \cline{2-9}

\multicolumn{1}{c||}{} &
\begin{tabular}[c]{@{}l@{}}Rouge-1 \\ \end{tabular} &
\begin{tabular}[c]{@{}l@{}}Rouge-L\\ \end{tabular} &
\begin{tabular}[c]{@{}l@{}}Rouge-1\\ \end{tabular} &
\begin{tabular}[c]{@{}l@{}}Rouge-L \\  \end{tabular} &
\begin{tabular}[c]{@{}l@{}}Rouge-1 \\ \end{tabular} &
\begin{tabular}[c]{@{}l@{}}Rouge-L \\  \end{tabular} &
\begin{tabular}[c]{@{}l@{}}MMLU \\ \end{tabular} &
\begin{tabular}[c]{@{}l@{}}MT-Bench \\ \end{tabular} 
\\ \hline
Vanilla & 0.2702 & 0.1660 & 0.2929 & 0.1753 & 0.2456 & 0.1487 & 0.6070 & 7.4563 \\
Prompting (a) & 0.2730 & 0.1650 & 0.2863 & 0.1739 & 0.2481 & 0.1515 & 0.6074 & 7.1438 \\
Prompting (dbrx) & 0.2726 & 0.1687 & 0.2817 & 0.1726 & 0.2412 & 0.1482 & 0.6053 & 7.3531 \\
MemFree Decode & 0.2711 & 0.1628 & 0.2846 & 0.1748 & 0.2458 & 0.1502 & 0.6074 & 7.1719 \\
GA & 0 & 0 & 0 & 0 & 0 & 0 & 0 & 0 \\
NPO & 0.2618 & 0.1561 & 0.2806 & 0.1665 & 0.2447 & 0.1475 & 0.6021 & 7.2688 \\
Gradient Difference & 0.2503 & 0.1464 & 0.2694 & 0.1595 & 0.2326 & 0.1408 & 0.6042 & 7.2375 \\
TV & 0.2038 & 0.1266 & 0.2131 & 0.1306 & 0.1929 & 0.1189 & 0.5582 & 6.4938 \\
SSU & 0.2431 & 0.1489 & 0.2553 & 0.1561 & 0.2352 & 0.1434 & 0.6000 & 7.2344 \\
\hline
\end{tabular}}
\caption{Overall results of Mistral-7B at time step 4, compared with several baselines for $D_f$, $D_{prev}$, and $D_{nor}$. Benchmark performance includes MMLU and MT-Bench scores.}
\vspace{-0.15in}
\label{tab:main_results_mistral_time_step_4}
\end{table*}

\begin{table*}[h]
\centering
\resizebox{1\textwidth}{!}{
\begin{tabular}{l||ll||ll||ll||ll}
\hline
\multicolumn{1}{c||}{\multirow{2}{*}{}} & 
\multicolumn{2}{c||}{\textbf{$\mathbf{D_f}$}} & 
\multicolumn{2}{c||}{\textbf{$\mathbf{D_{prev}}$}} & 
\multicolumn{2}{c||}{\textbf{$\mathbf{D_{nor}}$}} &
\multicolumn{2}{c}{\textbf{Benchmark}} \\ \cline{2-9}

\multicolumn{1}{c||}{} &
\begin{tabular}[c]{@{}l@{}}Rouge-1 \\ \end{tabular} &
\begin{tabular}[c]{@{}l@{}}Rouge-L\\ \end{tabular} &
\begin{tabular}[c]{@{}l@{}}Rouge-1\\ \end{tabular} &
\begin{tabular}[c]{@{}l@{}}Rouge-L \\  \end{tabular} &
\begin{tabular}[c]{@{}l@{}}Rouge-1 \\ \end{tabular} &
\begin{tabular}[c]{@{}l@{}}Rouge-L \\  \end{tabular} &
\begin{tabular}[c]{@{}l@{}}MMLU \\ \end{tabular} &
\begin{tabular}[c]{@{}l@{}}MT-Bench \\ \end{tabular} 
\\ \hline
Vanilla & 0.2709 & 0.1555 & 0.2829 & 0.1703 & 0.2456 & 0.1487 & 0.6070 & 7.4563 \\
Prompting (a) & 0.2673 & 0.1530 & 0.2904 & 0.1719 & 0.2481 & 0.1515 & 0.6074 & 7.1438 \\
Prompting (dbrx) & 0.2673 & 0.1561 & 0.2835 & 0.1714 & 0.2412 & 0.1482 & 0.6053 & 7.3531 \\
MemFree Decode & 0.2624 & 0.1515 & 0.2833 & 0.1735 & 0.2458 & 0.1502 & 0.6074 & 7.1719 \\
GA & 0 & 0 & 0 & 0 & 0 & 0 & 0 & 0 \\
NPO & 0.2633 & 0.1490 & 0.2819 & 0.1647 & 0.2423 & 0.1462 & 0.5986 & 7.4719 \\
Gradient Difference & 0.2503 & 0.1464 & 0.0483 & 0.0336 & 0.0491 & 0.0338 & 0.6063 & 2.8375 \\
TV & 0.1703 & 0.1078 & 0.1674 & 0.1099 & 0.1468 & 0.0986 & 0 & 1.0000 \\
SSU & 0.2324 & 0.1402 & 0.2571 & 0.1575 & 0.2333 & 0.1427 & 0.6000 & 7.3469 \\
\hline
\end{tabular}}
\caption{Overall results of Mistral-7B at time step 5, compared with several baselines for $D_f$, $D_{prev}$, and $D_{nor}$. Benchmark performance includes MMLU and MT-Bench scores.}
\vspace{-0.15in}
\label{tab:main_results_mistral_time_step_5}
\end{table*}

\begin{table*}[h]
\centering
\resizebox{1\textwidth}{!}{
\begin{tabular}{l||ll||ll||ll||ll}
\hline
\multicolumn{1}{c||}{\multirow{2}{*}{}} & 
\multicolumn{2}{c||}{\textbf{$\mathbf{D_f}$}} & 
\multicolumn{2}{c||}{\textbf{$\mathbf{D_{prev}}$}} & 
\multicolumn{2}{c||}{\textbf{$\mathbf{D_{nor}}$}} &
\multicolumn{2}{c}{\textbf{Benchmark}} \\ \cline{2-9}

\multicolumn{1}{c||}{} &
\begin{tabular}[c]{@{}l@{}}Rouge-1 \\ \end{tabular} &
\begin{tabular}[c]{@{}l@{}}Rouge-L\\ \end{tabular} &
\begin{tabular}[c]{@{}l@{}}Rouge-1\\ \end{tabular} &
\begin{tabular}[c]{@{}l@{}}Rouge-L \\  \end{tabular} &
\begin{tabular}[c]{@{}l@{}}Rouge-1 \\ \end{tabular} &
\begin{tabular}[c]{@{}l@{}}Rouge-L \\  \end{tabular} &
\begin{tabular}[c]{@{}l@{}}MMLU \\ \end{tabular} &
\begin{tabular}[c]{@{}l@{}}MT-Bench \\ \end{tabular} 
\\ \hline
Vanilla & 0.2694 & 0.1610 & 0.2778 & 0.1628 & 0.2456 & 0.1487 & 0.6070 & 7.4563 \\
Prompting (a) & 0.2653 & 0.1580 & 0.2806 & 0.1632 & 0.2481 & 0.1515 & 0.6074 & 7.1438 \\
Prompting (dbrx) & 0.2692 & 0.1617 & 0.2862 & 0.1679 & 0.2412 & 0.1482 & 0.6053 & 7.3531 \\
MemFree Decode & 0.2662 & 0.1585 & 0.2758 & 0.1638 & 0.2458 & 0.1502 & 0.6074 & 7.1719 \\
GA & 0 & 0 & 0 & 0 & 0 & 0 & 0 & 0 \\
NPO & 0.2711 & 0.1572 & 0.2714 & 0.1555 & 0.2457 & 0.1482 & 0.5979 & 7.4119 \\
Gradient Difference & 0.0273 & 0.0191 & 0.0270 & 0.0198 & 0.0485 & 0.0326 & 0.6070 & 2.6815 \\
TV & 0 & 0 & 0 & 0 & 0 & 0 & 0 & 0 \\
SSU & 0.2519 & 0.1476 & 0.2448 & 0.1463 & 0.2304 & 0.1416 & 0.5982 & 7.2938 \\
\hline
\end{tabular}}
\caption{Overall results of Mistral-7B at time step 6, compared with several baselines for $D_f$, $D_{prev}$, and $D_{nor}$. Benchmark performance includes MMLU and MT-Bench scores.}
\vspace{-0.15in}
\label{tab:main_results_mistral_time_step_6}
\end{table*}

\end{document}